\documentclass[10pt,journal,compsoc]{IEEEtran}

\usepackage{times}
\usepackage{epsfig}
\usepackage{graphicx}
\usepackage[cmex10]{amsmath}
\usepackage{amssymb}

\usepackage{booktabs}
\usepackage{placeins}
\usepackage{latexsym} 
\usepackage{subfigure}
\usepackage{url}
\usepackage{color} 
\usepackage{overpic}
\usepackage[table]{xcolor}
\usepackage{rotating}

\usepackage{epstopdf}
\usepackage{multirow}
\usepackage[ruled,vlined]{algorithm2e}
\SetCommentSty{mycommfont}
\usepackage[normalem]{ulem}

\newcommand{\ADD}[1]{{#1}}

\newcommand{\regressor}{regressor}

\ifCLASSOPTIONcompsoc
  \usepackage[nocompress]{cite}
\else
  \usepackage{cite}
\fi

\ifCLASSINFOpdf
\else
\fi

\begin{document}

\title{Single and Multiple Illuminant Estimation Using Convolutional
  Neural Networks}

\author{Simone~Bianco,
        Claudio~Cusano,
        and~Raimondo~Schettini
\IEEEcompsocitemizethanks{\IEEEcompsocthanksitem S. Bianco and R. Schettini are with the Department
of Informatics, Systems and Communication, University of Milano Bicocca,
Italy.\protect\\
E-mail: \{simone.bianco,raimondo.schettini\}@disco.unimib.it
\IEEEcompsocthanksitem C. Cusano is with the Department of Electrical,
Computer and Biomedical Engineering of the University of Pavia, Italy. \protect\\ E-mail: claudio.cusano@unipv.it}
}



\IEEEtitleabstractindextext{%
\begin{abstract}
  In this paper we present a method for the estimation of the color of
  the illuminant in RAW images.  The method includes a Convolutional
  Neural Network that has been specially designed to produce multiple
  local estimates.  A multiple illuminant detector determines whether
  or not the local outputs of the network must be aggregated into a
  single estimate.  We evaluated our method on standard datasets with
  single and multiple illuminants, obtaining lower estimation errors
  with respect to those obtained by other general purpose methods in the state of the
  art.
\end{abstract}

}

\maketitle

\IEEEdisplaynontitleabstractindextext

%
\IEEEpeerreviewmaketitle

\ifCLASSOPTIONcompsoc
\IEEEraisesectionheading{\section{Introduction}\label{sec:introduction}}
\else
\section{Introduction}
\label{sec:introduction}
\fi
\IEEEPARstart{T}{he} observed color of the objects in the scene depends on the intrinsic color of the object (i.e. the surface spectral reflectance), on the illumination, and on their relative positions. Many computer vision problems in both still images and videos can make use of color constancy processing as a pre-processing step to make sure that the recorded color of the objects in the scene does not change under different illumination conditions.

In general there are two methodologies to obtain reliable color description
from image data: computational color constancy and color invariance \cite{lee2014taxonomy}.
Computational color constancy is a two-stage operation: the former is specialized on estimating the color of the scene illuminant from the image data, the latter corrects the image on the basis of this estimate to generate a new image of the scene as if it was taken under a reference illuminant. 
Color invariance methods instead represent images by features which remain unchanged with respect to imaging conditions.

In this work we focus on illuminant estimation.  Our method is based
on supervised learning and includes a Convolutional Neural Network
(CNN) specially designed for the local estimation of the illuminant
color. Recently, deep neural networks have gained the attention of
numerous researchers outperforming state-of-the-art approaches on
various computer vision tasks
\cite{kavukcuoglu2010learning,krizhevsky2012imagenet}.  One of CNN’s
advantages is that it can take raw images as input and incorporate
feature design into the training process. With a deep structure, CNN
can learn complicated mappings while requiring minimal domain
knowledge.  In our method the outputs of the CNN provide a spatially
varying estimate of the illuminant that can optionally be aggregated
into a single global estimate by a local-to-global \regressor{} based
on non-linear Support Vector Regression (SVR).  To make a final
decision between the local and the global estimates we designed a
multiple illuminant detector exploiting a Kernel Density Estimator
(KDE).  To the best of our knowledge this is the first general purpose work in which
both single and multiple illuminants are dealt with in a comprehensive
way.

Preliminary findings reported in this paper appeared in
\cite{bianco2015color}, where we presented the basic architecture of
the CNN and evaluated its performance in the single
illuminant scenario.  This paper extends the previous one in several
ways:
\begin{itemize}
\item since one of the assumptions that is often violated in color
  constancy is the presence of a uniform illumination in the scene, we
  have extended the applicability of the proposed algorithm to the
  case of nonuniform illumination. The method is adaptive, being able
  to distinguish and process in different ways images of scenes taken
  under a uniform and those acquired under non-uniform
  illumination. 
\item In the case of uniform illumination, the multiple local
  estimates must be aggregated in a single global estimate.  To do
  so we designed a new local-to-global regression method that replaces
  the per-channel median operator used in \cite{bianco2015color} with
  a non-linear mapping based on a RBF kernel over local statistics of
  the CNN estimates.  The parameters of the mapping are obtained by
  applying a regression procedure that minimizes the median angular
  error on the training set.
\item Preliminary results reported in \cite{bianco2015color} included
  only images having a single color target in the scene, thus allowing
  only the comparisons with global illuminant estimation methods. We
  present a much more detailed experimental evaluation using both a
  multiple illuminant synthetic dataset and a dataset of RAW images
  containing at least two known color targets for benchmarking. 
\end{itemize}

We show experimentally that the proposed method advances the
state-of-the-art on standard datasets of RAW images for both the cases
of single and multiple illuminants.  

The rest of the paper is organized as follows:
Section~\ref{sec:problem} formalizes the problem of illuminant
estimation and reviews the main approaches in the state of the art.
Section~\ref{sec:method} illustrates in detail the proposed method.
Section~\ref{sec:setup} describes the data and the algorithms used in
the experimentation, while Section~\ref{sec:results} discusses the
results obtained.  
\ADD{Section~\ref{sec:netarch} reviews the architecture of the CNN on which our method is based, and gives insights on the learned model from a computational color constancy point of view. }
Finally, Section~\ref{sec:conclusions} summarizes
the findings of our experimentation and proposes new directions of research in this field.

%
%
\section{{Problem formulation and related works}}
\label{sec:problem}
The image values for a Lambertian surface located at the pixel with coordinates $(x,y)$ can be seen as a function $\boldsymbol{\rho}(x,y)$, mainly dependent on three physical factors: the illuminant spectral power distribution $I(x,y,\lambda)$, the surface spectral reflectance $S(x,y,\lambda)$ and the sensor spectral sensitivities $\mathbf{C}(\lambda)$. Using this notation $\boldsymbol{\rho}(x,y)$ can be expressed as
\begin{equation}
  \boldsymbol{\rho}(x,y) = \int  I(x,y,\lambda) S(x,y,\lambda) \mathbf{C}(\lambda) \mathrm{d}\lambda,
\label{eq:colorformationequation}
\end{equation}
{where $\lambda$ is the wavelength}, $\boldsymbol{\rho}$ and $\mathbf{C}(\lambda)$ are three-component vectors and the integration is performed over the visible spectrum. 
The goal of color constancy is to estimate the color $\mathbf{I}(x,y)$ of the scene illuminant, i.e. the projection of $I(x,y,\lambda)$ on the sensor spectral sensitivities $\mathbf{C}(\lambda)$:
\begin{equation}
  \mathbf{I}(x,y) = \int  I(x,y,\lambda) \mathbf{C}(\lambda) \mathrm{d}\lambda.
\end{equation}
{
{Usually the illuminant color is estimated up to a scale factor as it is more important to estimate the chromaticity of the scene illuminant than its overall intensity \cite{HorFin04}}. 
}
%
%
Thus, the error metric usually considered, as suggested by Hordley and
Finlayson \cite{HorFin04}, is the angle between the RGB triplet of
estimated illuminant ($\hat{\mathbf{I}}(x,y)$) and the RGB triplet of
the measured ground truth illuminant ($\mathbf{I}(x,y)$):
\begin{equation}
  e_{\text{ANG}}(x,y) = \arccos \left( \frac{\mathbf{I}(x,y)^t \hat{\mathbf{I}}(x,y)}{\| \mathbf{I}(x,y) \|  \| \hat{\mathbf{I}}(x,y)\| } \right) .
\end{equation}

Since the only information available are the sensor responses $\boldsymbol{\rho}$ across the image, color constancy is an under-determined problem \cite{FunBarMar98} and thus further assumptions and/or knowledge are needed to solve it. 
{
Several computational color constancy algorithms have been proposed, each based on different assumptions. The most common assumption is that the color of the light source is uniform across the scene, i.e. $\mathbf{I}(x,y)=\mathbf{I}$.} The next two sections review single and multiple illuminant estimation algorithms in the state of the art.

\subsection{Single illuminant estimation}

Methods for single illuminant estimation can be divided into two main classes: statistic approaches, and learning-based
approaches. Statistic approaches estimate the scene illumination only on the base of the content in a single
image making assumptions about the nature of color images exploiting statistical or physical
properties; learning-based approaches require training
data in order to build a statistical image model, before the estimation of the illumination. 

\subsubsection*{Statistic-based algorithms}
Van~de~Weijer et al. \cite{vandeWGev07} have unified a variety of algorithms. These algorithms estimate the illuminant color $\mathbf{I}$ by implementing instantiations of the following equation:
\begin{equation}
  \label{eq:general-method}
  \mathbf{I}(n, p, \sigma) = \frac{1}{k} \left( \iint \left| \nabla^n
  \boldsymbol{\rho}_\sigma(x,y) \right|^p \mathrm{d}x \ \mathrm{d}y
  \right)^{\frac{1}{p}},
\end{equation}
where $n$ is the order of the derivative, $p$ is the Minkowski norm, $\boldsymbol{\rho}_\sigma(x,y) = \boldsymbol{\rho}(x,y) \otimes G_\sigma(x,y)$ is the convolution of the image with a Gaussian filter $G_\sigma(x,y)$ with scale parameter $\sigma$, and $k$ is a constant to be chosen such that the illuminant color $\mathbf{I}$ has unit length ({using the $2-$norm}). The integration is performed over all pixel coordinates. Different $(n,p,\sigma)$ combinations correspond to different illuminant estimation algorithms, each based on a different assumption. For example, the Gray World algorithm \cite{GW} --- generated by setting $(n,p,\sigma)=(0,1,0)$ --- is based on the assumption that the average color in the image is gray and that the illuminant color can be estimated as the shift from gray of the averages in the image color channels; the White Patch algorithm \cite{retinex} --- generated by setting $(n,p,\sigma)=(0,\infty,0)$ --- is based on the assumption that the maximum response is caused by a perfect reflectance: a surface with perfect
reflectance properties will reflect the full range of light that it
captures and consequently, the color of this perfect reflectance is
exactly the color of the light source. In practice, the assumption
of perfect reflectance is alleviated by considering the color
channels separately, resulting in the maxRGB algorithm. The Gray Edge algorithm \cite{vandeWGev07} --- generated by setting for example $(n,p,\sigma)=(1,0,0)$ --- is based on the assumption that the average color of the edges is gray and that the illuminant color can be estimated as the shift from gray of the averages of the edges in the image color channels.

The Gamut Mapping method does not follows (\ref{eq:general-method})
and assumes that, for a given illuminant, one observes only a limited
gamut of colors \cite{gamutmapF}. It has a preliminary phase in which
a canonical illuminant is chosen and the canonical gamut is computed
observing as many surfaces under the canonical illuminant as
possible. Given an input image with an unknown illuminant, its gamut
is computed and the illuminant is estimated as the mapping that can be
applied to the gamut of the input image, resulting in a gamut that
lies completely within the canonical gamut and produces the most
colorful scene. If the spectral sensitivity functions of the camera
are known, the Color by Correlation approach could be also used
\cite{CbC}.

\subsubsection*{Learning-based algorithms}
The learning-based illuminant estimation algorithms, that estimate the
scene illuminant using a model that is learned on training data, can
be subdivided into two main subcategories: probabilistic methods and
fusion/selection based methods. 

One of the first learning-based algorithms is \cite{funt1996learning},
where a Neural Network was trained on binarized chromaticity
histograms: input neurons are set either to zero indicating that a
chromaticity is not present in the image, or to one indicating that it
is present.

Bayesian approaches \cite{cambridge}
model the variability of reflectance and of illuminant as random
variables, and then estimate illuminant from the posterior
distribution conditioned on image intensity data.

Given a set illuminant estimation algorithms, in \cite{BiaCioCusSch08}
an image classifier is trained to classify the images as indoor and
outdoor, and different experimental frameworks are proposed to exploit
this information in order to select the best performing algorithm on
each class.  In \cite{BiancoPR} it has been shown how intrinsic, low
level properties of the images can be used to drive the selection of
the best algorithm (or the best combination of algorithms) for a given
image. The algorithm selection and combination is made by a decision
forest composed of several trees on the basis of the values of a set
of heterogeneous features.
In \cite{NIS2011} the Weibull parametrization has been used to train a maximum likelihood classifier based on mixture of Gaussians to select the best performing illuminant estimation method for a certain image.

In \cite{chzcc2011} a statistical model for the spatial distribution of colors in white balanced images is developed, and then used to infer illumination parameters as those being most likely under their model.
High level visual information has been used to select the best illuminant out of a set of possible illuminants \cite{hilevelinfo}. This is achieved by restating the problem in terms of semantic interpretability of the image.
Several illuminant estimation methods are applied to generate a set of illuminant hypotheses. For each illuminant hypothesis, they correct the image, evaluate the likelihood of the semantic content of the corrected image, and select the most likely illuminant color. 
In \cite{bianco2012color,bianco2014adaptive} the use of automatically detected objects having intrinsic color is investigated. In particular, they shown how illuminant estimation can be performed exploiting the color statistics extracted from the faces automatically detected in the image. 
When no faces are detected in the image, any other algorithm in the state-of-the-art can be used.  
In \cite{joze2012exemplar,joze2014exemplar} the surfaces in the image are exploited and the illuminant estimation problem is addresses by unsupervised learning of an appropriate model for each training surface in training images. The model for each surface is defined using both texture features and color features.
In a test image the nearest neighbor model is found for each surface and its illumination is estimated  by comparing the statistics of pixels belonging to nearest neighbor surfaces  and the target surface. The final illumination estimation results from combining these estimated illuminants over surfaces to generate a unique estimate.

In \cite{finlayson2013corrected} it was showed how simple moment based
algorithms can, with the addition of a simple correction step deliver
much improved illuminant estimation performance. The approach employs
first, second and higher moments of color and color derivatives and
linearly corrects them to give an illuminant estimate.

In \cite{cheng2015effective} four simple image features are used for
training an ensemble of decision trees. Each of these trees is
computed from samples in the training data that are biased to a local
region in chromaticity space of the ground truth illuminations. The
final estimate is made by finding consensus among the different
features’ trees estimations.

In \cite{bianco2015color} two different approaches using CNNs were
investigated: in the first one an ad-hoc CNN for the color constancy
problem was trained; in the second one a pre-trained one was used by
extracting a 4096-dimensional feature vector from each image using the
Caffe \cite{jia2014caffe} implementation of the deep CNN described by
Krizhevsky et al. \cite{krizhevsky2012imagenet}. Features were
computed by forward propagation of a mean-subtracted $227 \times 227$
RGB RAW image through five convolutional layers and two fully
connected layers. More details about the network architecture can be
found in \cite{krizhevsky2012imagenet,jia2014caffe}.  The CNN was
discriminatively trained on a large dataset (ILSVRC 2012) with
image-level annotations to classify images into 1000 different
classes. Features are obtained by extracting activation values of the
last hidden layer. The extracted features were then used as input to a
linear Support Vector Regression (SVR) \cite{SVR} to estimate the
illuminant color for each image.

In \cite{chakrabarti2015color} illuminant color is predicted from
luminance-to-chromaticity based on a conditional likelihood function
for the true chromaticity of a pixel, given its luminance. Two
approaches have been proposed to learn this function. The first was
based purely on empirical pixel statistics, while the second was based
on maximizing accuracy of the final illuminant estimate.

\subsection{Multiple illuminant estimation}
The great majority of state-of-the-art illuminant estimation methods assumes that a uniform illumination is present in the scene. This assumption is often violated in real-world images. It is not trivial to extend the existing illuminant estimation algorithms to work locally instead of globally, since the spatial support on which they accumulate the statistics is reduced, and the final local estimate could be biased by local image properties. 
One of the first methods following this strategy is Retinex~\cite{retinex}, which is able to deal with non-uniform illumination assuming that an abrupt change in chromaticity is caused by a change in reflectance properties. This implies that the illuminant smoothly varies across the image and does not change between adjacent or nearby locations. 
Ebner \cite{LSAC} proposed a method that assumes that the illuminant transition is smooth. The method uses the local space average color for local estimation of the illuminant by convolving the image with a Gaussian kernel function. 
Bleier et al. \cite{Bleier} investigated whether existing color constancy methods, originally developed assuming uniform illumination, can be adapted to local illuminant color estimation using image sub-regions. Multiple independent estimations are then combined through regression to obtain a more robust final estimate. 
Gijsenij et al. \cite{MLS} proposed a method that makes use of local image patches, which can be selected by any sampling method. After sampling of the patches, illuminant estimation techniques are applied to obtain local illuminant estimates, and these estimates are combined into more robust estimations, since it is assumed that the number of different lights is less than the number of patches. This combination of local estimates is done with two different approaches: clustering if the number of lights is known, segmentation otherwise.
Recently Bianco and Schettini \cite{bianco2014adaptive}, and Joze and Drew \cite{joze2014exemplar} respectively extended the face-based and exemplar-based color constancy algorithms to deal with multiple illuminations.
A different class of algorithms is based on user guidance to deal with the case of two \cite{hsu2008light} and multiple lights \cite{boyadzhiev2012user}.


\section{{The proposed approach}}
\label{sec:method}

In the last years deep learning techniques allowed to obtain
significant improvements in the solution of several computer vision
problems.  Their success often depends on the availability of a large
amount of annotated training data.  Compared to other image-related
problems, in illuminant estimation annotated data is scarce.
Therefore, the straightforward procedure of learning the most probable
illuminant color directly from the image pixels needs some major
adjustments.

We propose a three-stage method: the first stage is patch based, that
is, a CNN is trained to predict the illuminant color from a small
square portion of the input image.  A large training set of patches
can be obtained even from a relatively small data set of images, making it
possible the use of deep learning techniques.  This first stage allows
to obtain multiple local estimates of the illuminant across the input
image.

The second stage determines whether or not there are multiple
illuminants in the scene.  This decision is taken on the basis of a
statistical analysis of the local estimates produced by the first
stage.  When multiple illuminants are detected, the local estimates
can be directly used as the final output of the whole method.

\begin{figure*}
  \centering
  \includegraphics[width=\textwidth]{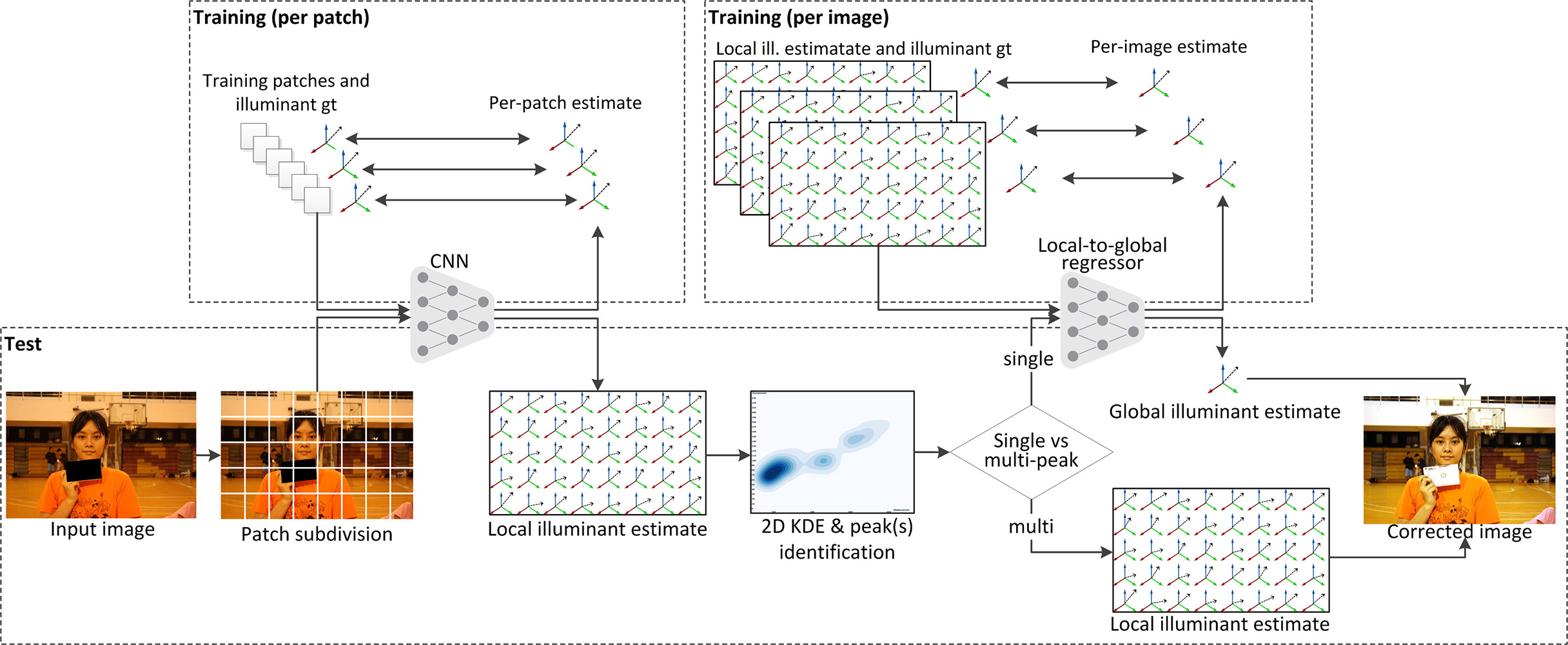}
  \caption{Scheme of the proposed illuminant estimation method showing
    the connections among the modules during both training and
    test phases.}
  \label{fig:sistemacompleto}
\end{figure*}

The optional third stage is applied when the second one determines
that the scene has been taken under a single illuminant.  In this case
it is better to aggregate the local estimates into a single
prediction.  For this purpose, in our previous work \cite{bianco2015color} we experimented
with the mean and the per-channel median operators.  In this work we
propose a local-to-global aggregation procedure based on supervised
learning.  More in detail, statistical features are extracted from the
local estimates, and then fed to a non-linear mapping whose output is
the final global estimate of the color of the illuminant.  Differently
from the first stage, this stage is image based.  Therefore, its
complexity is limited by the small number of annotated images.  For
this reason, instead of using a deep learning approach, we adopted a
``shallow'' non-linear regression scheme.
Figure~\ref{fig:sistemacompleto} shows a schematic view of the
proposed method.

\subsection{Local illuminant estimation}

\begin{figure*}
  \centering
  \includegraphics[width=1.45\columnwidth]{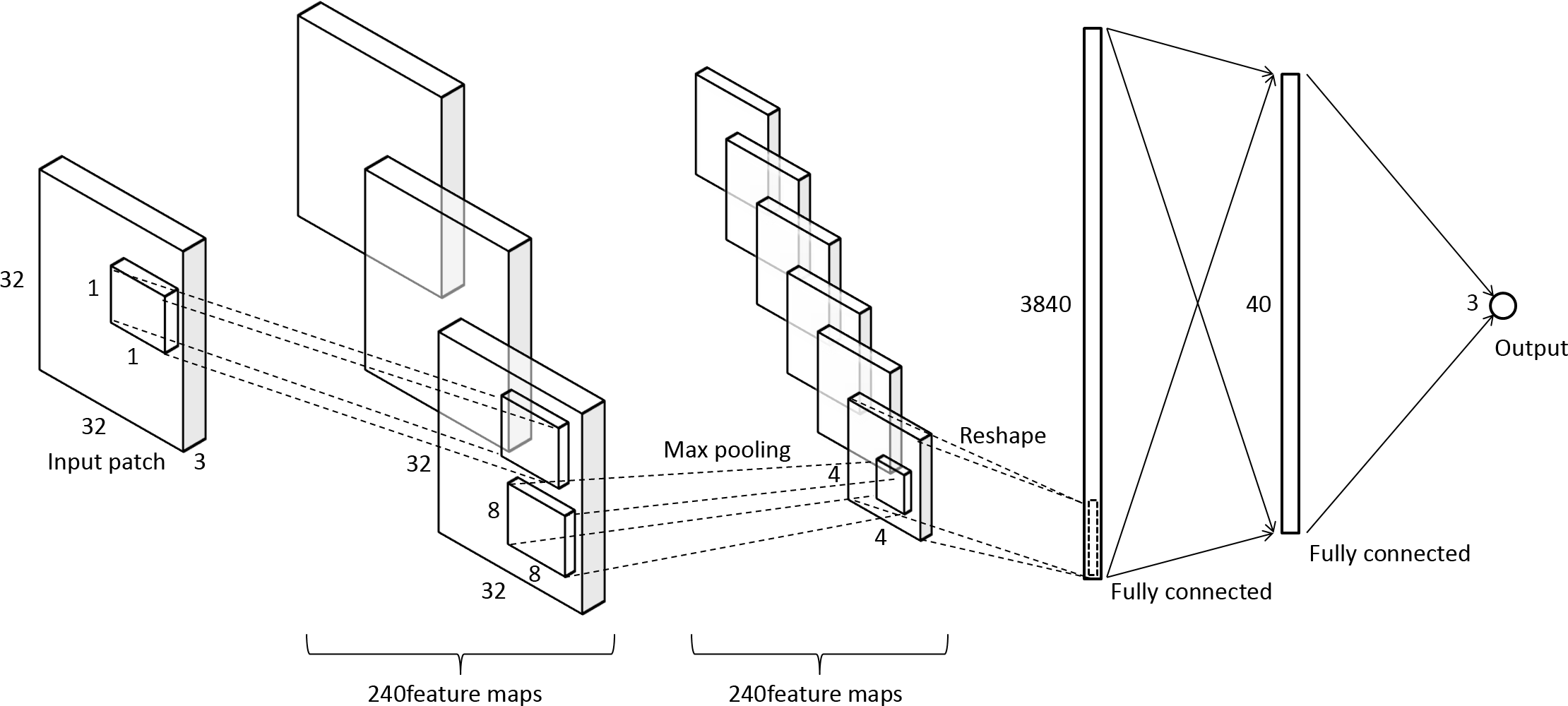}
  \caption{The architecture of the CNN that produces the
    local estimates.}
  \label{fig:architecture}
\end{figure*}

In the first stage a convolutional neural network produces local
estimates of the illuminant. The network, described in greater detail
in Section \ref{sec:netarch}, takes as input non-overlapping patches that
have been previously subjected to a stretching of the histogram so
that the output estimate is invariant with respect to the local
contrast.  
The network 
is composed by the the following sequence of layers
(see also Figure~\ref{fig:architecture} for a graphical
representation):
\begin{itemize}
\item input RGB patches of size $32 \times 32 \times 3$;
\item a bank of 240 convolutional $1 \times 1 \times 3$ filters
  producing an output of size $32 \times 32 \times 240$;
\item downsampling via an $8 \times 8$ max pooling layer to a size of
  $4 \times 4 \times 240$;
\item reshaping of the result of pooling into a 3840-dimensional
  vector;
\item a linear $3840 \times 40$ layer producing a 40-dimensional
  feature vector;
\item a ReLU activation function;
\item a linear $40 \times 3$ layer producing the output RGB estimate.
\end{itemize}

Taking into account all the linear coefficients and the biases, the
network include a total of 154,723 parameters that have been learned
by applying the standard back propagation algorithm to minimize the
average Euclidean squared difference between the estimated and the
ground truth illuminant colors (we also tried to minimize the cosine
loss without any improvement).  
Beside its size, compared to the networks used for scene and object
recognition we notice two major differences: (i) $1 \times 1$
convolutional filters, 
and (ii) the large $8 \times 8$ pooling.  
These differences can be motivated by considering
that with respect to object/scene recognition, illuminant estimation
is a dual problem: instead of trying to identify the content of the
image regardless the illuminant, here we need to estimate the
illuminant regardless the content of the image.  
\ADD{A detailed interpretation of the model from a color constancy point of view is given in Section \ref{sec:netarch1}.}

\subsection{Detection of multiple illuminants}
Since our CNN is applied to each patch independently, it can be easily
used to predict local illuminants.  However, local estimates tend to
be noisy and sometimes (when there is a single illuminant, or when the
color of all the light sources is very similar) it is better to
replace them with a single global estimate.  What we need is an
automatic rule to switch between the two modalities.  In order to
decide if the image contains single or multiple illuminants, the per
patch illuminant estimates are normalized and projected onto the
normalized chromaticity plane.  Then, an efficient 2D kernel density
estimation (KDE) \cite{botev2010kernel} is applied. The modes
$(R_i, B_i)$, $i=1,\ldots,n$, i.e.~the red/blue chromaticities (the
green channel is scaled to one) with the highest densities are
identified using a scale-space filtering \cite{SSHF}. Only the modes
with a value higher than $t$ times the maximum are retained:
\begin{equation}
  J=\left\{ j \in \{ 1, \dots, m \} : \frac{density(R_j,B_j)}{\max_{i=1,\ldots,n} density(R_i,B_i)} \geq t \right\}.
  \label{eq:ssubset}
\end{equation}
The angular difference between each pair of the retained modes
($(R_j, 1, B_j)$, $j \in J$) is computed.  If the maximum difference
exceeds a set threshold then the scene is considered as taken under
multiple illuminants.  Otherwise, we proceed by assuming the
presence of a single illuminant.  Following
\cite{Hor06,bianco2014adaptive} we set the threshold to 3$^\circ$,
since it has been judged to be a noticeable but acceptable
difference.

\subsection{Local to global aggregation of the estimates}
In our previous work \cite{bianco2015color} we generated a single illuminant estimation per image by pooling the predicted illuminants on the image patches. By taking image patches as input, we have a much larger number of training samples compared to using the whole image on a given dataset, which particularly meets the needs of
CNNs, but we loose the information that certain patches belong to the same image. Thus, we fine-tuned the learned net by adding knowledge about the way local
estimates are pooled to generate a single global estimate for each image. 

\begin{figure}
  \centering
	\includegraphics[width=\columnwidth]{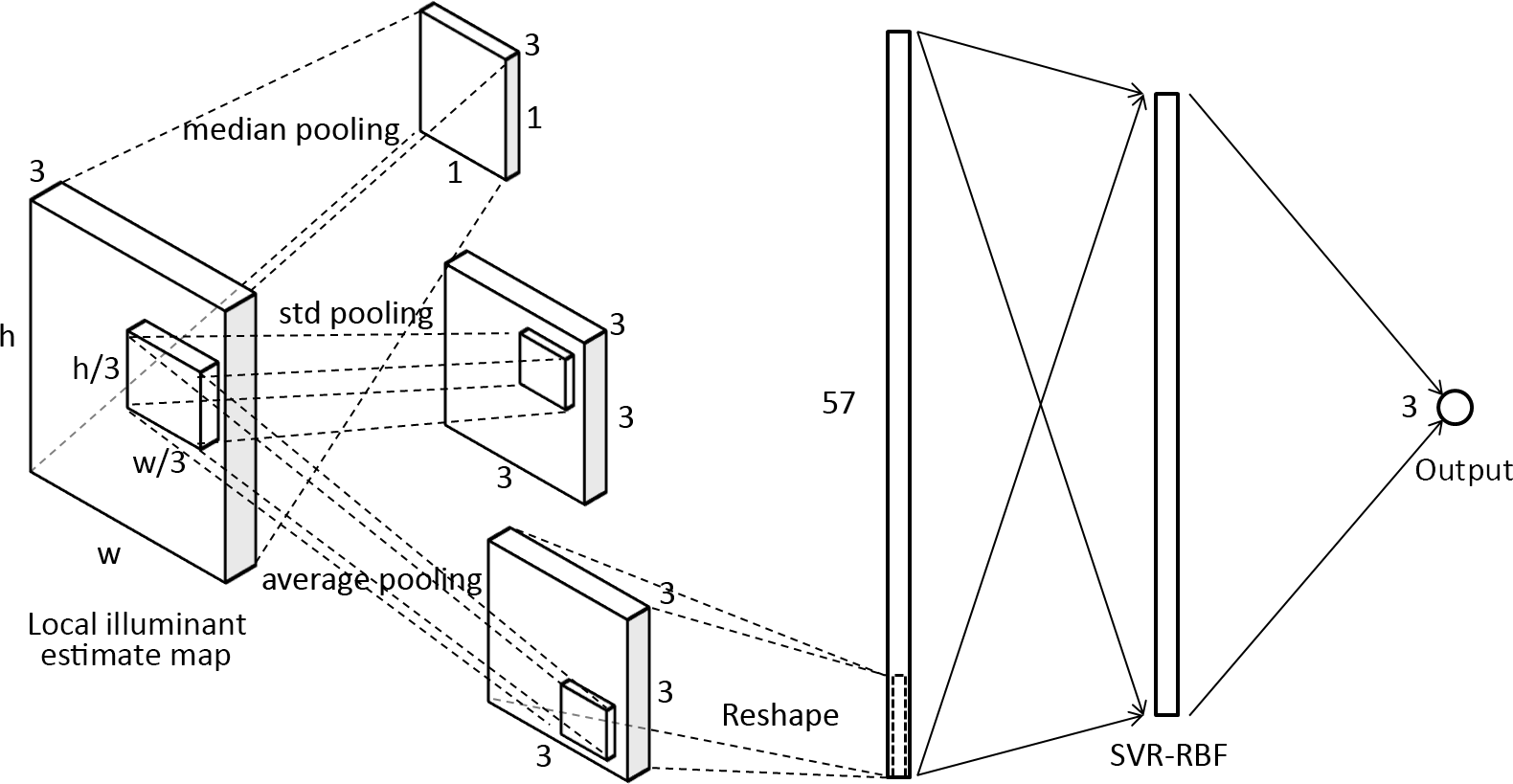}
  \caption{The architecture of our local-to-global \regressor.}
  \label{fig:architecture2}
\end{figure}

In this work we extend the per-channel average and median pooling operators used in \cite{bianco2015color} with a non-linear mapping based on a RBF kernel over local statistics
of the CNN estimates. The parameters of the mapping are obtained by applying a regression procedure that minimizes the median angular error on the training set.
%
%
Given as input the map of the per-patch illuminant estimates having a
size of $w \times h$, the first step in this module is the smoothing
via convolution with a $5 \times 5$ Gaussian filter. The response is then
independently pooled in three different ways: average pooling and
standard deviation pooling both with size $w/3 \times h/3$ (i.e.~on a
subdivision in nine rectangular regions), and median pooling with size
$w \times h$ (i.e.~on the whole image). These values are reshaped and
given as input to a SVR (with RBF kernel) which predicts the global
illuminant by minimizing the median angular error over the training
set. The architecture of this module is reported in
Figure~\ref{fig:architecture2}.
%

\begin{figure*}
  \centering
	\resizebox{\textwidth}{!}{
	\setlength{\tabcolsep}{2pt}
	\begin{tabular}{cccccc}
	\includegraphics[height=0.5\columnwidth]{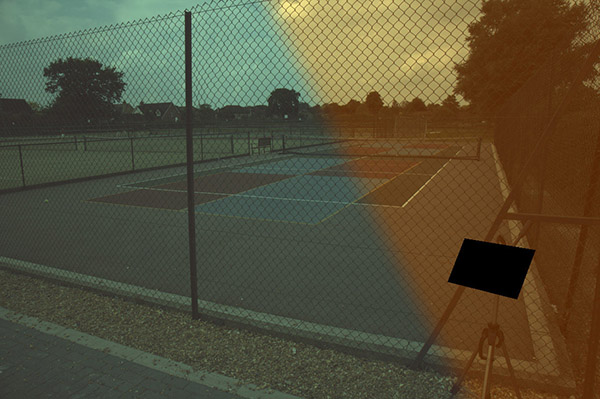}	&
	\includegraphics[height=0.5\columnwidth]{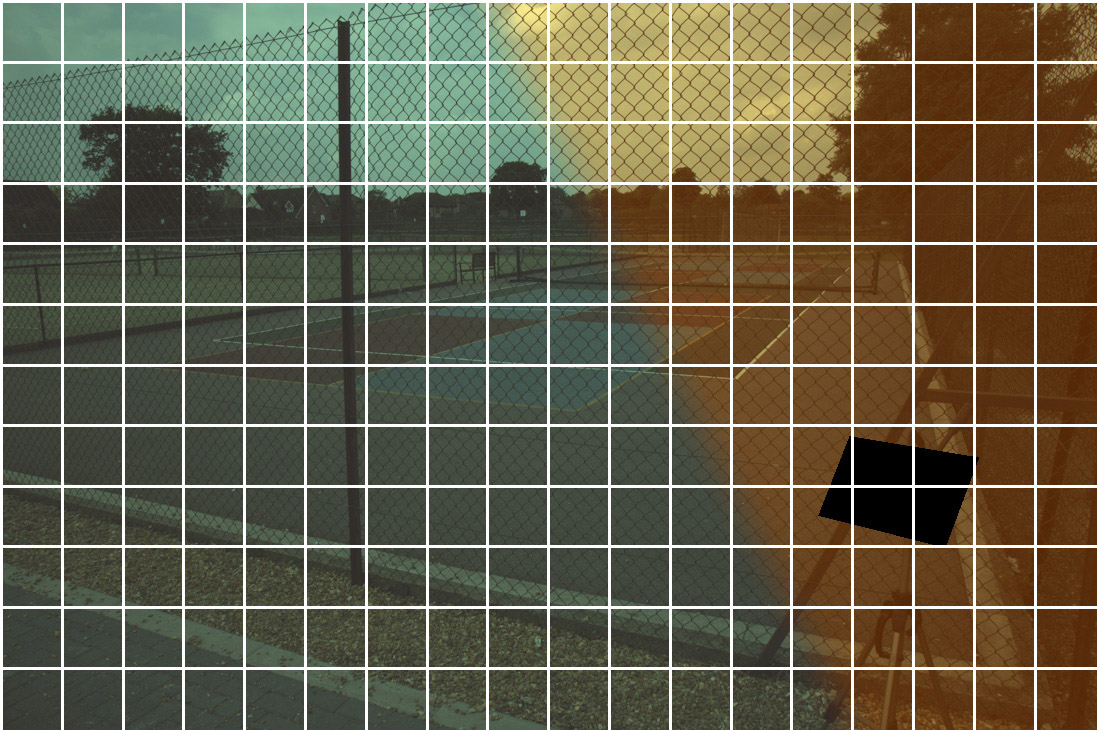} &
	\includegraphics[height=0.5\columnwidth]{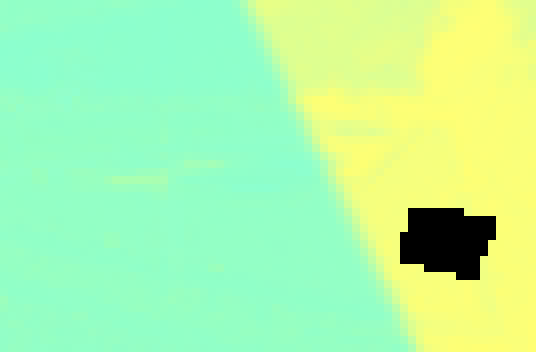} & 
	\includegraphics[height=0.5\columnwidth]{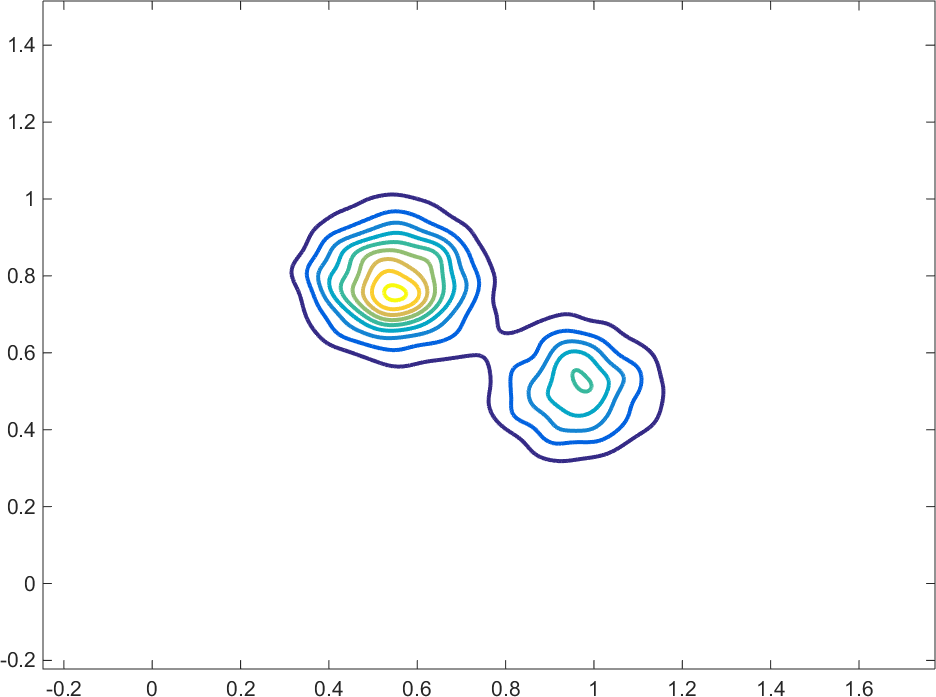} & 
	\includegraphics[height=0.5\columnwidth]{imesempiopipe_multi_stima.png} & 
	\includegraphics[height=0.5\columnwidth]{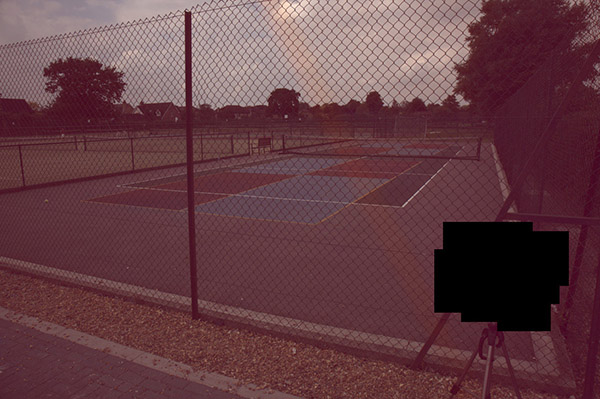} \\
	\includegraphics[height=0.5\columnwidth]{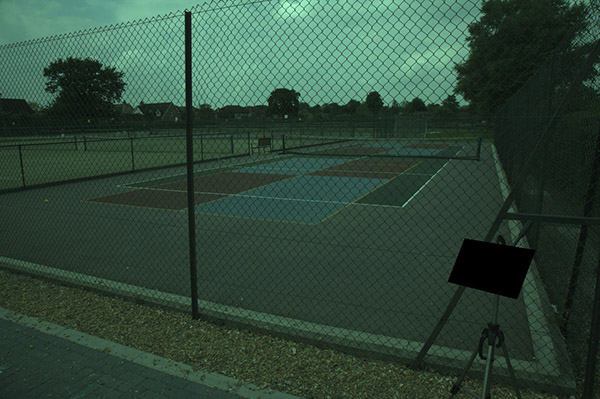}	&
	\includegraphics[height=0.5\columnwidth]{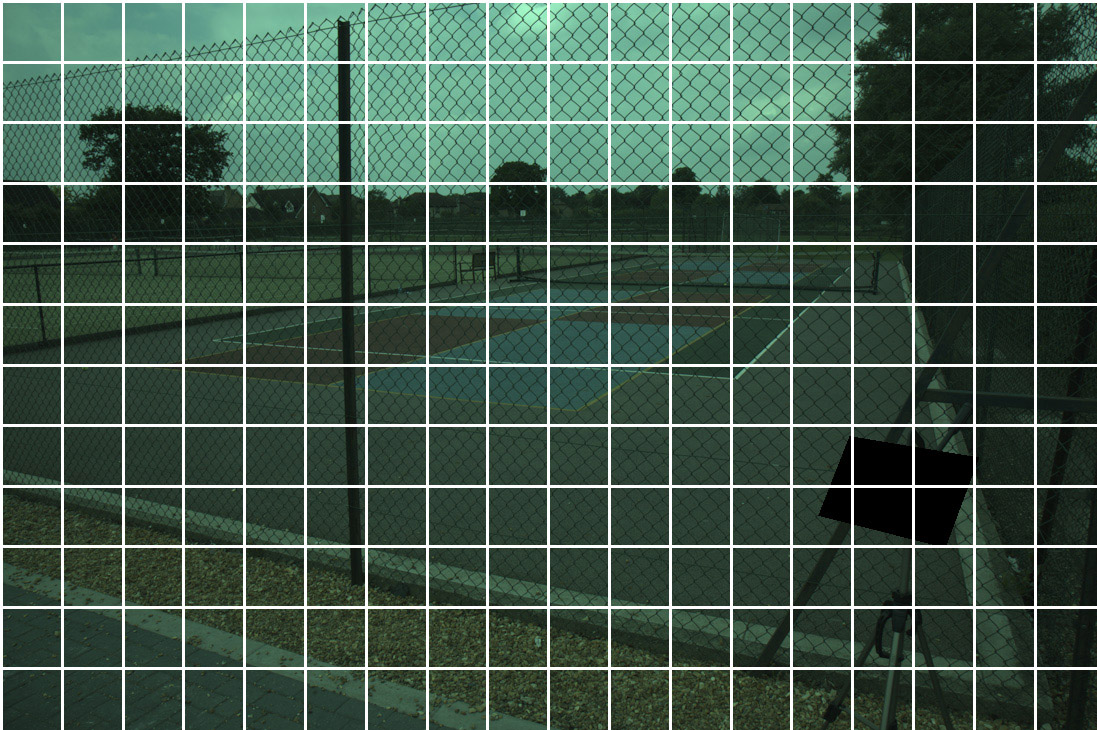} &
	\includegraphics[height=0.5\columnwidth]{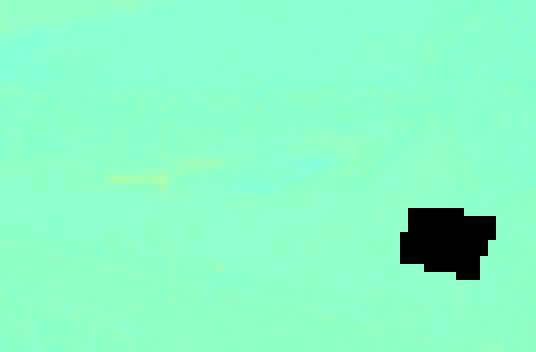} & 
	\includegraphics[height=0.5\columnwidth]{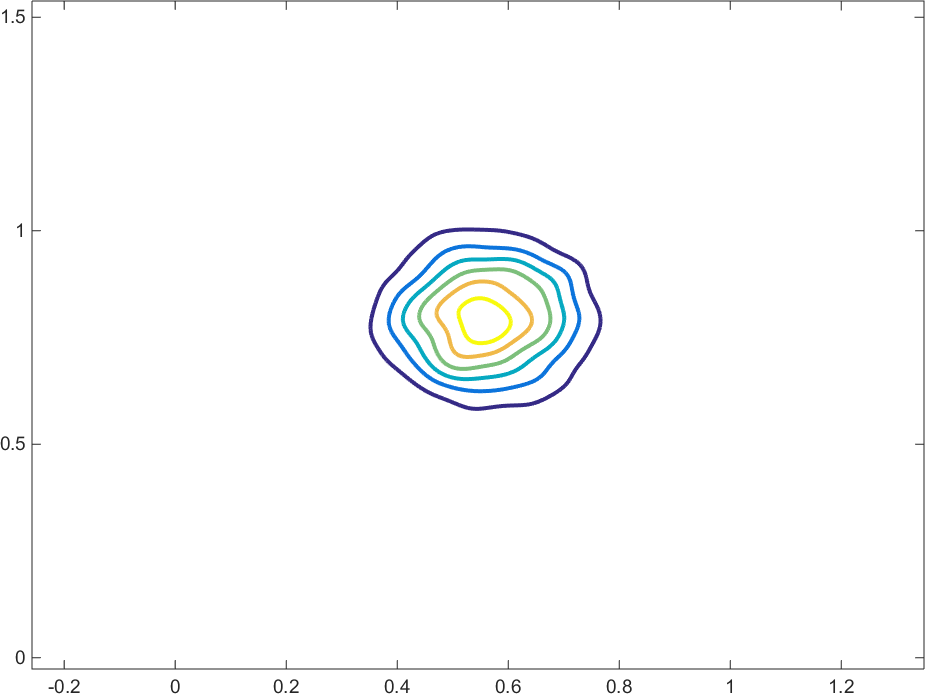} & 
	\includegraphics[height=0.5\columnwidth]{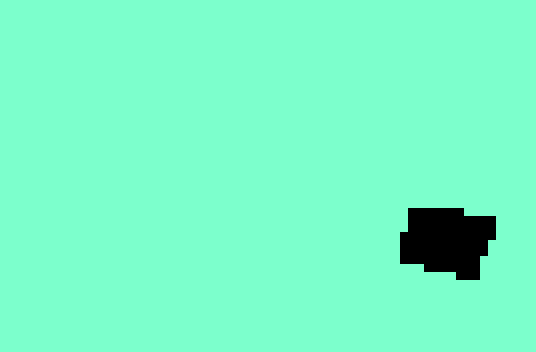} & 
	\includegraphics[height=0.5\columnwidth]{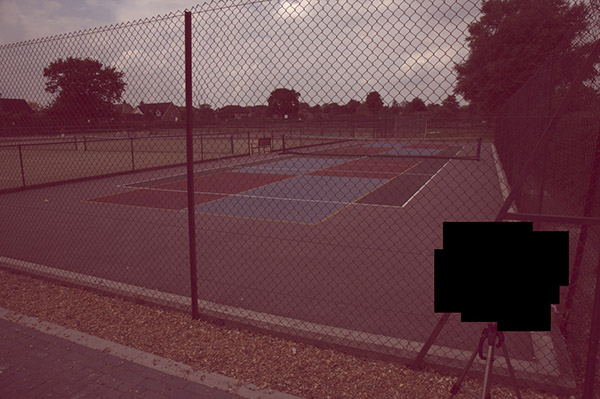} \\
	\LARGE Input image & 
	\LARGE Patch subdivision & 
	\LARGE Local illuminant estimate & 
	\LARGE KDE & 
	\LARGE Final illuminant estimate & 
	\LARGE Corrected image \\
	\end{tabular}
	}
  \caption{Output of each stage of the proposed illuminant estimation method is showed in the case of multiple (top row) and single illuminants (bottom row). From left to right: input image, subdivision in patches, local illuminant estimate, output of KDE where it is possible to see the peaks of the different illuminants found: two for the first image and one for the second one; final illuminant estimate: local illuminant estimate for the first image (since the number of peaks found by KDE is greater than one), global illuminant estimate for the second one (since only one peak is found by KDE); corrected images.}
  \label{fig:architectureExample}
\end{figure*}

In Figure~\ref{fig:architectureExample} the output of each stage of the proposed illuminant estimation method is showed in the case of multiple and single illuminants. 


\section{Experimental Setup}
\label{sec:setup}
The aim of this section is to investigate if the proposed algorithm
can outperform state-of-the-art algorithms in the single and multiple
illuminant estimation on standard datasets of RAW images. 

\subsection{Image Datasets and Evaluation Procedure}
To test the performance of the proposed algorithm for the global
illuminant estimation, \ADD{two} standard datasets of RAW camera images having
a known color target are used. In the first dataset, images have been captured using
high-quality digital SLR cameras in RAW format, and are therefore free
of any color correction.  The dataset \cite{cambridge} was originally
available in sRGB-format, but Shi and Funt \cite{Shi} reprocessed the
raw data to obtain linear images with a higher dynamic range (14 bits
as opposed to standard 8 bits).  The dataset has been acquired using a
Canon 5D and a Canon 1D DSLR cameras and consists of a total of 568
images. 
The Macbeth ColorChecker (MCC) chart is included in every scene, and
this allows to accurately estimate the actual illuminant of each
acquired image.  
\ADD{The second dataset is the NUS dataset \cite{cheng2014illuminant}. The dataset is similar to the previous one: it has been captured using digital SLR cameras in RAW format with a MCC included in every scene. The differences with the previous dataset are that it has been captured by 9 different cameras (Canon 1Ds Mk III, Canon 600D, Fujifilm X-M1, Nikon D5200, Olympus E-PL6, Panasonic Lumix DMC-GX1, Samsung NX2000, Sony SLT-A57 and Nikon D40) and that there is a larger number of images, i.e. 1853 with around 200 images for each camera.} 

To test the performance of the proposed algorithm for the multiple illuminant estimation, \ADD{three} different datasets have been used. The first one is synthetically generated from the Gehler-Shi dataset: each image is relighted using two, three and four random illuminants taken from the same datasets. 
This synthetic dataset thus contains a total of 1704 images. 
The second dataset used is a subset of the Milan portrait dataset \cite{bianco2014adaptive}. It has been acquired in RAW format using four different DSLR cameras: Canon 40D, Canon 350D, Canon 400D, and Nikon D700. The dataset is the union of different subsets that have been acquired in three different world locations: Italy, Taiwan, and Japan. The dataset includes portraits of a single person with a single MCC up to multiple persons with multiple MCCs. In this work we used the subset containing multiple MCCs, for a total of 197 images. 
\ADD{Finally, the third one is the multiple illuminant dataset by Beigpour et al. \cite{beigpour2014multi}. It has been acquired using a Sigma SD10 single-lens reflex (SLR) digital camera which uses a Foveon X3 sensor and is available in linear RAW format. The dataset consist of two parts: the first one is taken in controlled laboratory setting for a total of 10 scenes taken under six distinct illumination conditions; the second one is taken in uncontrolled setting for a total of 20 indoor and outdoor scenes. The datasets comes with pixel-wise ground truth information.}

\ADD{The network has been trained on the Gehler-Shi dataset and adapted to the other datasets by re-training each time the local-to-global \regressor{} to cope with the different cameras and sensor type used.}

\begin{figure}%
	\begin{tabular}{c}
	\includegraphics[width=0.95\linewidth]{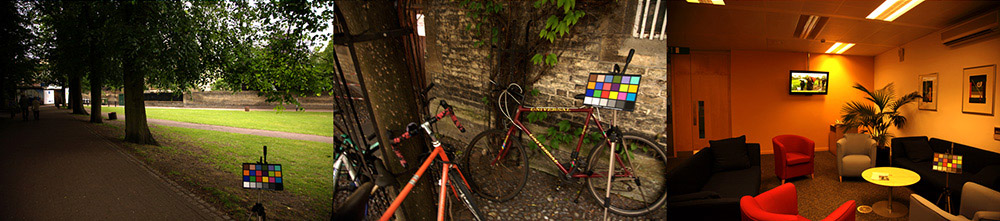} \\
	\includegraphics[width=0.95\linewidth]{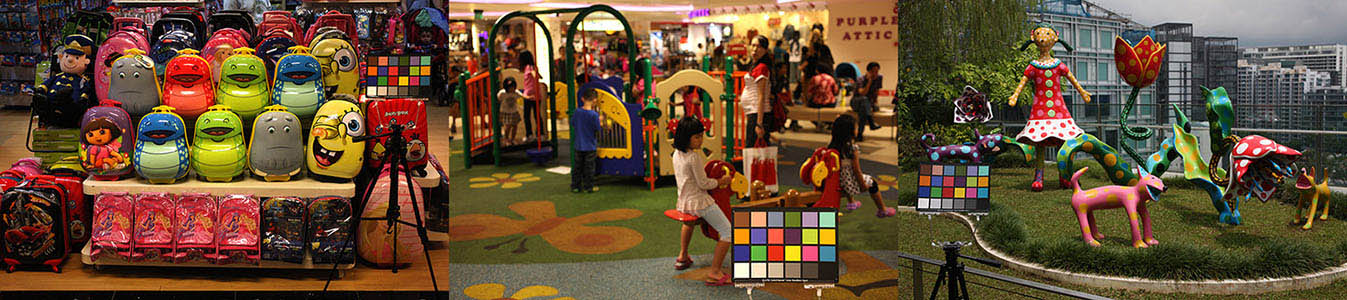} \\
	\includegraphics[width=0.95\linewidth]{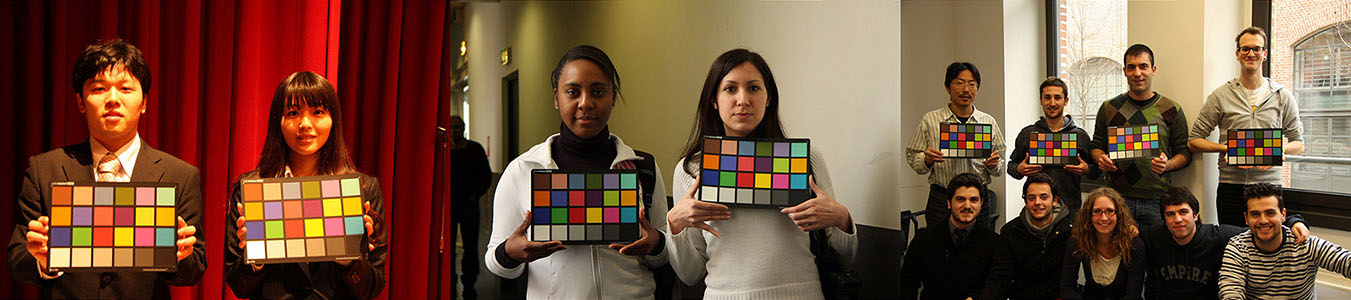} \\
	\includegraphics[width=0.95\linewidth]{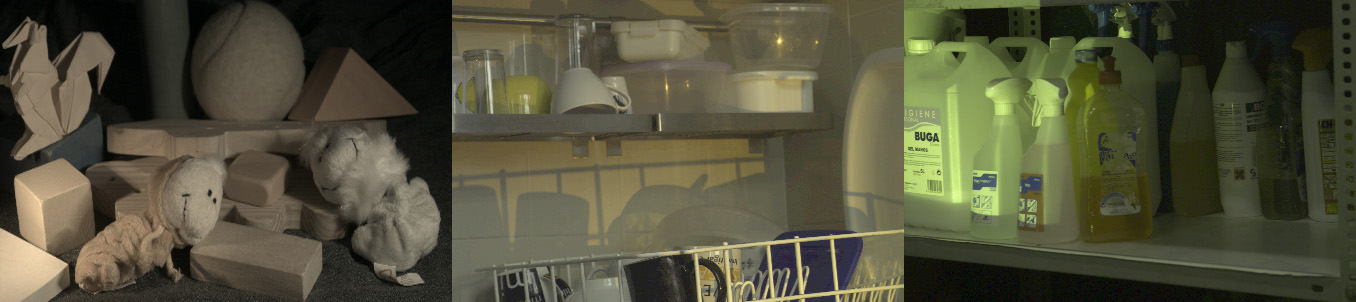} \\
	\end{tabular}
	\caption{Example of images within the image datasets considered. Top to bottom: Gehler-Shi, NUS, Milan portrait and Beigpour et al. datasets.}
	\label{fig:datasetall}
\end{figure}

\ADD{Examples of images within the datasets considered are reported in Figure \ref{fig:datasetall}.}

%
%

\subsubsection{Relighted Gehler-Shi dataset}
We synthetically generated a relighted version of the Gehler-Shi dataset: each image is balanced using the corresponding ground truth illuminant and relighted using two, three and four random illuminants taken from the original dataset. 
Their position in the image was set randomly with the constraint of being at least $\min \{w,h\}/3$ apart, with $w$ and $h$ being image width and height respectively.
The ground truth for each image has been generated by nearest-neighbor assignment followed by Gaussian smoothing to simulate illuminant mixing. 
This synthetic dataset thus contain a total of 1704 images. 
The average maximum angular distance among the illuminants in each image are 8.6$^\circ$, 12.2$^\circ$, 14.8$^\circ$ for the subsets relighted with two, three, and four illuminants respectively. 


\subsection{Benchmark algorithms}
\label{sec:bench}

Different benchmarking algorithms for color constancy are
considered. Since each image of the dataset contains only one MCC,
only global color constancy algorithms based on the assumption of
uniform illumination can be compared. Six of them are generated
varying the three variables $(n,p,\sigma)$ in
Equation~\ref{eq:general-method}, and correspond to well known and
widely used illuminant estimation algorithms.  The values chosen for
$(n,p,\sigma)$ are reported in Table \ref{tab:parametri} and set as in
\cite{ginseng11}.  The algorithms are used in the original authors'
implementation which is freely available online
(\url{http://lear.inrialpes.fr/people/vandeweijer/code/ColorConstancy.zip}). The
seventh algorithm is the pixel-based Gamut Mapping
\cite{gamutmappingGevers}. The value chosen for $\sigma$ is also
reported in Table \ref{tab:parametri}.  The other algorithms
considered are illumination chromaticity estimation via Support Vector
Regression (SVR \cite{funt2004estimating}); the Bayesian (BAY
\cite{cambridge}); the Natural Image Statistics (NIS \cite{NIS2011});
the High Level Visual Information \cite{hilevelinfo}: bottom-up (HLVI
BU), top-down (HLVI TD), and their combination (HLVI BU\&TD); the
Spatio-Spectral statistics \cite{chzcc2011}: with Maximum Likelihood
estimation (SS ML), and with General Priors (SS GP); the Automatic
color constancy Algorithm Selection (AAS) \cite{BiancoPR} and the
Automatic Algorithm Combination (AAC) \cite{BiancoPR}; the
Exemplar-Based color constancy (EB) \cite{joze2012exemplar}; the
Face-Based (FB) color constancy algorithm \cite{bianco2012color} using
GM or SS ML when no faces are detected; the CNN-based algorithms
\cite{bianco2015color} and the AlexNet fine-tuned with a linear
Support Vector Regression (SVR) \cite{SVR} to estimate the illuminant
color for each image \cite{bianco2015color} (AlexNet+SVR); the
ensemble of regression trees applied to simple color features
\cite{cheng2015effective} (SF); the corrected-moment illuminant
estimation \cite{finlayson2013corrected} (CM); the one predicting
chromaticity from pixel luminance (PCL) \cite{chakrabarti2015color}; the one exploiting bright pixels (BP) \cite{joze2012role} and the one exploiting both bright and dark pixels (BDP) \cite{cheng2014illuminant}.


\begin{table}[!ht]
\caption{{Values chosen for $(n,p,\sigma)$ for the state-of-the-art algorithms which are instantiations of Eq.\ref{eq:general-method}.}}
\label{tab:parametri}
\centering
\begin{tabular}{lrrr}
  \toprule
  Algorithm & $n$ & $p$ &  $\sigma$ \\
  \midrule
Gray World (GW)     							&   0 &   1          &  0 \\
White Patch (WP)     							&   0 &   $\infty $  &  0 \\
Shades of Gray (SoG)    					&   0 &   4          &  0 \\
general Gray World (gGW)    		  &   0 &   9          &  9 \\
1st-order Gray Edge (GE1)    			&   1 &   1          &  6 \\
2nd-order Gray Edge (GE2)    			&   2 &   1          &  1  \\
Gamut Mapping (GM)								&   0 &   0          &  4  \\
	\bottomrule
\end{tabular}
\end{table}

The last algorithm considered {is} the Do Nothing (DN) algorithm which gives the same estimation for the color of the illuminant ($\mathbf{I}=[1 \ 1 \ 1]$) for every image, i.e. it assumes that the image is already correctly balanced.

\subsection{Learning of the main modules}
We train our CNN on $32 \times 32$ patches randomly taken from
training images of the Gehler-Shi dataset in RAW format (patches including portions of the
reference MCC are excluded from training). Images have been resized to
$\max (w,h)=1200$ pixels. The net is learned using a three-fold cross
validation on the folds provided with the dataset: for each run one is
used for training, one for validation and the remaining one for
test. For training, we assign each patch with the illuminant ground
truth associated to the image to which it belongs. At testing time, we
generate a single illuminant estimation per image by pooling the the
predicted patch illuminants.  By taking image patches as input, we
have a much larger number of training samples compared to using the
whole image on a given dataset, which particularly meets the needs of
CNNs. Net parameters have been learned using Caffe~\cite{jia2014caffe} with Euclidean loss.

The learned net is then applied to each whole image in the training
set by masking the MCC to obtain an illuminant estimation map.  The
pooled features computed from these maps are the input to our
local-to-global \regressor{} to give a single global illuminant
estimate for each image. We train our \regressor{} using the same
three-fold cross validation as before using an $\epsilon$-SVR
\cite{SVR} with RBF kernel in which we used a modified cost function
to minimize the median angular distance between illuminant estimates
and ground-truths. The \regressor{} is able to give a more accurate
global estimate than a simple average or median pooling
\cite{bianco2015color} for two reasons: (i) it is learning-based and
is able to leverage the different local estimates coming from the
patches belonging to the same image; (ii) it is trained by explicitly
minimizing the error metric using in the evaluation of illuminant
estimation methods.

\section{Results and Discussion}
\label{sec:results}

We evaluated the proposed method in both single and multiple
illuminant estimation.

\subsection{Global illuminant estimation}

In Table \ref{tab:errori} the median, the average, the 90$^{th}$-percentile, and the maximum of the
angular errors obtained by the considered state-of-the-art algorithms
and the proposed approach on the Gehler-Shi dataset are reported.
\begin{table}[!ht]
\caption{Angular error statistics obtained by the state-of-the-art algorithms considered on the Gehler-Shi dataset.}
\label{tab:errori}
\centering
\begin{tabular}{lrrrr}
  \toprule
	Algorithm &  Med &  Avg &  90$^{th}$prc & Max  \\
  \midrule
DN    												&   13.55 &  13.62 &  16.45 &  27.37  \\
GW    												&    6.30 &   6.27 &   10.12 & 24.84  \\
WP    												&    5.61 &   7.46 &  15.68 &  40.59  \\
SoG   												&    4.04 &   4.85 &   9.71&   19.93  \\
gGW   												&    3.45 &   4.60 &   9.68 &  22.21  \\
GE1  												  &    4.55 &   5.21 &   9.78 & 19.69  \\
GE2   												&    4.43 &   5.01 &   \bf{8.93} &  \bf{16.87}  \\
GM  \cite{gamutmappingGevers} &    \bf{2.28} &   \bf{4.10}  & 11.08 & 23.18  \\
\midrule
SVR		\cite{funt2004estimating}	&  6.67 & 7.99 & 14.61 & 26.08 \\
{BAY} 	 \cite{cambridge}				&    3.44 &   4.70 &  10.21 &  24.47 \\
{NIS}  \cite{NIS2011}     		  &    3.13 &   4.09 &   8.57 &  26.20 \\
{HLVI BU} \cite{hilevelinfo}		&    2.54 &   3.30 &   6.59 &  17.51 \\
{HLVI TD} \cite{hilevelinfo}		&    2.63 &   3.65 &   7.53 &  25.24 \\
{HLVI BU\&TD} \cite{hilevelinfo} &    2.47 &   3.38 &  6.97 &   25.24 \\
{SS ML} \cite{chzcc2011}       &    2.93 &   3.55 &    7.23 & 15.25 \\
{SS GP} \cite{chzcc2011}  			&    2.90 &   3.47 &   7.00 &  \bf{14.80} \\
{AAS}	\cite{BiancoPR}			      &    3.16 &   4.18 &   9.15 &  22.21 \\
{AAC}	\cite{BiancoPR}			      &    2.90 &   3.74 &   7.93 &  14.98 \\
EB 	 \cite{joze2012exemplar}	  &     2.24   & 2.77   &  \bf{5.52} &  19.44 \\
FB+GM   \cite{bianco2012color} 	&    2.01 &   3.67 &  9.50 &   23.18  \\
{FB+SS GP} \cite{bianco2012color} &   2.57 & 3.18 & 6.67 & \bf{14.80} \\
CM~\cite{finlayson2013corrected}  &  2.04 & 2.86 &  -- &  -- \\
SF~\cite{cheng2015effective}      &   {\bf 1.65} & {\bf 2.42}  &  -- & -- \\
PCL~\cite{chakrabarti2015color}   &  1.67 & 2.56 &  5.56 & -- \\
BP~\cite{joze2012role}            &  2.61 & 3.98 &  -- &  -- \\
BDP~\cite{cheng2014illuminant} & 2.14 & 3.52 &  -- & 28.35 \\
\midrule
AlexNet+SVR \cite{bianco2015color}		     &  3.09  & 4.74 &  11.18 & 29.15\\ 
CNN per patch	\cite{bianco2015color}			 &   2.69  & 3.67 &  7.79 & 30.93 \\
CNN	average-pooling	\cite{bianco2015color} &  2.44  & 3.18 &   6.37 & 14.84 \\ 
CNN median-pooling \cite{bianco2015color}	 &  2.32  & 3.07 &   6.15 & 19.04 \\
CNN fine-tuned \cite{bianco2015color}	 		 &    \bf{1.98}  & \bf{2.63} & \bf{5.54} &  \bf{14.77} \\ 
\midrule
Proposed single estimate &  1.44 &	2.36	& 5.72	& 16.98 \\
\bottomrule
\end{tabular}
\end{table} 
%
The table is divided into three blocks and for each of them the best result for each statistic is reported in bold. The first block includes statistic-based algorithms, the second one learning-based algorithms, and the third one the different variants of the proposed approach.  

From the results it is possible to see that the deep CNN pre-trained
on ILSVRC 2012 \cite{krizhevsky2012imagenet} coupled with SVR
(i.e. AlexNet+SVR) is already able to outperform most statistic-based
algorithms and some learning-based ones.  The CNN introduced in our
previous work~\cite{bianco2015color} in its various instantiations
allowed to obtain a median angular error below 2 degrees which is
better than almost all the other methods considered.  Even better
results have been obtained with the recent method by Cheng et
al.~\cite{chzcc2011} for which the median error is 1.65 degrees.  The
method proposed here obtained the lowest error (1.44 degrees if we
consider the median).  The ranking of the algorithms does not change
if we consider the mean error instead of the median; the best maximum
error, instead has been obtained by the fine-tuned
CNN~\cite{bianco2015color}).

\begin{figure}
  \centering
  \includegraphics[width=0.99\columnwidth]{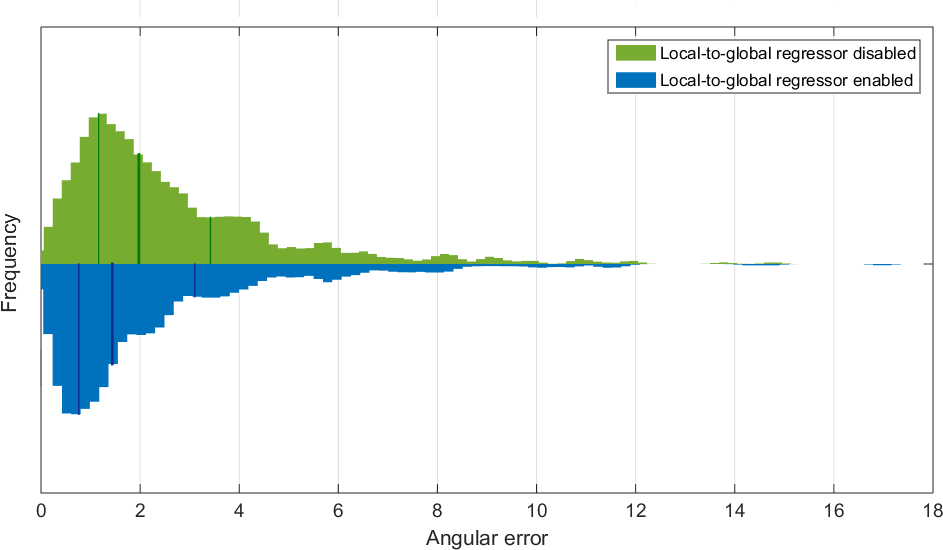} 
  \caption{Distribution of the angular errors obtained on the
    Gehler-Shi dataset with (bottom) and without (top) the
    local-to-global aggregation module.  On each distribution, the
    black lines indicate the quartiles.}
  \label{fig:errorDistribution}
\end{figure}

Note that for this experiment we did not apply the multiple illuminant
detection module and we always performed the local-to-global
aggregation.  This last step brings a significant improvement.  In
fact, without it the median error raises by more than one degree,
reaching the 2.69 degrees corresponding to the ``CNN per patch''
result.  It is also a significant improvement with respect to the
other aggregation methods considered in our previous work: average
pooling, median pooling and fine tuning, that obtained median errors
of 2.44, 2.32 and 1.98, respectively.
Figure~\ref{fig:errorDistribution} shows the distribution of the
angular errors obtained with and without the local-to-global
\regressor. It is possible to see how the introduction of the
aggregation module pushes the angular error distribution towards zero.

\begin{figure*}
	\centering
	\resizebox{0.95\textwidth}{!}{
	\setlength{\tabcolsep}{2pt}
	\begin{tabular}{cccc}
	\includegraphics[width=0.40\columnwidth]{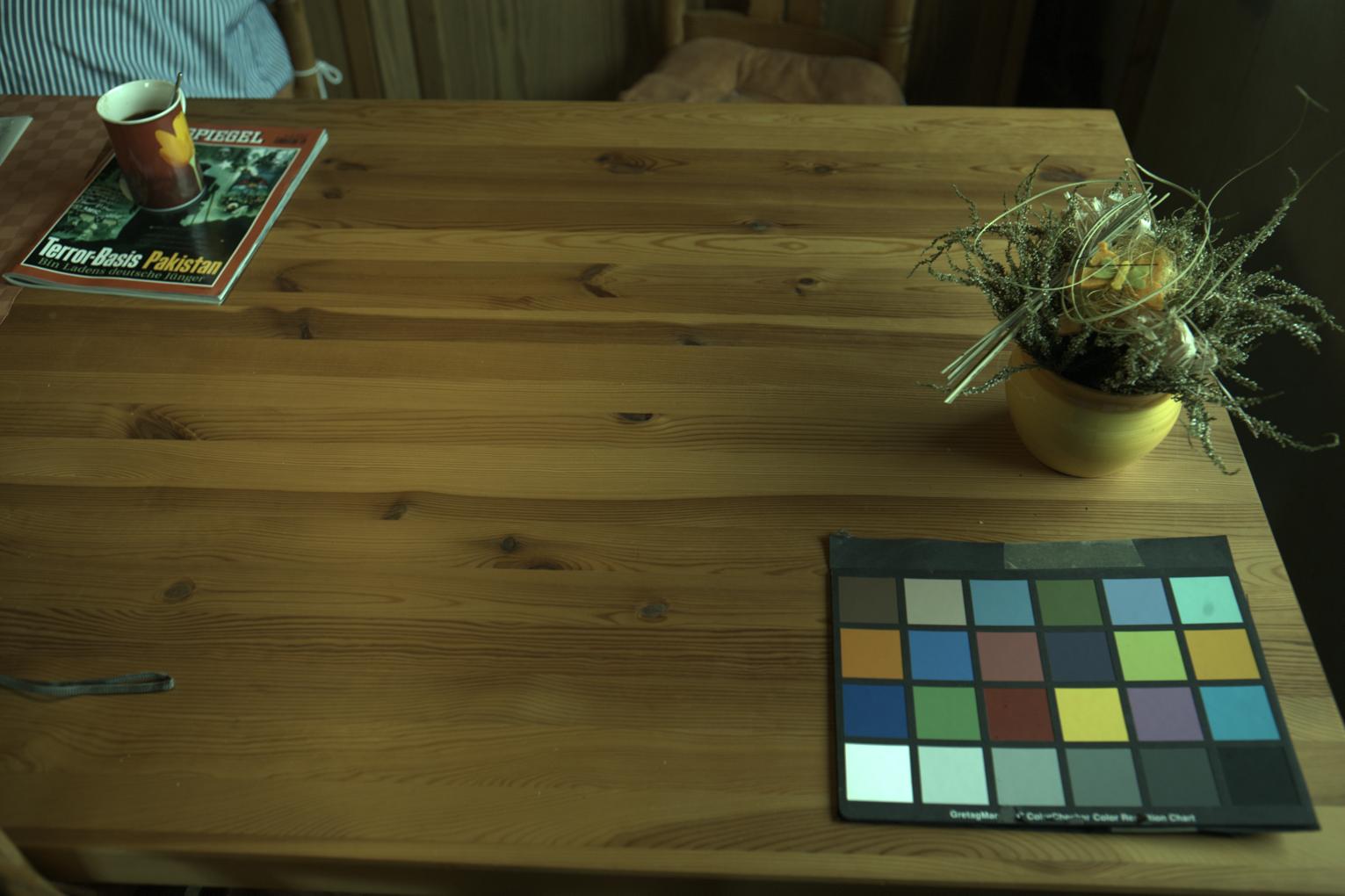} &
	\includegraphics[width=0.40\columnwidth]{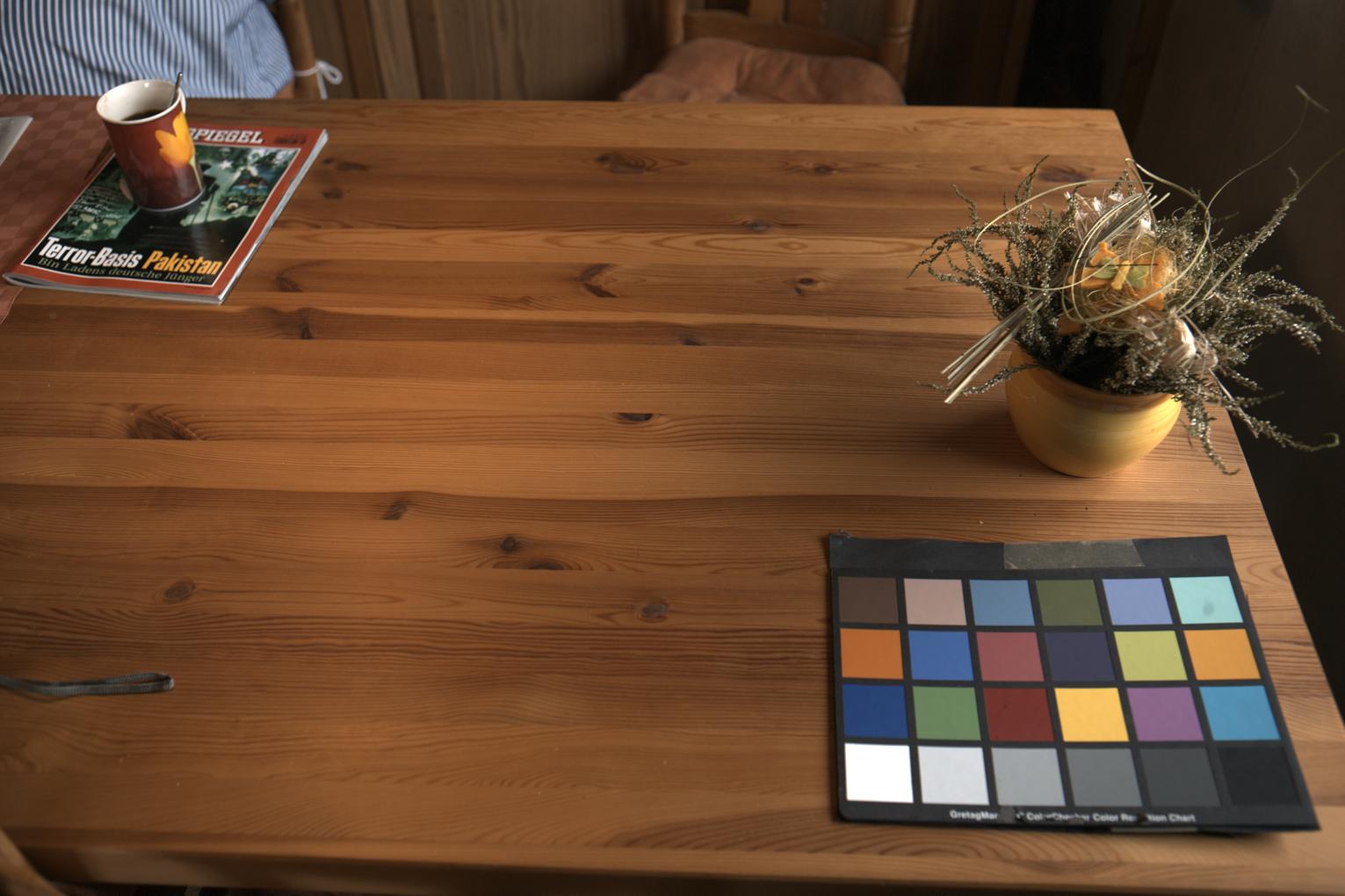} &
	\includegraphics[width=0.40\columnwidth]{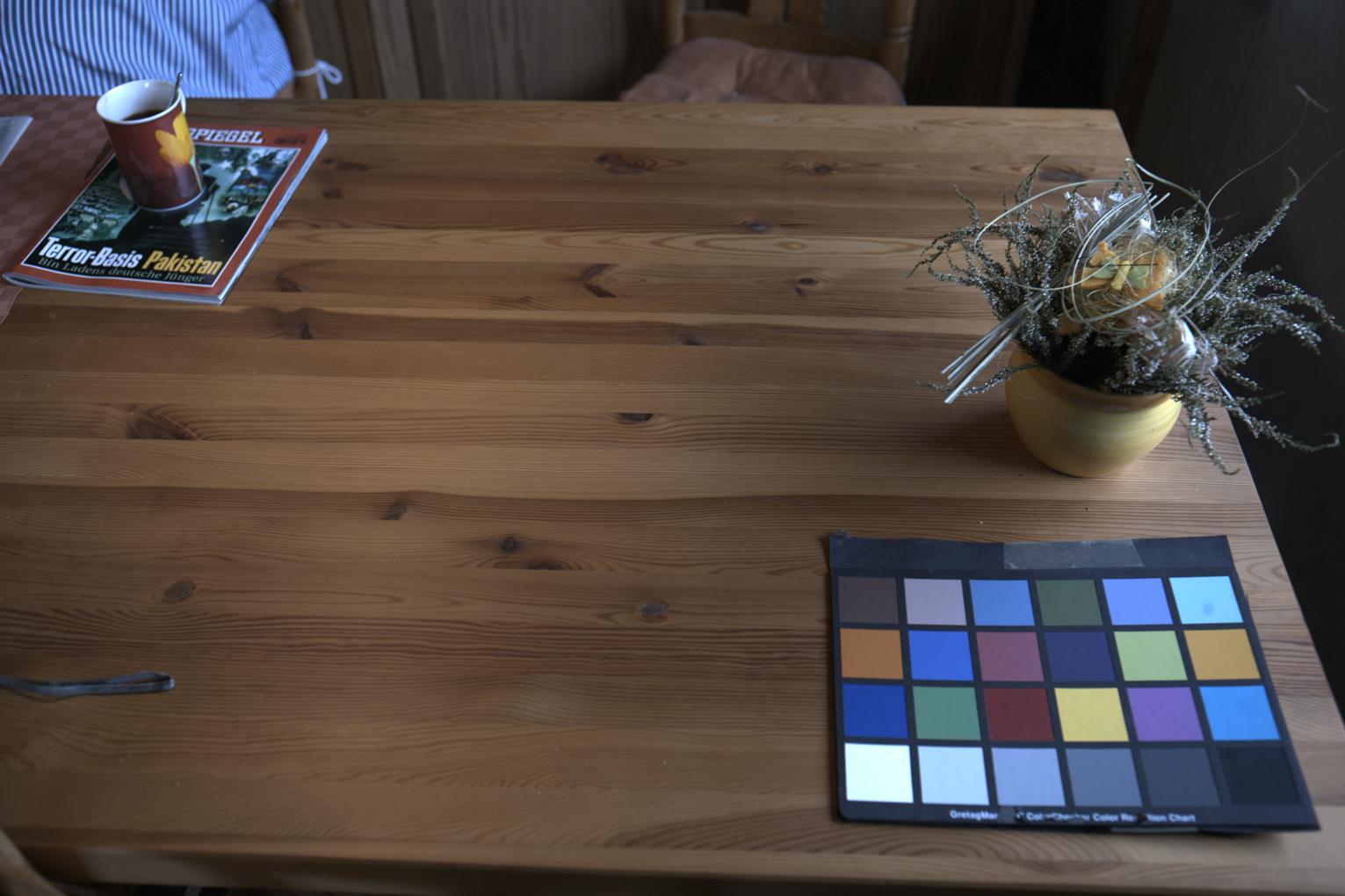} &
	\includegraphics[width=0.40\columnwidth]{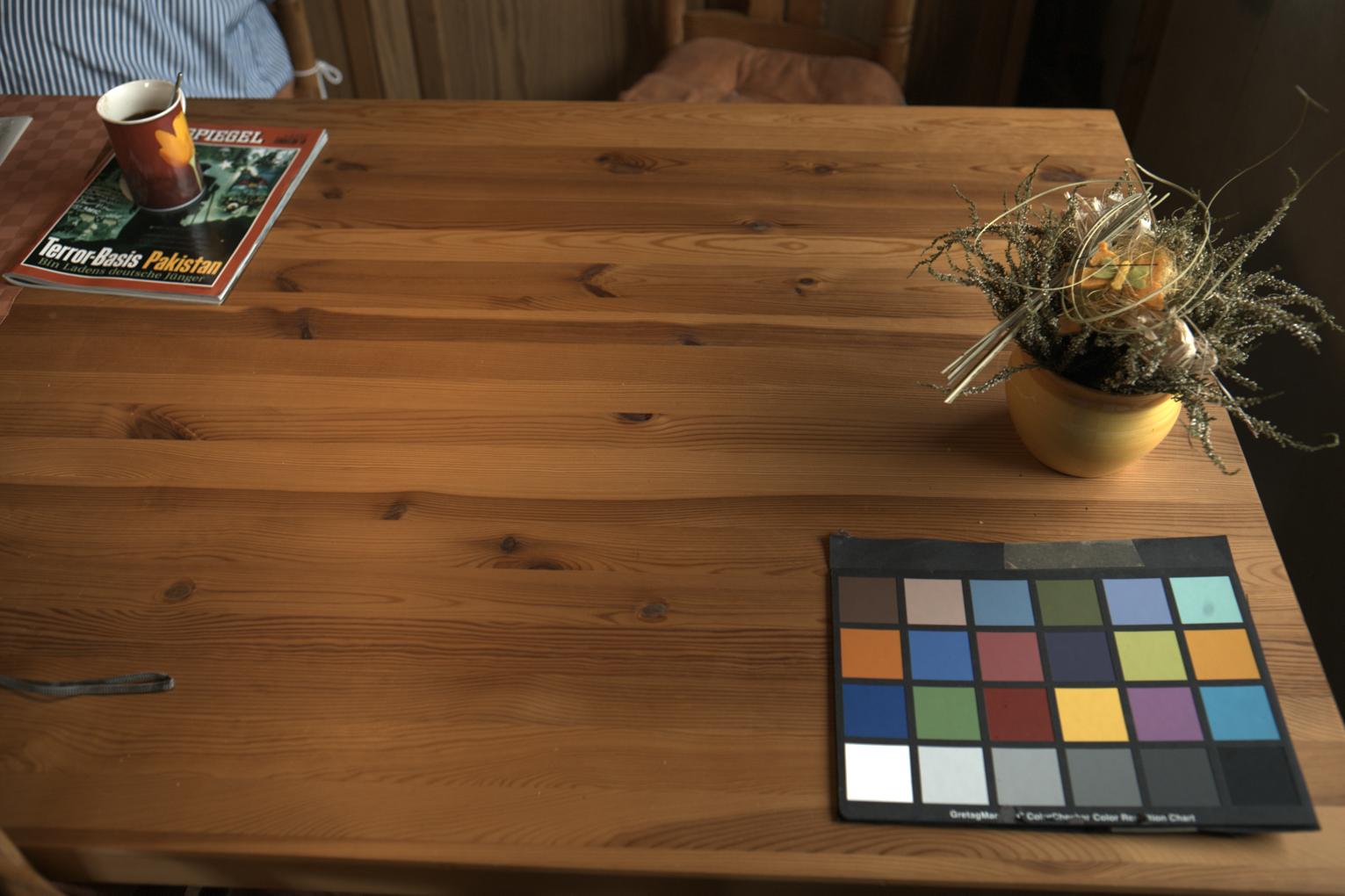} \\
	\small Input image & \small Ground truth ($0^\circ$) & \small Proposed (16.98$^\circ$) & \small AAS (1.48$^\circ$) \\
	\includegraphics[height=0.40\columnwidth,angle=90]{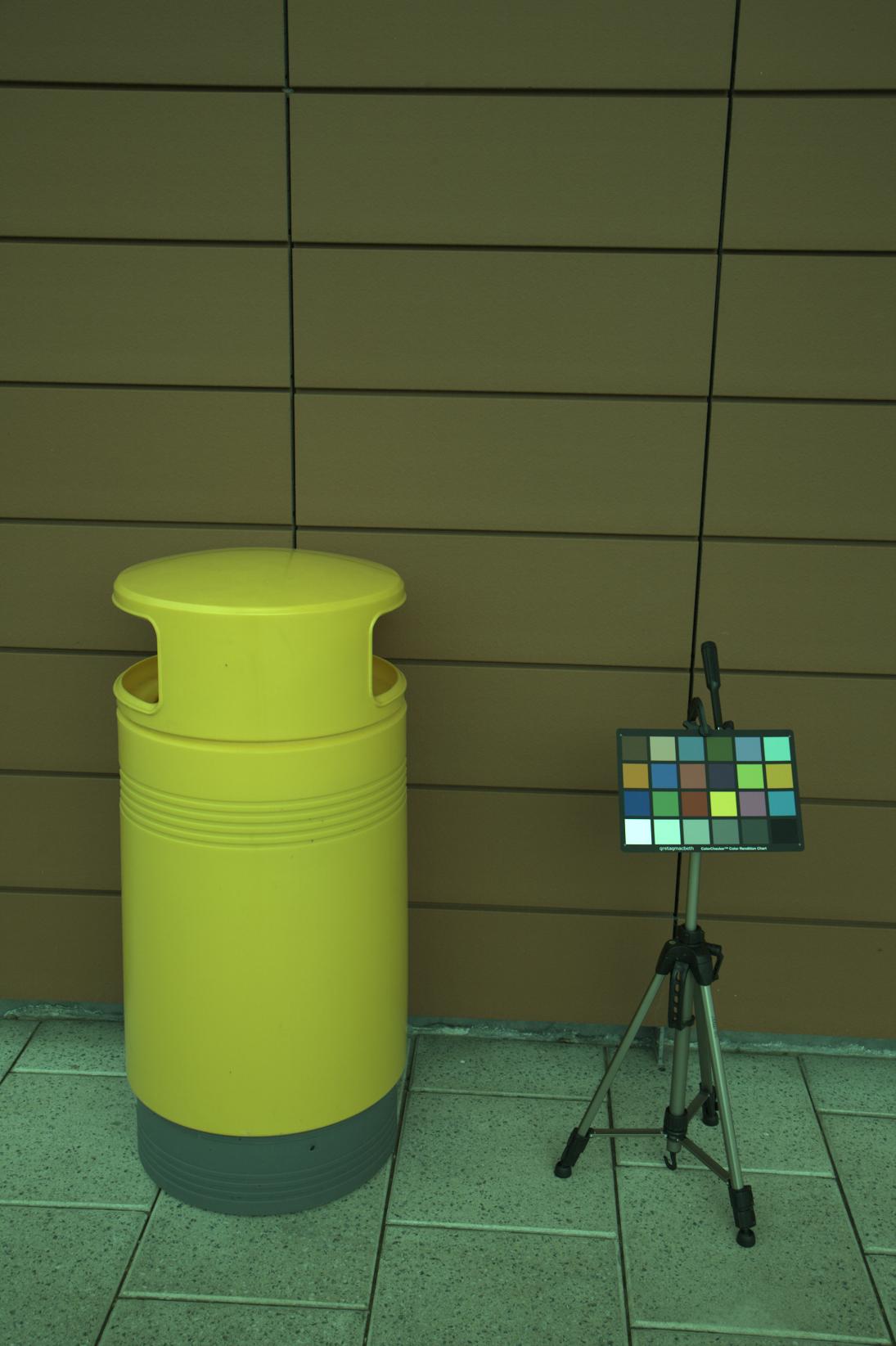} &
	\includegraphics[height=0.40\columnwidth,angle=90]{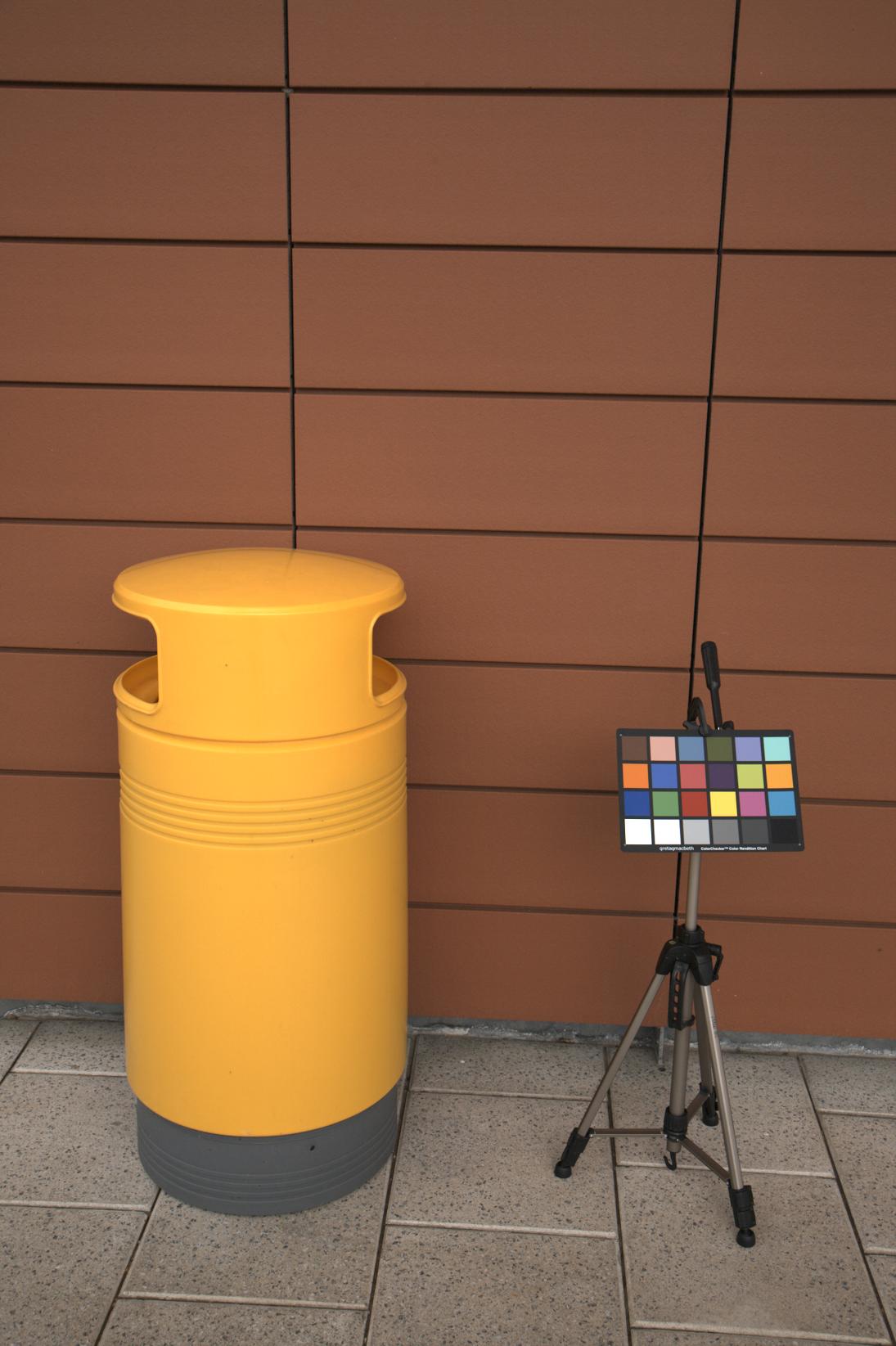} &
	\includegraphics[height=0.40\columnwidth,angle=90]{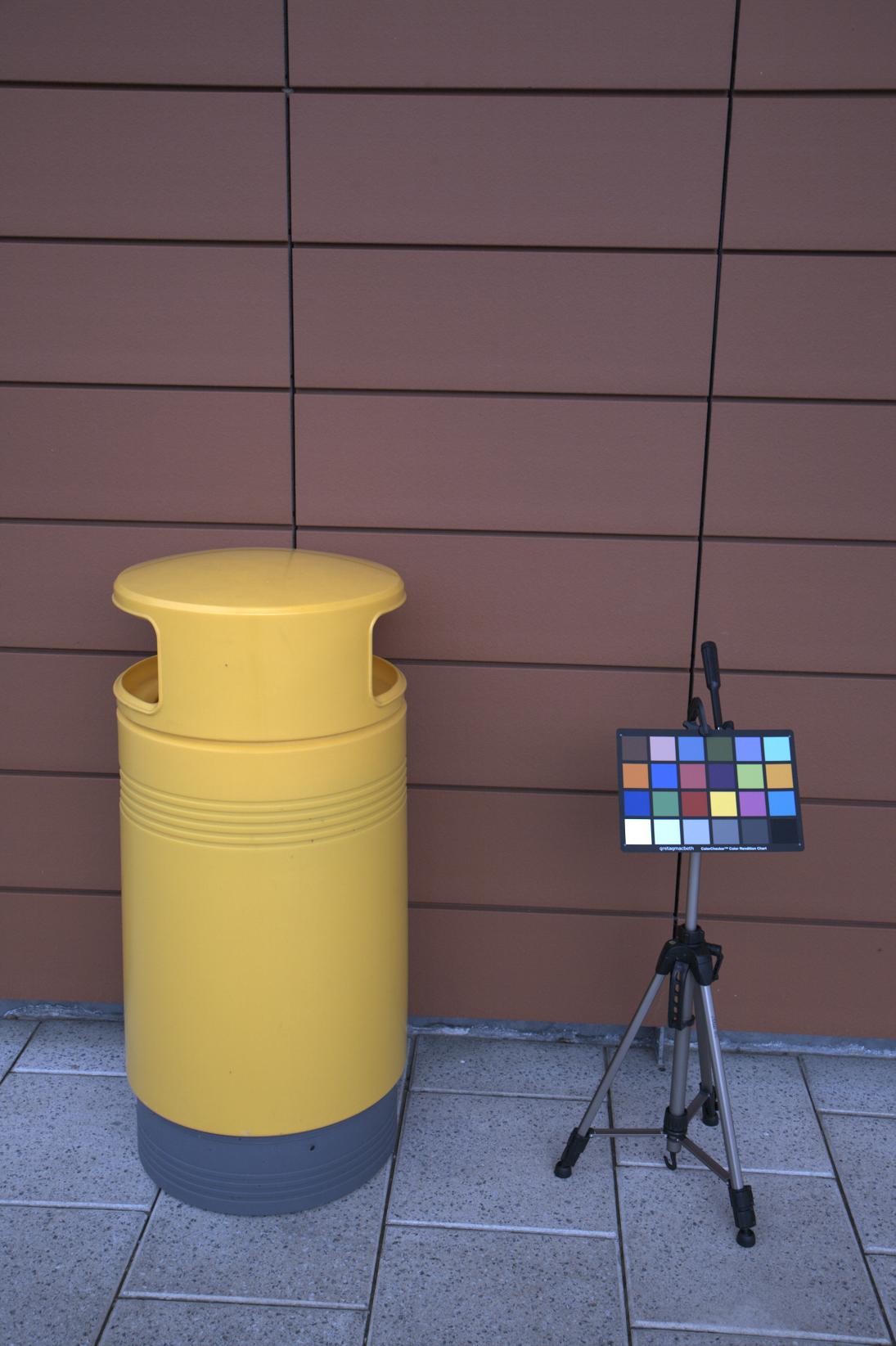} &
	\includegraphics[height=0.40\columnwidth,angle=90]{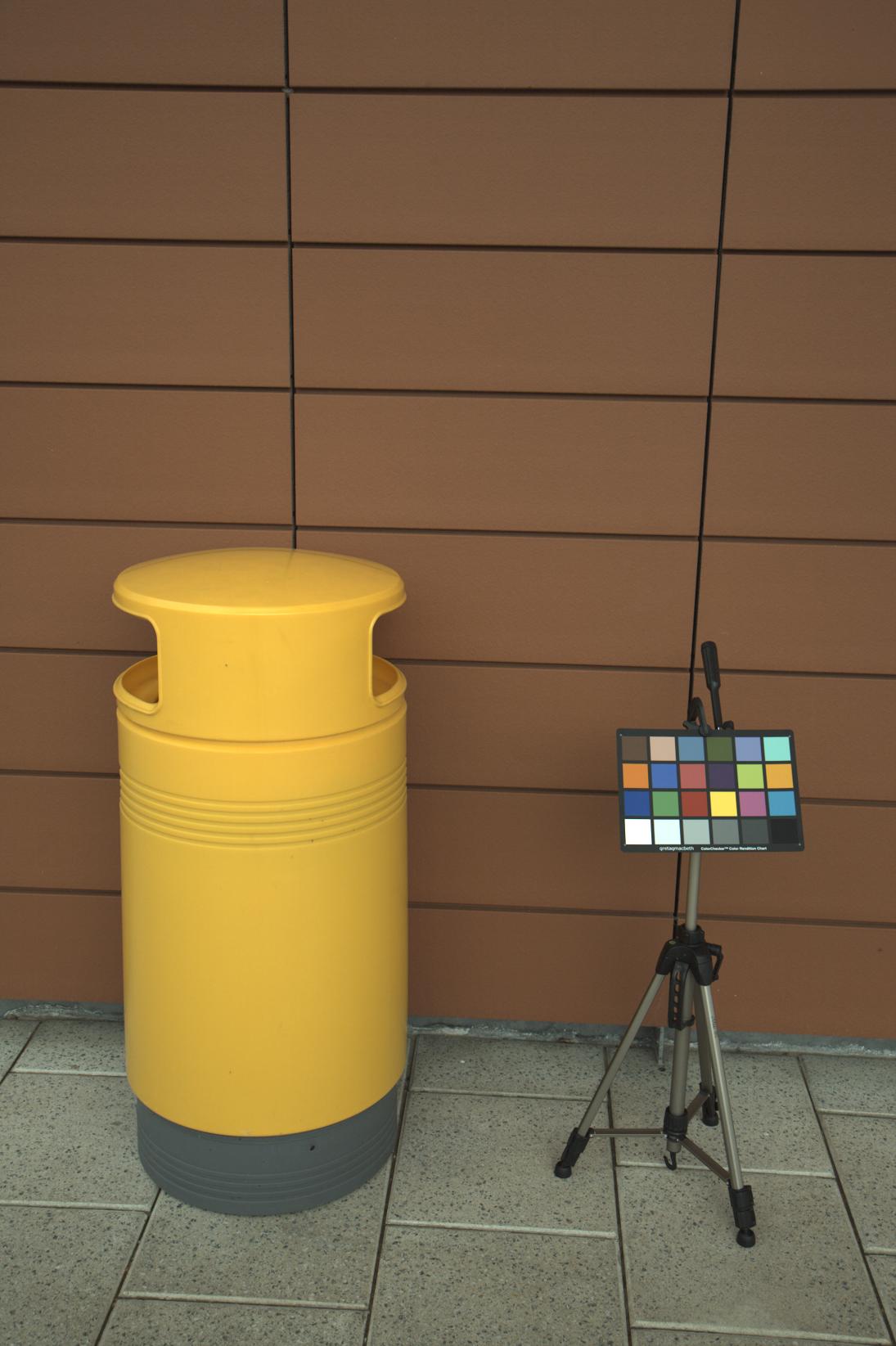} \\
	\small Input image & \small Ground truth ($0^\circ$) & \small Proposed (14.77$^\circ$) & \small GM  (0.82$^\circ$)\\
	\includegraphics[width=0.40\columnwidth]{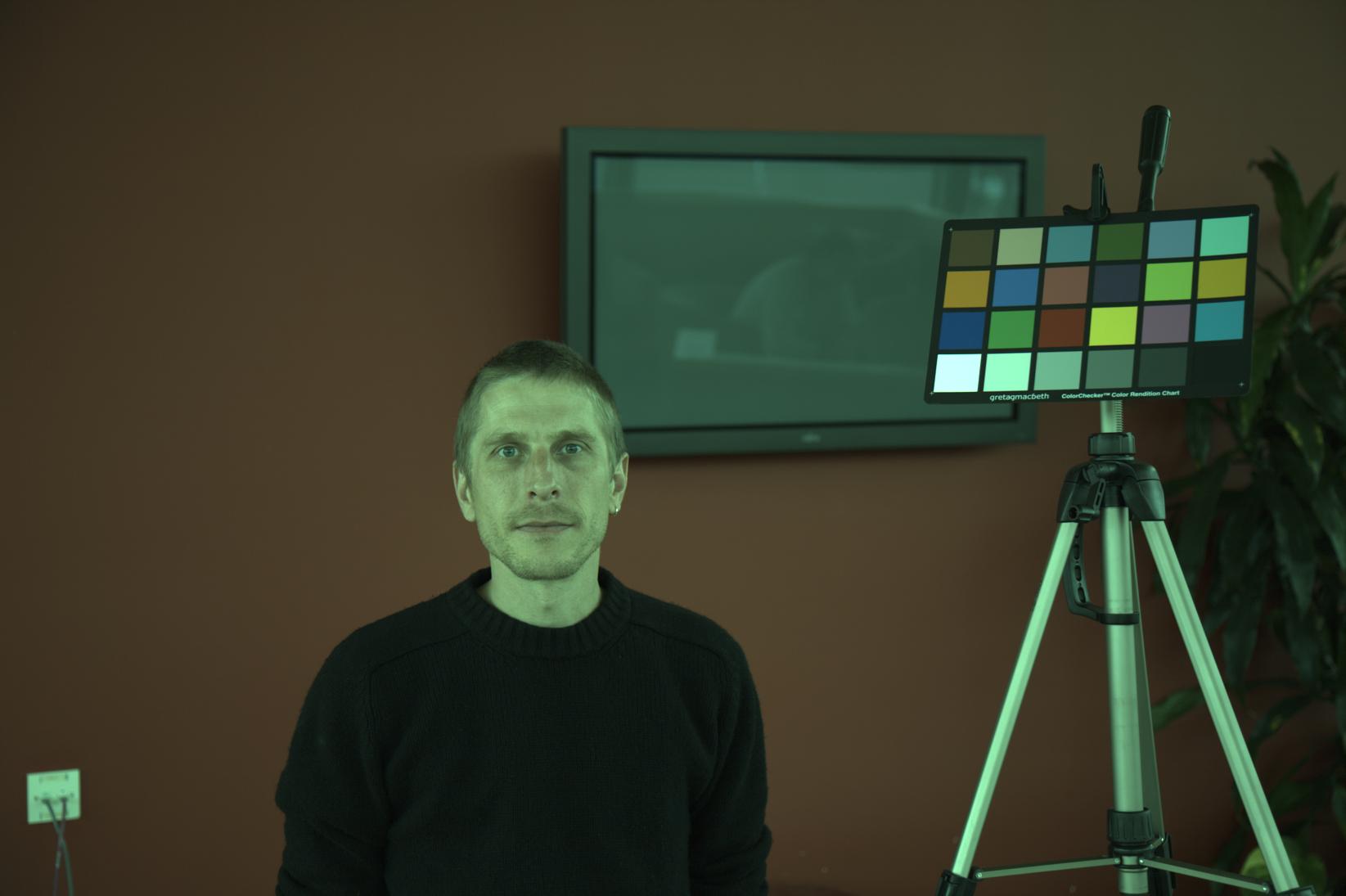} &
	\includegraphics[width=0.40\columnwidth]{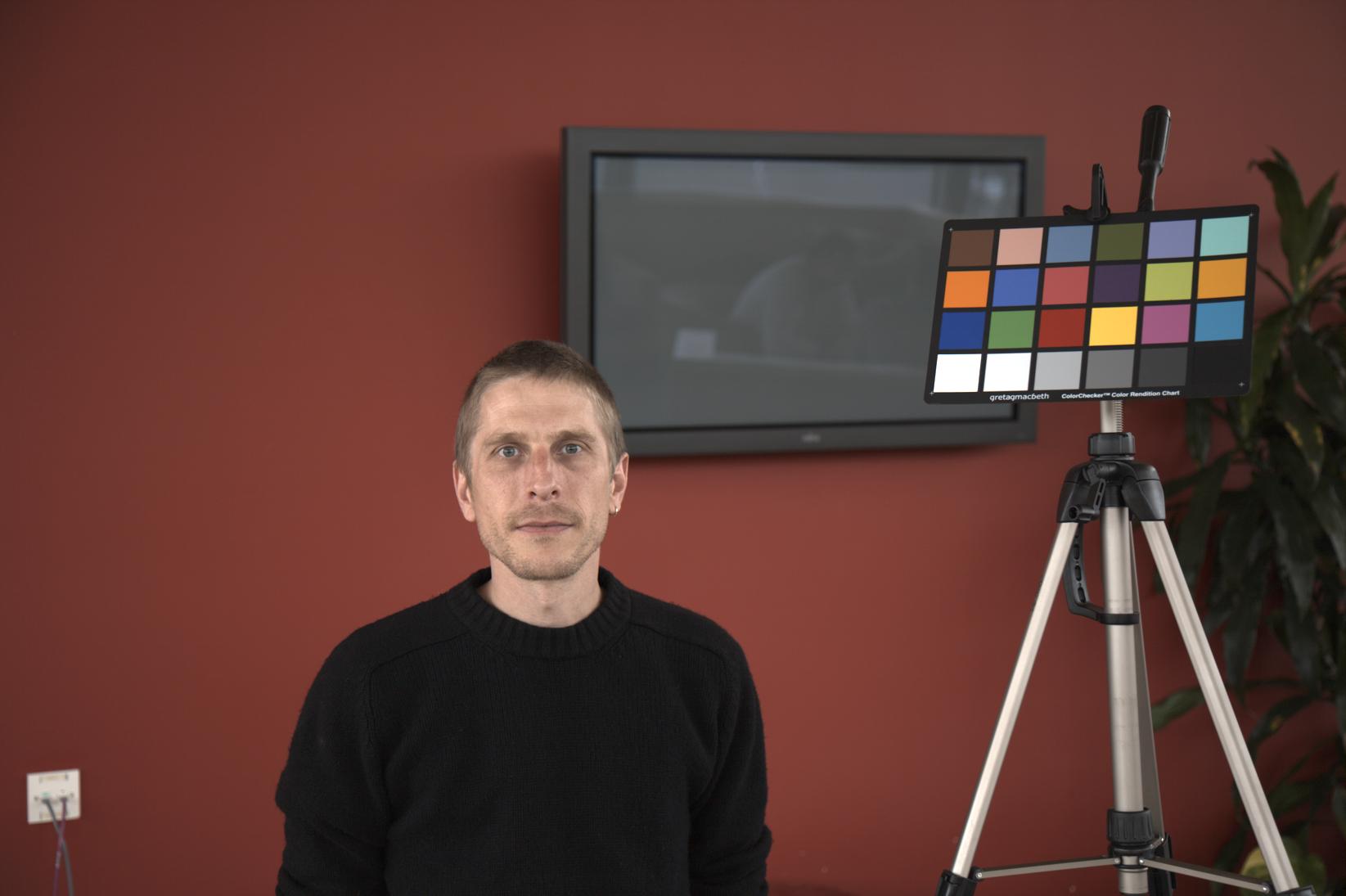} &
	\includegraphics[width=0.40\columnwidth]{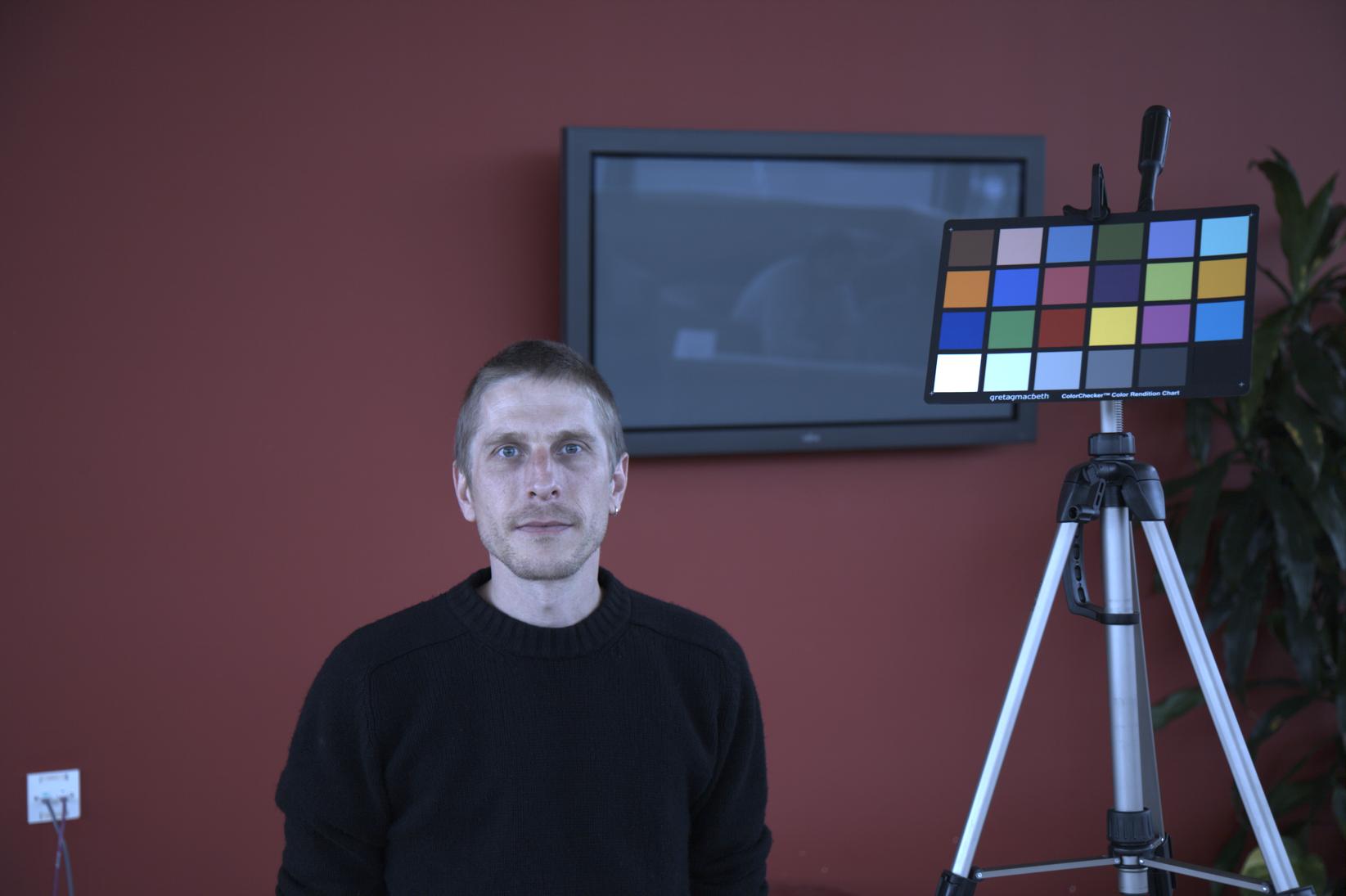} &
	\includegraphics[width=0.40\columnwidth]{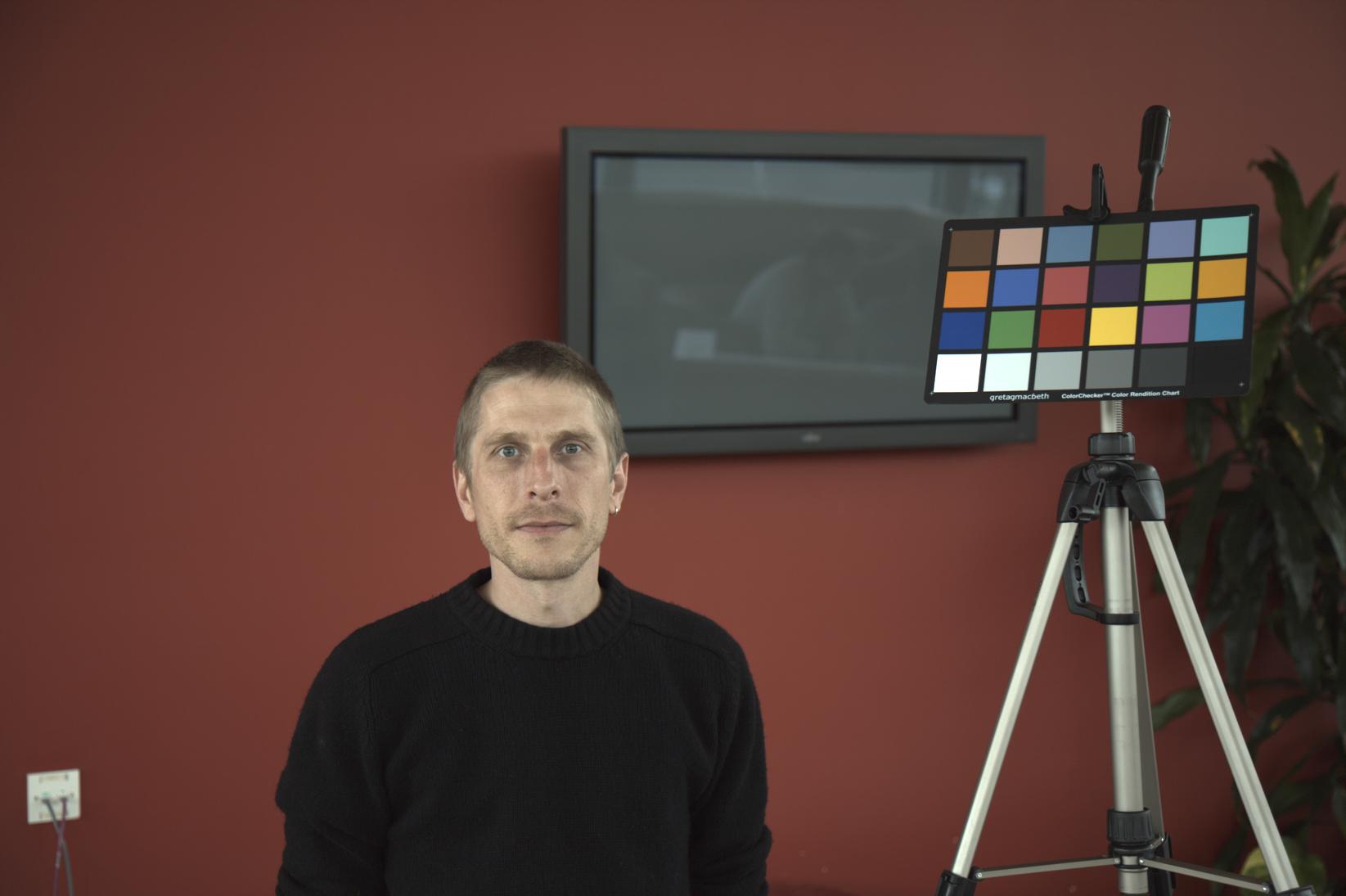} \\
	\small Input image & \small Ground truth ($0^\circ$) & \small Proposed (14.29$^\circ$) & \small FB+GM  (0.27$^\circ$) \\
	\end{tabular}
	}
	\caption{Examples of images on which the method makes the
          largest estimation errors, in the case of a single
          illuminant. Left to right: input RAW image, correction with the
          ground truth illuminant, correction
          with the illuminant estimated by the proposed method (with the local-to-global regressor enabled), and correction with the algorithm in the state-of-the-art making the best estimate on the given image.}
	\label{fig:NewWorstErrors}
\end{figure*}

Figure \ref{fig:NewWorstErrors} reports some examples of images on
which the proposed illuminant estimation method makes the largest
errors. \ADD{Even if during the illuminant estimation phase, the patches overlapping the MCC are ignored, they are left unmasked in the figure to better appreciate the results.}
Once we have an estimate of the global illuminant color $\mathbf{I}$, each pixel in the image is color corrected using the von Kries model \cite{von1902chromatic}, i.e.: $\boldsymbol{\rho}_{out}(x,y)=diag(\mathbf{I}^{-1})\boldsymbol{\rho}_{in}(x,y)$. 

\ADD{In Table \ref{tab:erroriNUS} the median angular errors obtained by the considered state-of-the-art algorithms
and the proposed approach on the NUS dataset are reported. As commonly done, results are reported separately for each camera. From the results it is possible to notice that our method outperforms the other algorithms on all cameras with an average improvement over the best algorithm of 0.35 degrees corresponding to the 15.8\%.}

\begin{table*}[!ht]
\caption{\ADD{Median angular errors obtained by the state-of-the-art algorithms considered on the NUS dataset.}}
\label{tab:erroriNUS}
\centering
\resizebox{1.00\textwidth}{!} {
\begin{tabular}{lrrrrrrrrrrrrrrrr}
  \toprule
			& GW & WP &SoG &gGW &BP &GE1 &GE2 &GM(P) &GM(E) &GM(I) &BAY &SS ML &SS GP &NIS & BDP &Prop.\\
	\midrule
Canon1 & 4.15 & 6.19 &2.73 &2.35 &2.45 &2.48 &2.44 & 4.30 & 4.68 & 4.72 &2.80 &2.80 &2.67 &3.04 & 2.01 & \bf{1.71}\\
Canon2 & 2.88 &12.44 &2.58 &2.28 &2.48 &2.07 &2.29 &14.83 &15.92 &14.72 &2.35 &2.32 &2.03 &2.46 & 1.89 & \bf{1.85}\\
Fuji   & 3.30 &10.59 &2.81 &2.60 &2.67 &1.99 &2.00 & 8.87 & 8.02 & 5.90 &3.20 &2.70 &2.45 &2.95 & 2.15 & \bf{1.75}\\
Nikon1 & 3.39 &11.67 &2.56 &2.31 &2.30 &2.22 &2.19 &10.32 &12.24 & 9.24 &3.10 &2.43 &2.26 &2.40 & 2.08 & \bf{1.88}\\
Oly    & 2.58 & 9.50 &2.42 &2.15 &2.18 &2.11 &2.18 & 4.39 & 8.55 & 4.11 &2.81 &2.24 &2.21 &2.17 & 1.87 & \bf{1.65}\\
Pan    & 3.06 &18.00 &2.30 &2.23 &2.15 &2.16 &2.04 & 4.74 & 4.85 & 4.23 &2.41 &2.28 &2.22 &2.28 & 2.02 & \bf{1.59}\\
Sam    & 3.00 &12.99 &2.33 &2.57 &2.49 &2.23 &2.32 & 7.91 & 6.12 & 6.37 &3.00 &2.51 &2.29 &2.77 & 2.03 & \bf{1.88}\\
Sony   & 3.46 & 7.44 &2.94 &2.56 &2.62 &2.58 &2.70 & 4.26 & 3.30 & 3.81 &2.36 &2.70 &2.58 &2.88 & 2.33 & \bf{1.63}\\
Nikon2 & 3.44 &15.32 &3.24 &2.92 &3.13 &2.99 &2.95 &10.99 &11.64 &11.32 &3.53 &2.99 &2.89 &3.51 & 2.72 & \bf{2.00}\\
		\bottomrule
\end{tabular}
}
\end{table*} 

\subsection{Local illuminant estimation}
Our CNN predicts the illumination on small image patches, so it can be easily used to predict  local illuminants as well as giving a global illuminant estimate for the entire image.
Given the performance of the per patch error in Table \ref{tab:errori} we expect our CNN to perform well even on local estimation. We perform here a preliminary test by using our learned CNN as-is on the synthetically relighted Geheler-Shi dataset.

\begin{figure*}
	\centering
	\setlength{\tabcolsep}{1.5pt}
	\resizebox{\textwidth}{!}{
	\begin{tabular}{cccccc}
\includegraphics[height=0.25\columnwidth]{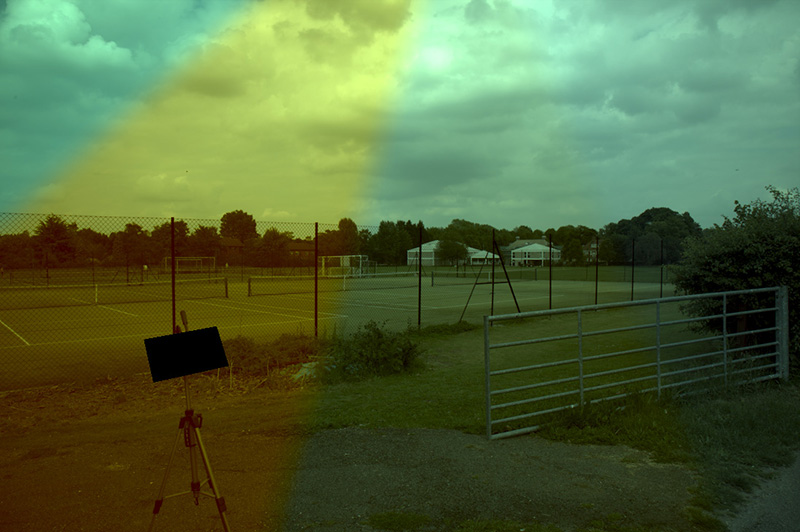} 			 & \includegraphics[height=0.25\columnwidth]{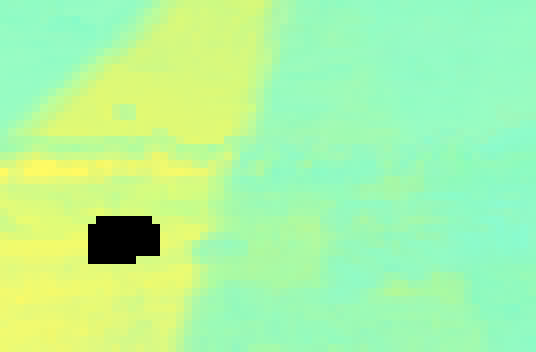} &
\includegraphics[height=0.25\columnwidth]{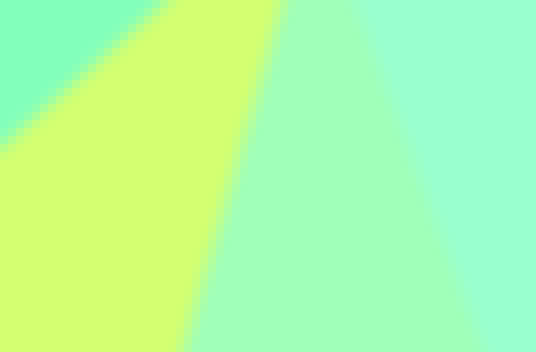}  & \includegraphics[height=0.25\columnwidth]{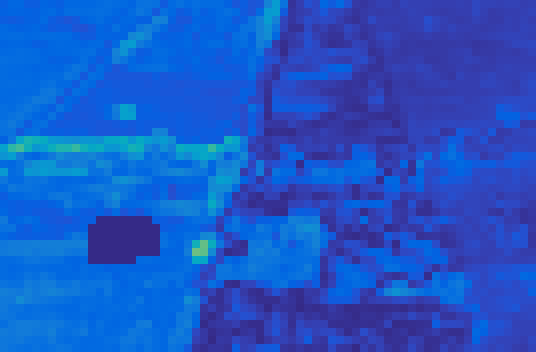} & \includegraphics[height=0.25\columnwidth]{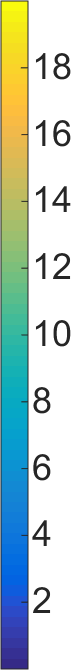} &  \includegraphics[height=0.25\columnwidth]{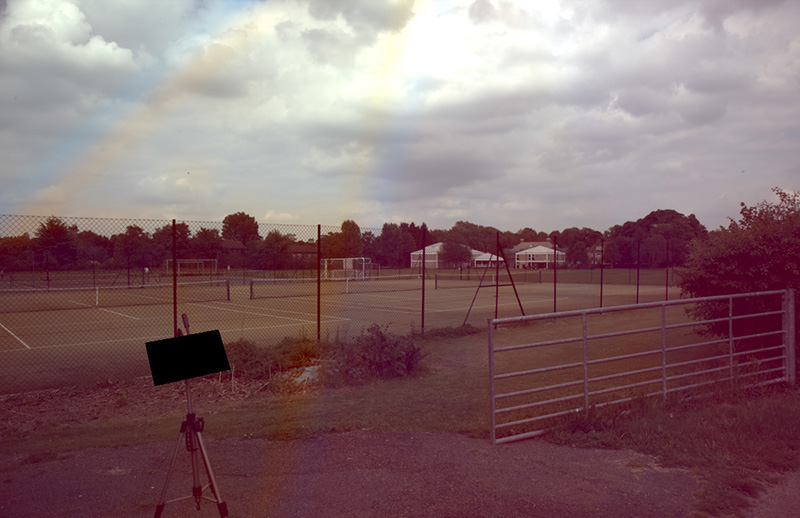}\\
\includegraphics[height=0.25\columnwidth]{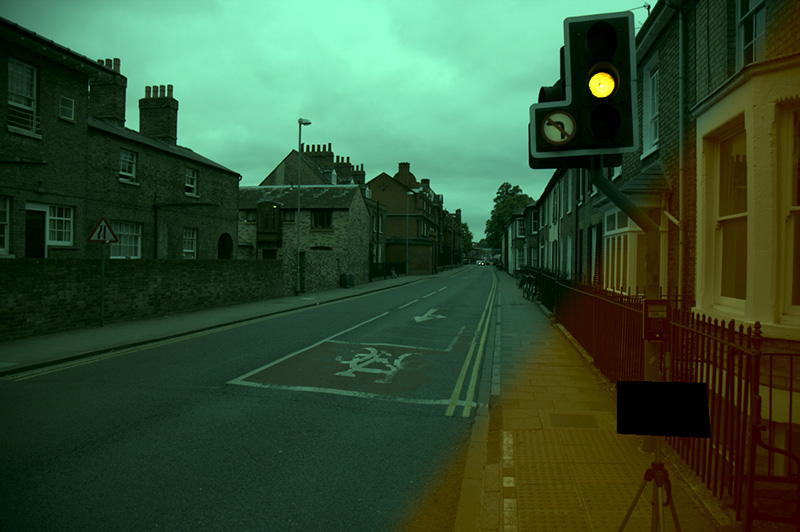} 			 & \includegraphics[height=0.25\columnwidth]{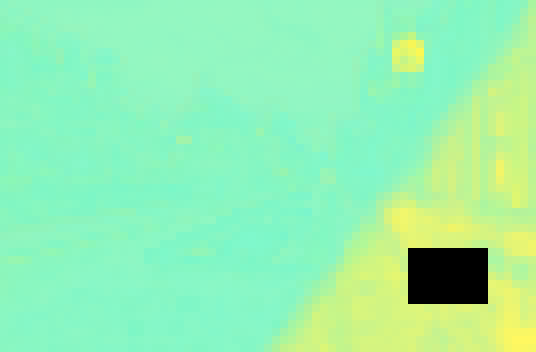}  &
\includegraphics[height=0.25\columnwidth]{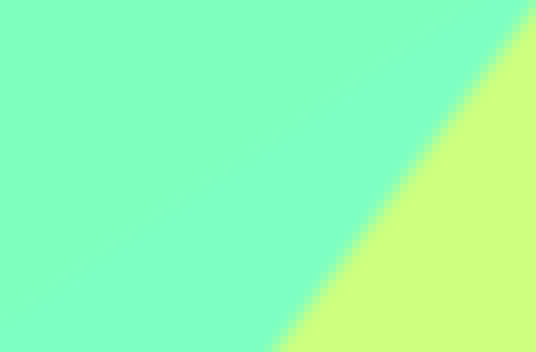}  & \includegraphics[height=0.25\columnwidth]{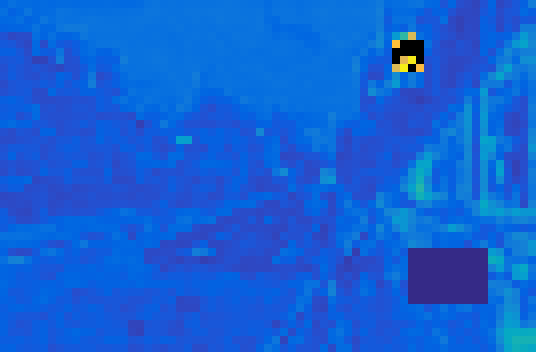} & \includegraphics[height=0.25\columnwidth]{errorbar.png} & \includegraphics[height=0.25\columnwidth]{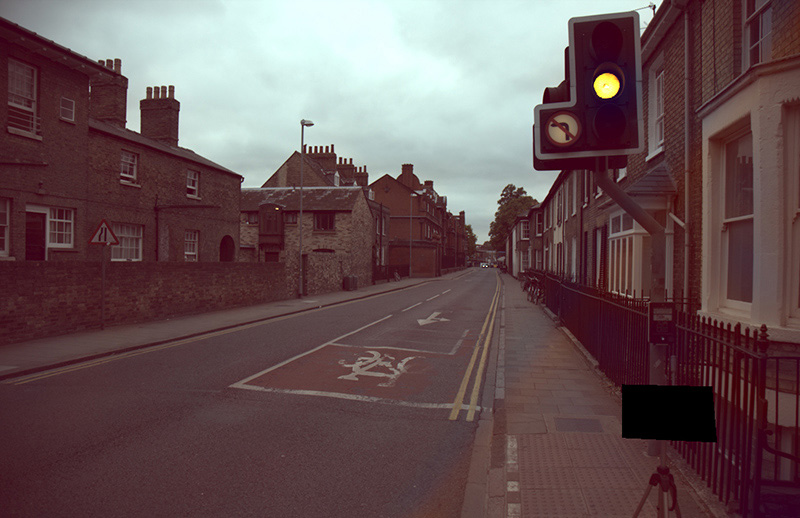}\\
Input image & Local illuminant estimate & Ground truth & Angular error map & & Corrected image\\
\end{tabular}
}
\caption{Examples of the illuminant estimation on the relighted Gehler-Shi dataset. From left to right: relighted input image, local illuminant estimate, illuminant ground truth, angular error map between estimate and ground truth, corrected image.}
\label{fig:switch}
\end{figure*}

Among the algorithms in the state-of-the-art able to deal with
non-uniform illumination,
e.g. \cite{retinex,provenzi2008spatially,LSAC,Bleier,joze2014exemplar,bianco2014adaptive}
we report as comparison the results of the Multiple Light Sources
(MLS) \cite{MLS} using White Patch (WP) and Gray World (GW)
algorithms, grid based sampling, in the clustering version setting the
number of clusters equal to the number of lights in the scene.  The
numerical results are reported in Table \ref{tab:erroriSV}, while some examples are given in Figure \ref{fig:switch}.  It is
clear that the proposed method obtain significantly better results
than all the other methods considered; the second best obtained about
twice the median error (5.92 degrees) than the proposed one (2.96
degrees).

\begin{table}[!ht]
\caption{Angular error statistics obtained on the synthetic relighted Gehler-Shi dataset with spatially varying illumination.}
\label{tab:erroriSV}
\centering
\begin{tabular}{lrrrr}
  \toprule
	Algorithm &   Med &  Avg & 90$^{th}$prc &  Max  \\
  \midrule
DN 								&  13.47 & 13.49 &  15.53 & 26.75  \\
LSAC \cite{LSAC} 	&  8.60 & 9.00 &  13.78 & 32.78  \\
RETINEX \cite{retinex} &  8.61 & 9.03 &  13.75 & 32.76  \\
MLS+WP \cite{MLS} &  5.92 & 6.90 &  12.55 & 31.09  \\
MLS+GW \cite{MLS} &  8.91 & 9.35 &  14.43 & 33.33  \\
\midrule
Proposed multiple estimate &  2.96 & 3.75  & 6.79 & 23.87 \\
\bottomrule
\end{tabular}
\end{table}

Note that this comparison has been made by disabling the detection of
multiple illuminant and by always taking the local estimates.  In a
further experiment we evaluated the performance in a mixed
single/multi illuminant scenario.  The dataset used is the single
illuminant version of the Gehler-Shi and one-third of the
synthetically relighted version so that the numbers of images having
single and multiple illuminants are equal. The numerical results are
reported in Table \ref{tab:erroriDBswitch}, where the performance of
the four variants of the proposed method are reported:
%
i) single illuminant, that always applies the local-to-global \regressor; 
ii) multi illuminant, that always keeps the local estimates; 
iii) the fully automatic, that uses the multiple illuminant detector to decide if the local-to-global \regressor{} must be applied or not;
iv) the oracle, that applies the local-to-global \regressor{} only when, according to the ground truth, the image present a single illuminant.
The results obtained show that the use of the multiple illuminant
detector allows to obtain better results with respect to adopting a
single strategy.  Its performance are very close to those that can be
obtained by exploiting the ground truth information about the presence
of single or multiple illuminants (i.e.~the oracle version). 
%
\begin{table}[!ht]
\caption{Angular error statistics obtained by variants of the proposed method on a mixture of the original and the relighted Gehler-Shi dataset.}
\label{tab:erroriDBswitch}
\centering
\begin{tabular}{lrrrr}
  \toprule
	Algorithm & Med &  Avg & 90$^{th}$prc &  Max  \\
  \midrule
single illuminant	&  2.73	& 3.54	&	7.62 &	23.61 \\
multi illuminant	&  3.08	& 4.10	&	8.23 &	23.42 \\
fully automatic	  &  2.50	& 3.05	&	5.94 &	20.10 \\
oracle						&  2.48	& 3.05	&	5.98 &	20.10 \\
	\bottomrule
\end{tabular}
\end{table} 

The first experiment on real world data is performed on the subset of the Milan portrait
dataset containing multiple MCCs. The numerical results are reported
in Table \ref{tab:erroriNewDBmultiMCC}, where the performance of the
proposed method are reported enabling the multiple illuminant detector
to decide if the local-to-global \regressor{} must be applied or
not. The results obtained show that the proposed method performs
better than all the single illuminant estimation algorithms as well as
all the general purpose multiple illuminant estimation ones. The only
algorithm able to outperform the one proposed here is the face-based
\cite{bianco2014adaptive}, which is specifically designed to leverage
skin properties in images containing faces.

An example taken from the Milan portrait dataset is reported in Figure \ref{fig:sv1}. \ADD{Since ground truth illuminant is available only on the MCCS, pixel-level ground truth is obtained by linear interpolation. As usual, MCCs are ignored during illuminant estimation but are left unmasked in the figure to better understand the results.}

\begin{table}[!ht]
\caption{Angular error statistics obtained on the Milan portrait dataset.}
\label{tab:erroriNewDBmultiMCC}
\centering
\begin{tabular}{lrrrrrr}
  \toprule
	Algorithm &  Med &  Avg & 90$^{th}$prc &  Max  \\
  \midrule
DN														&   17.30 &  17.53 &  19.74 &  28.60 \\
WP														&  12.16 &  11.39 &  19.08 &  28.60 \\
GW														&  4.26 &   4.86 &   9.26 &  20.04 \\
SoG														&    4.39 &   5.93	&  14.03 &  20.02 \\
gGW														&    5.25 &   6.42	&  15.07 &  20.80 \\
GE1														&    4.59 &   5.08	&   9.50 &  18.22 \\
GE2														&    4.93 &   5.39	&   9.69 &  {15.36} \\
SS ML													&    2.94 &   {3.72} &   8.03 &  16.28  \\
LSAC      \cite{LSAC}					&    4.23 &   4.79 &   8.66 &  18.99  \\
RETINEX    \cite{retinex}   	&    4.28 &   4.83 &   8.39 &  20.54 \\
MLS + WP          \cite{MLS}  &    3.21 &   4.04 &   {7.55} &  17.19 \\
MLS + GW          \cite{MLS}  &    3.33 &   4.18 &   8.82 &  17.97 \\
Fusion Grad. Tree Boost.  \cite{Bleier}    &  4.48 & 5.29 & 9.95 & 31.26 \\ 
Fusion Rand. Forest Regr.  \cite{Bleier}   &  3.23 & 3.96 & 7.61 & 27.76 \\ 
Face-based	\cite{bianco2014adaptive}		&    \bf{2.11} &   \bf{2.66} &   \bf{5.15} &  \bf{11.43} \\ 
\midrule
Proposed (fully automatic) &  2.75 &	3.30 & 	6.24	& 15.22 \\	
	\bottomrule
\end{tabular}
\end{table}

{\setlength{\tabcolsep}{1.5pt}
\begin{figure*}[!ht]%
\centering
\resizebox{\textwidth}{!} {
\begin{tabular}{ccccc}
\includegraphics[width=0.50\columnwidth]{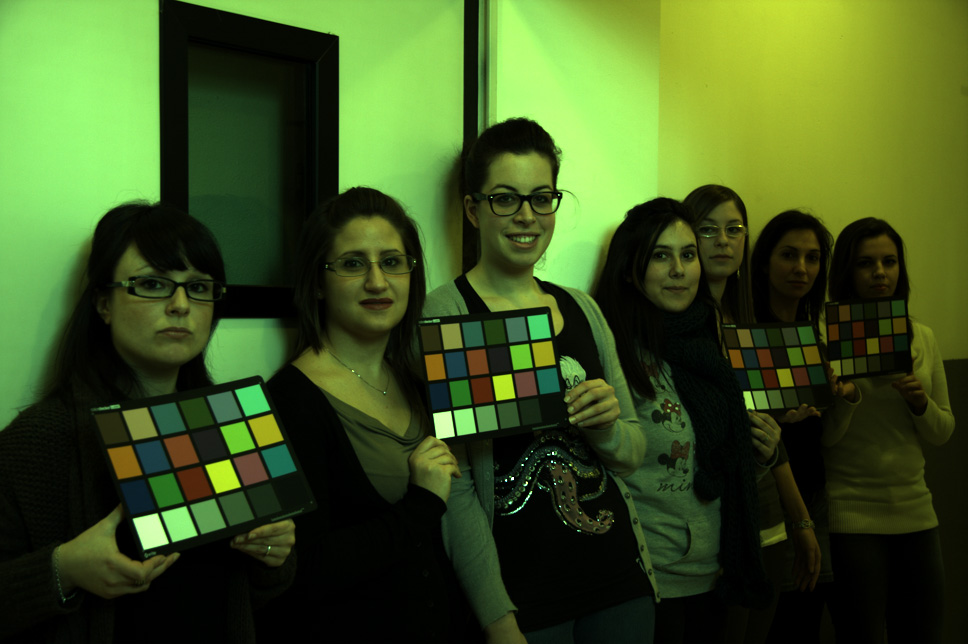} &
\includegraphics[width=0.50\columnwidth]{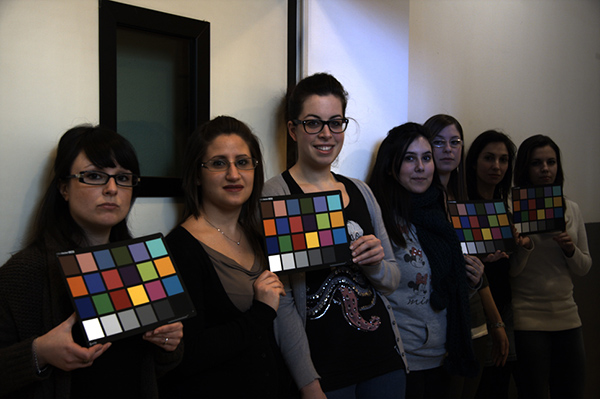} &
\includegraphics[width=0.50\columnwidth]{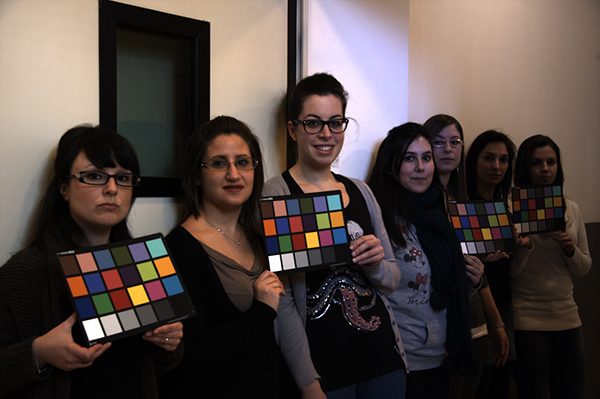} &
\includegraphics[width=0.50\columnwidth]{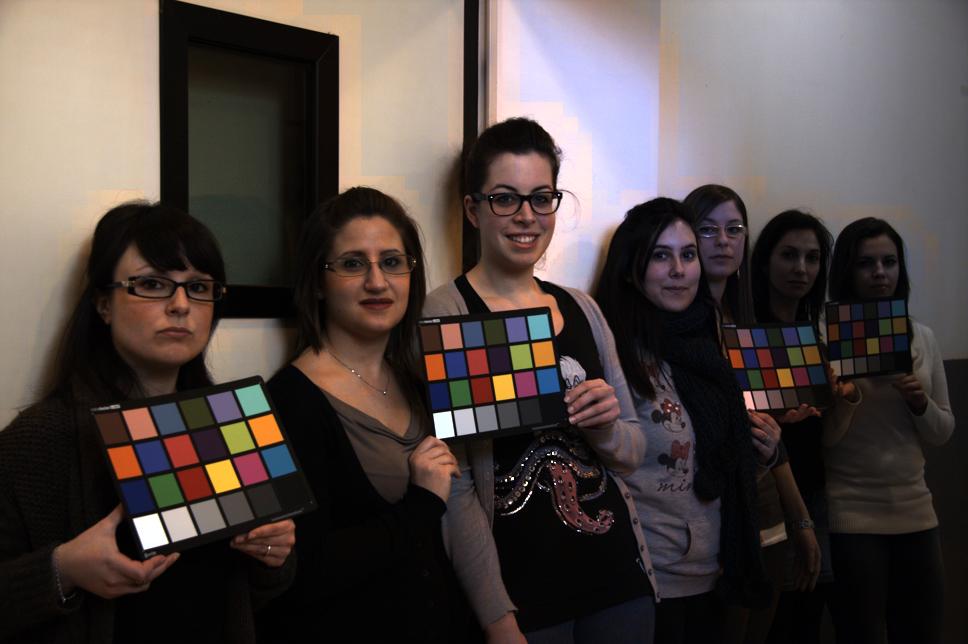} & \\
\small Input image & \small correction using the ground truth & \small correction using Face-based \cite{bianco2014adaptive}  & \small correct. using the proposed meth. & \\
\includegraphics[width=0.50\columnwidth]{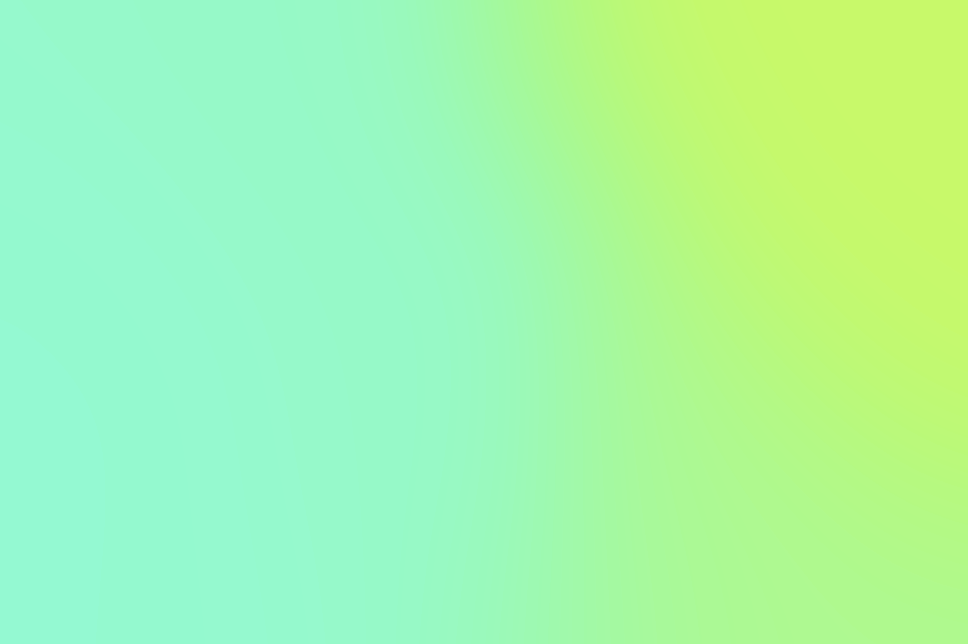} &
\includegraphics[width=0.50\columnwidth]{img_65_AWB_SP_SV_gt_orig.png} &
\includegraphics[width=0.50\columnwidth]{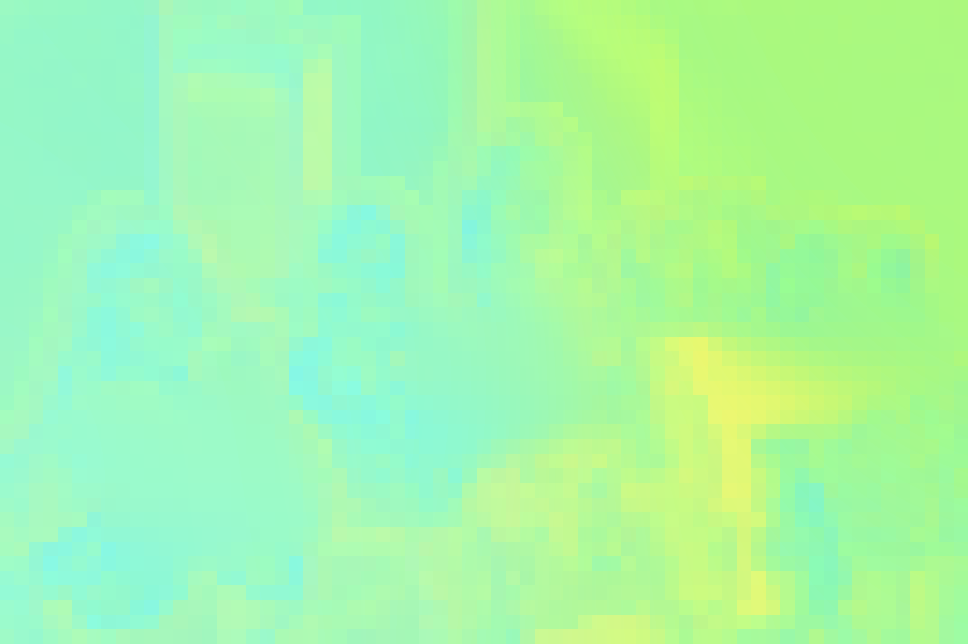} &
\includegraphics[width=0.50\columnwidth]{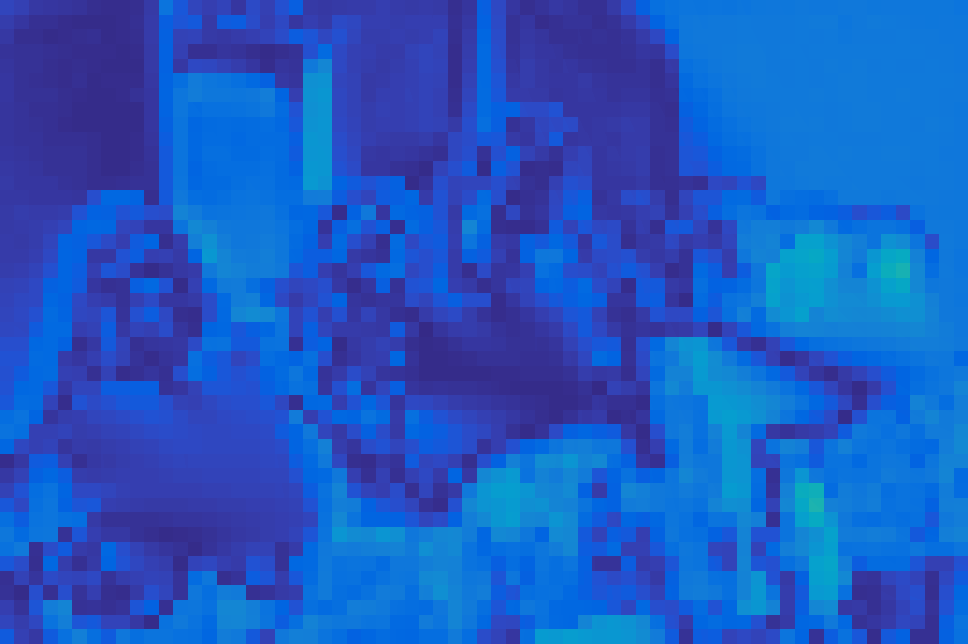} &
\includegraphics[height=0.33\columnwidth]{errorbar.png} \\
\small illuminant ground truth & \small Face-based illuminant estimate & \small our illuminant estimate & \small angular error & \\
\end{tabular}
}
\caption{Example image with multiple illuminants taken from the Milan portrait dataset. Top row: input image, correction using the ground truth, correction using the illuminant estimate from the Face-based method \cite{bianco2014adaptive}, and correction using the estimate of the proposed method: since multiple illuminant were detected in the input image, the method outputs a local illuminant estimate. Bottom row: illuminant ground truth, illuminant estimate from the Face-based method, estimate from the proposed method, and angular error map between our estimate and the ground truth.}
\label{fig:sv1}
\end{figure*}
}

\ADD{The last experiment concerning local illuminant estimation is performed on the multiple illuminant dataset by Beigpour et al.  The numerical results are reported in Table \ref{tab:erroriDBMIRF}, where the performance of the
proposed method are reported enabling the multiple illuminant detector
to decide if the local-to-global \regressor{} must be applied or
not. The results are reported separately for the laboratory and real-world settings. In both cases the results obtained show that the proposed method performs better than all the algorithms considered with an average reduction of the median error of almost 
14\%.}

\begin{table}[!htbp]
\caption{\ADD{Average and median angular errors obtained on the two parts of the Beigpur et al. dataset \cite{beigpour2014multi}: laboratory (left) and real-world (right).}}
\label{tab:erroriDBMIRF}
\centering
\begin{tabular}{cc}
\begin{tabular}{lrr}
  \toprule
	Algorithm & Avg & Med \\
  \midrule
DN											& 10.6 & 10.5 \\
GW											&  3.2 &  2.9 \\
WP											&  7.8 &  7.6 \\
GE1											&  3.1 &	2.8 \\
GE2											&  3.2 &	2.9 \\
IEbv										&  8.5 &  8.3  \\
\midrule
MLS + GW  \cite{MLS}    &  6.4 &  5.9 \\
MLS + WP  \cite{MLS}    &  5.1 &  4.2 \\
MLS + GE1 \cite{MLS}		&  4.8 &  4.2 \\
MLS + GE2 \cite{MLS}		&  5.9 &  5.7 \\
\midrule
MIRF + GW	\cite{beigpour2014multi}		&  3.1 &  2.8 \\
MIRF + WP	\cite{beigpour2014multi}		&  3.0 &  2.8 \\
MIRF + GE1 \cite{beigpour2014multi}	&  2.7 &  2.6 \\
MIRF + GE2 \cite{beigpour2014multi}	&  2.6 &  2.6 \\
MIRF + IEbV \cite{beigpour2014multi}	&  4.5 &  3.0 \\
\midrule
Proposed (fully automatic)								&  \bf{2.3} &  \bf{2.2} \\
	\bottomrule
\end{tabular}
&
\begin{tabular}{rr}
  \toprule
	 Avg & Med \\
  \midrule
											  8.8 &  8.9 \\
											  5.2 &  4.2 \\
											  6.8 &  5.6 \\
											  5.3 &	 3.9 \\
											  6.0 &	 4.7 \\
										    6.0 &  4.9  \\
\midrule
      4.4 &  4.3 \\
      4.2 &  3.8 \\
		  9.1 &  9.2 \\
		 12.4 & 12.4 \\
\midrule
  3.7 &  3.4 \\
  4.1 &  3.3 \\
  4.0 &  3.4 \\
  4.9 &  4.5 \\
  5.6 &  4.3 \\
\midrule
  \bf{3.3} &  \bf{3.1} \\
	\bottomrule
\end{tabular}
\end{tabular}
\end{table}

\section{Network architecture}
\label{sec:netarch}
\ADD{In this section we discuss the design of the network, how its
performance is affected by the parameters, and how we can relate the
behavior of the learned model to that of other methods for
computational color constancy.}

\ADD{The architecture of the network has been designed by starting from a
deep CNN similar to the LeNet~\cite{lecun1998gradient} and by removing layers until no
further improvement in performance was possible. The final model is a
simplified convolutional neural network with a single convolutional
layer, max pooling, and two fully connected layers.  Differently from
other computer vision tasks, deepening the network causes slightly
worse results.  This fact probably depends on the small variability in
content provided by the annotated data sets for computational color
constancy.  In fact, our training patches come from a few hundreds of
images, while deep networks are often trained on millions of annotated
images.}

\begin{figure}
	\centering
		\setlength{\tabcolsep}{-0pt}
	\resizebox{\columnwidth}{!}{
	\begin{tabular}{cc}
		\includegraphics[width=0.63\columnwidth]{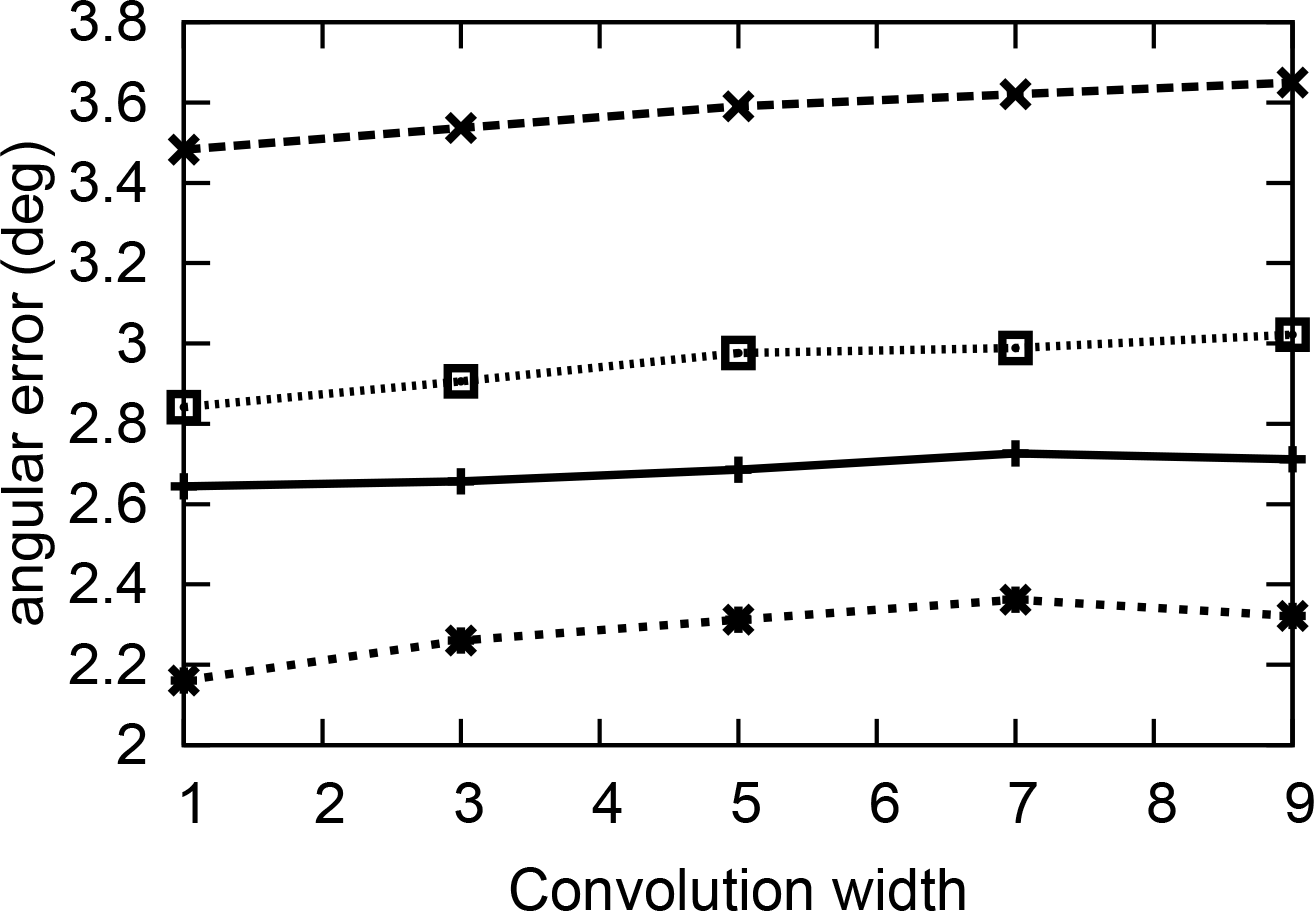} &
		\includegraphics[width=0.615\columnwidth]{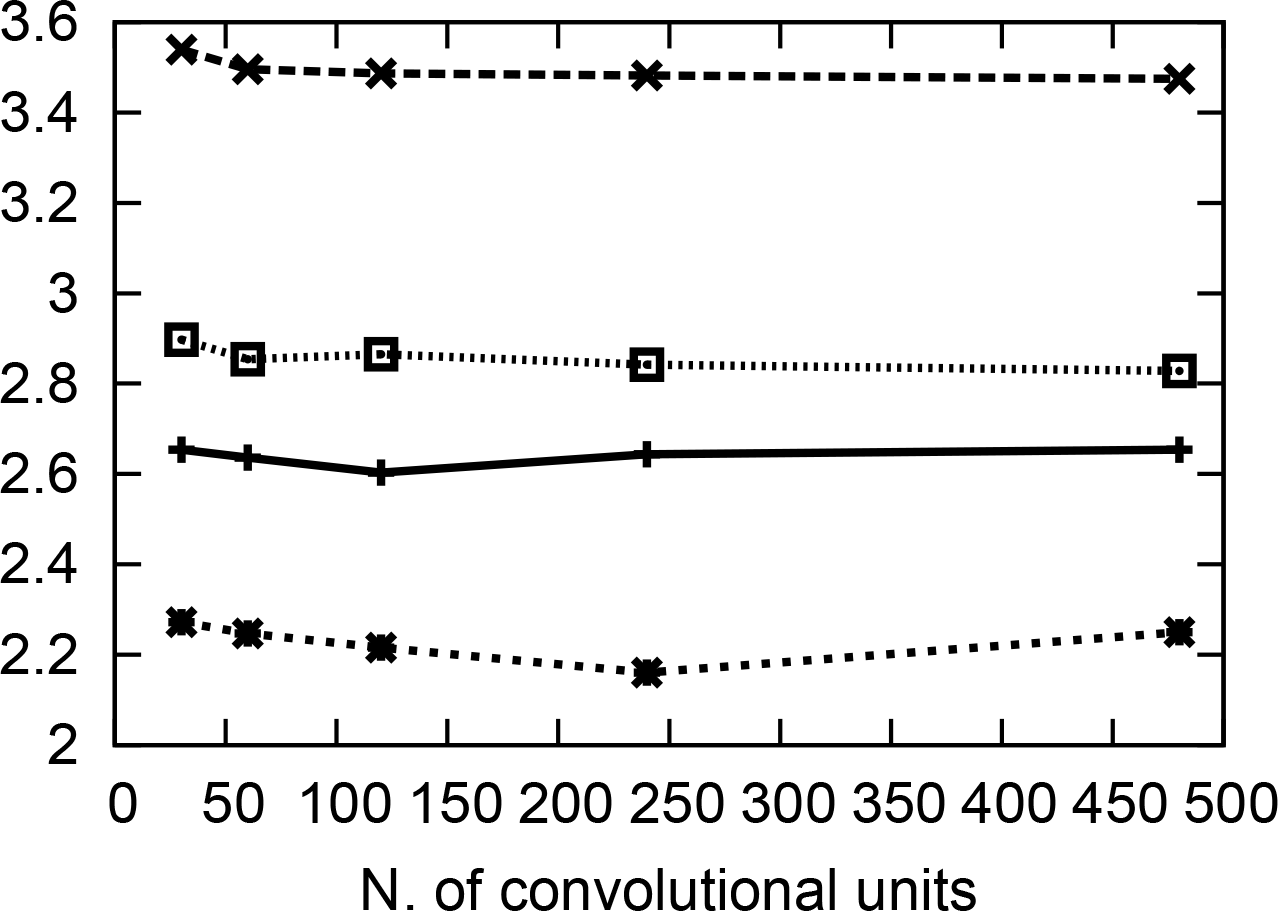} \\
		a) & b) \\
		\includegraphics[width=0.63\columnwidth]{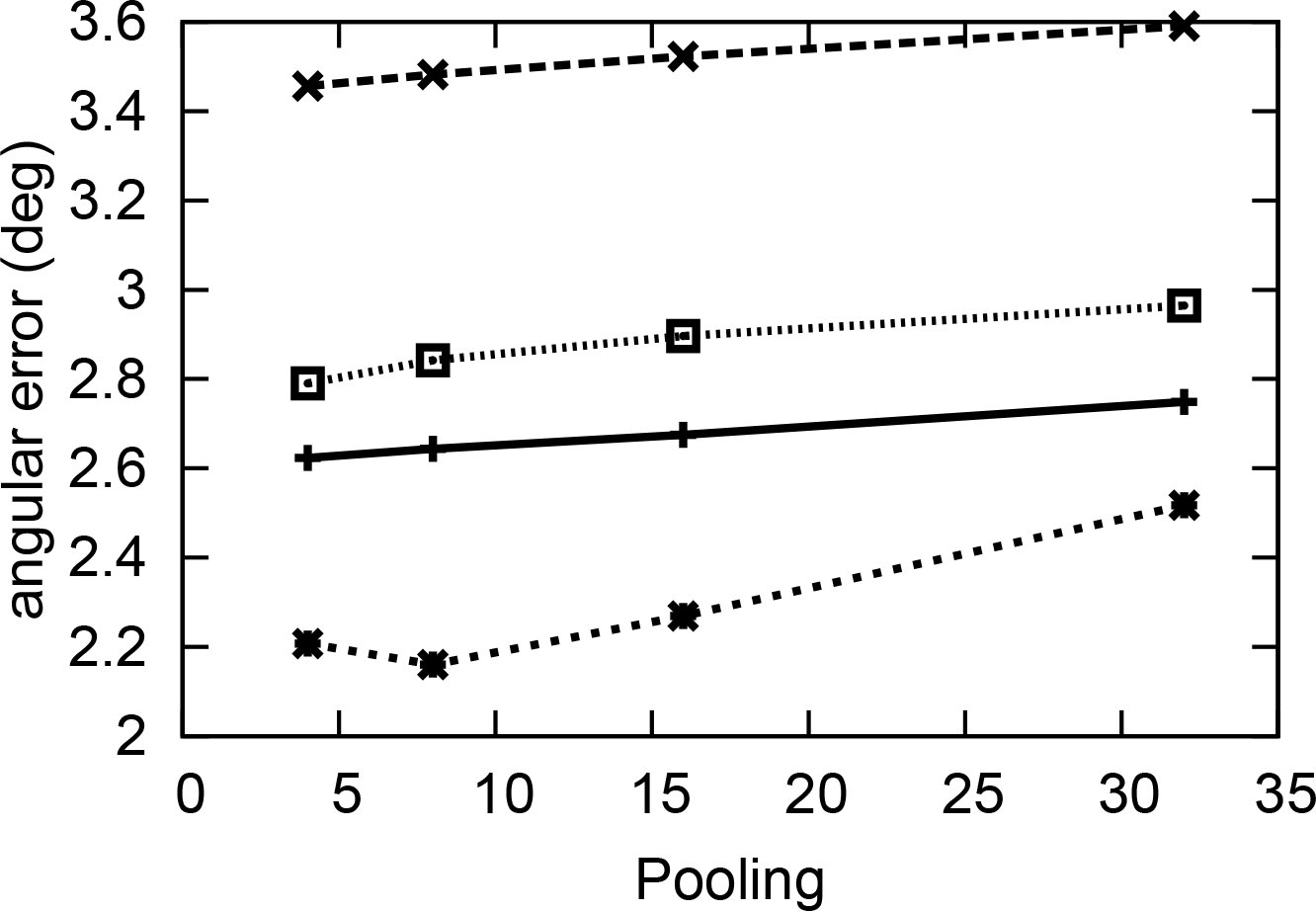} &
		\includegraphics[width=0.605\columnwidth]{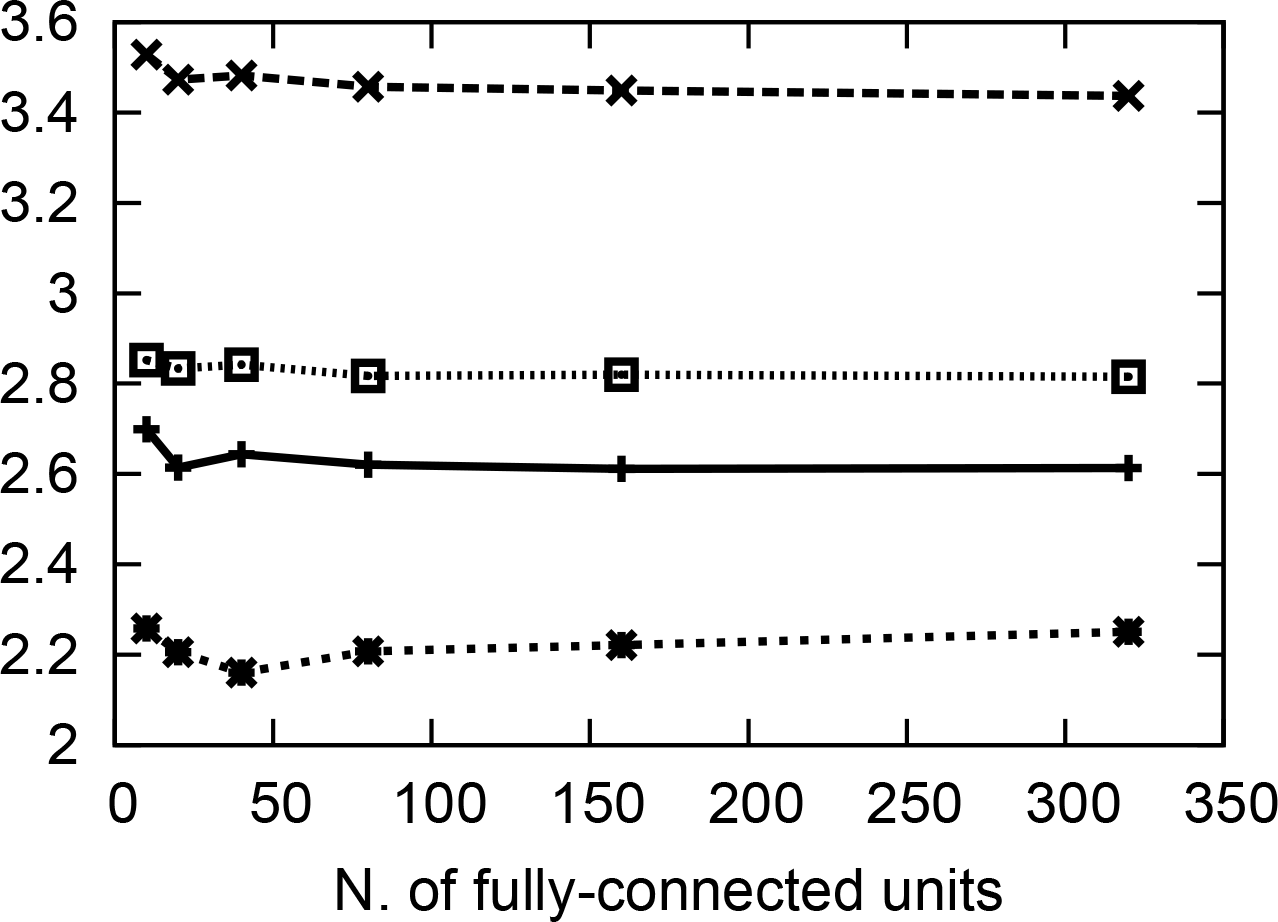} \\
		c) & d) \\
		\includegraphics[width=0.63\columnwidth]{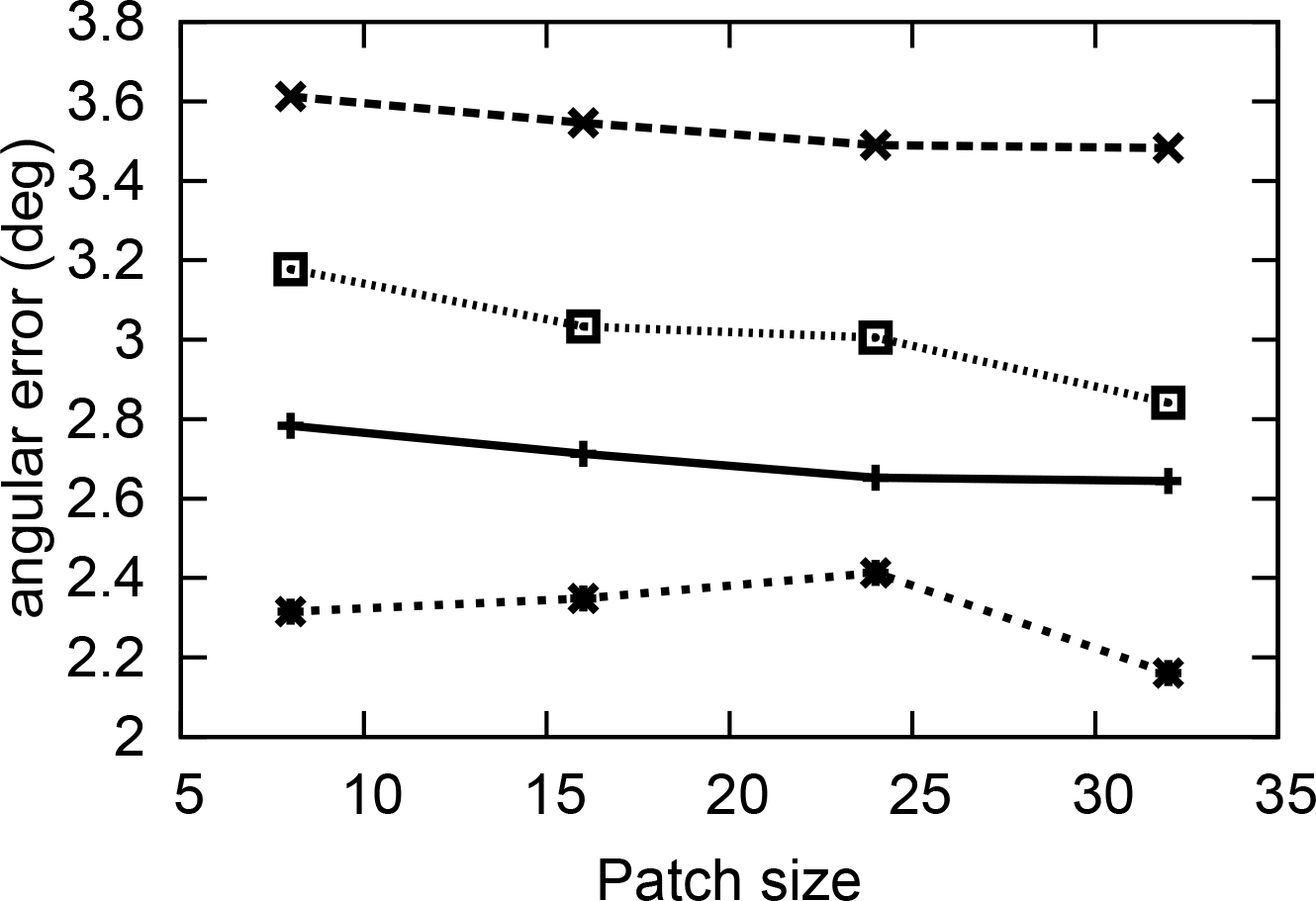} & 
		\includegraphics[width=0.61\columnwidth]{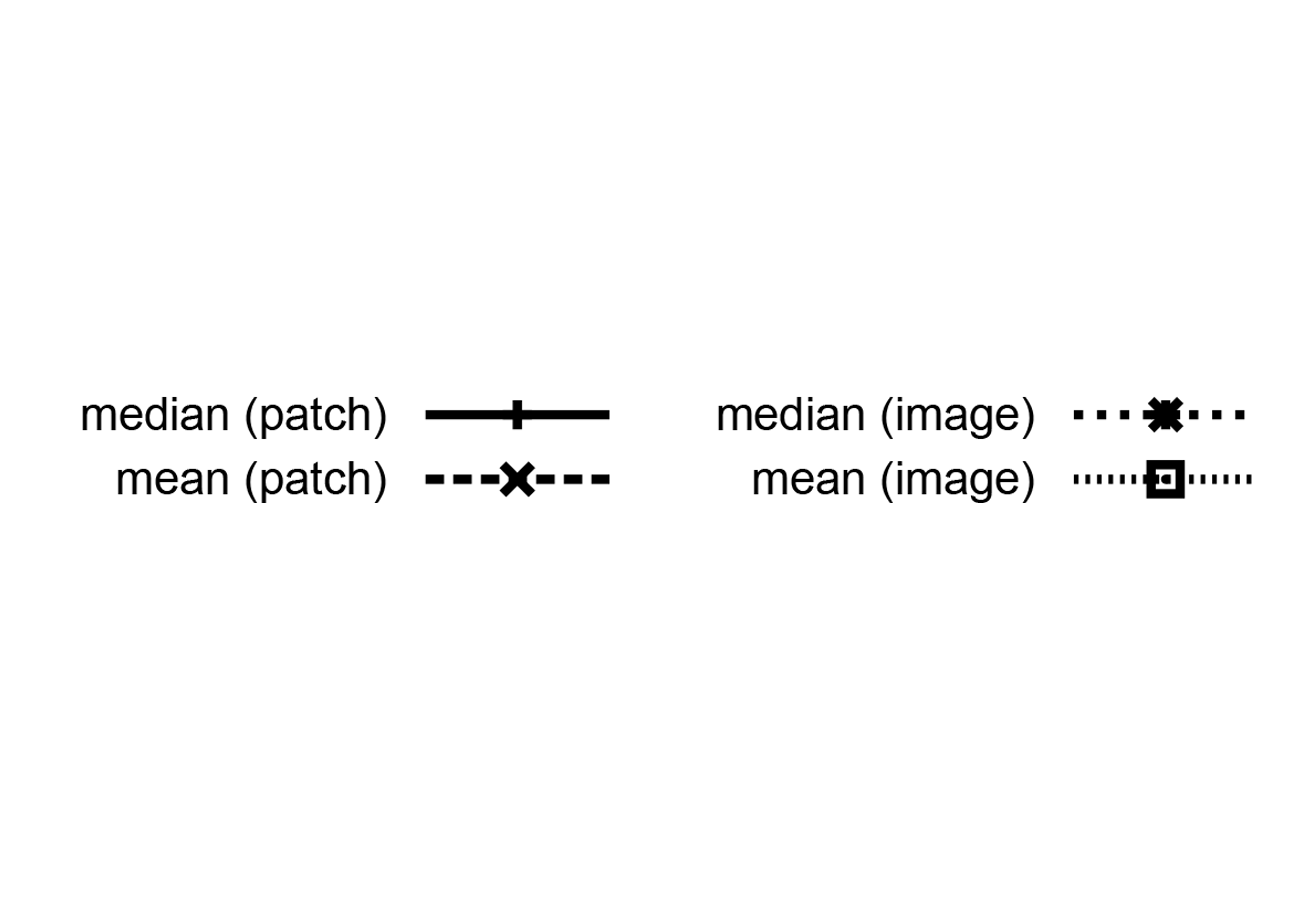}\\
		e) & \\
	\end{tabular}
	}
	\caption{Effects of the parameters on the CNN performance. Angular error with respect to varying convolution kernel width (a), number of convolutional kernels (b), pooling size (c), number of fully connected units (d) and input patch size (e). Each point corresponds to the best performance that can be obtained by fixing a single parameter at the value indicated by trying all the combinations for the other ones.}
	\label{fig:parametri}
\end{figure}

\ADD{The performance of the network are quite robust with respect to its
parameters as shown in Figure~\ref{fig:parametri}, that reports the
variation in accuracy as a function of the size of the input patches,
of the width and number of convolutional kernels, of the size of the
receptive field of the pooling units, and of the number of fully
connected units in the second to last layer.  The plots are obtained
by changing one parameter at a time while setting all the others as in
the optimal configuration.  In additional tests, not reported here, we
also measured the performance obtained by varying multiple parameters
without obtaining any surprising result.}

\ADD{The most striking element of the final network is the use of
$1\times1$ ``convolutional'' units.  At first this could be
surprising, since in different domains larger kernels are
preferred. However, it is not the first time that such small kernels
are used, see~\cite{szegedy2014going}.  In our case, networks built
with larger convolutions failed to reproduce the spatial filters (edge
detectors etc.) that are usually observed in CNNs trained for image
classification.  From the color constancy point of view, this choice of
kernel size seems to confirm the finding by Cheng at
al.~\cite{cheng2014illuminant}, that local spatial information does not
provide any additional information that cannot be obtained directly
from the color distributions.  The number of the convolutional kernels
seems less important and we found that the optimal value was around
240.}

\ADD{Another interesting element is represented by the relatively large
($8 \times 8$) receptive fields of the pooling units.  As a
consequence the max pooling layer strongly reduces the dimensionality
of the incoming data, while retaining just some spatial information.
Smaller receptive fields resulted in a decrease in the performance of
the network.  We observe a sort of duality with respect to the
parameters used for CNNs for image classification that usually prefer
large convolutional kernels and small pooling units.  
}

\ADD{Concerning the remaining parameters, we found that the optimal number
of fully-connected units was intermediate (40) and that the network
prefers large $32 \times 32$ patches over smaller ones.}


\subsection{Model interpretation}
\label{sec:netarch1}

\ADD{After training the network, we analyzed the resulting weights for the
three layers with learning capabilities.  The last layer maps the 40
intermediate values (``features'', in the following) in the three
components of the illuminant estimated for the input patch.  The
transformation is affine and is represented by a matrix of
$40 \times 3$ coefficients and by 3 biases.  A layer of this kind has
been already shown to perform well by Funt et
al.~\cite{funt1996learning}, where it was used to process the
responses of indicator functions over a regular quantization of the
image chromaticities.  It is also similar to combinational
methods~\cite{li2014evaluating} where the outputs of different color
constancy algorithms are combined to give the final illuminant
estimate.}

\begin{figure*}[!ht]%
\includegraphics[width=\textwidth]{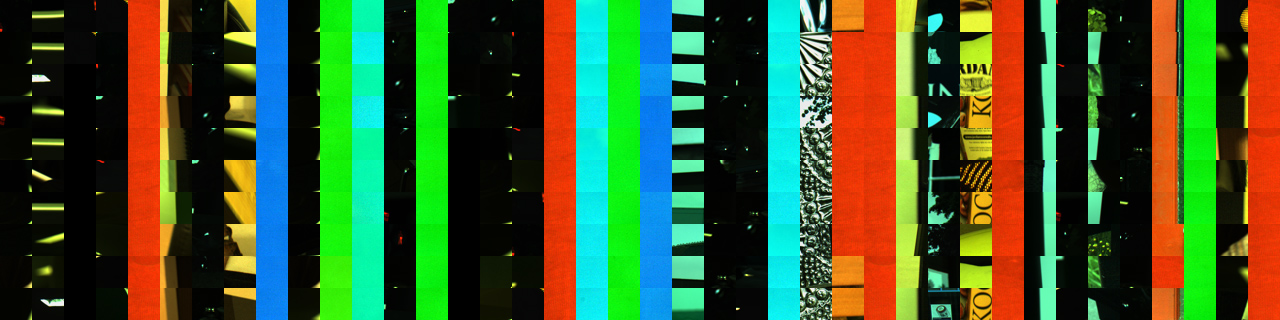}%
\caption{Image patches producing the highest activations of the 40 neurons in the fully connected layer: each column represents a different neuron and reports in decreasing order the patches corresponding to its top ten activations. }%
\label{fig:patchactivations}%
\end{figure*}

\begin{figure*}[!ht]%
\centering
\resizebox{\textwidth}{!} {
\begin{tabular}{ccccccccccc}
\includegraphics[width=0.350\columnwidth]{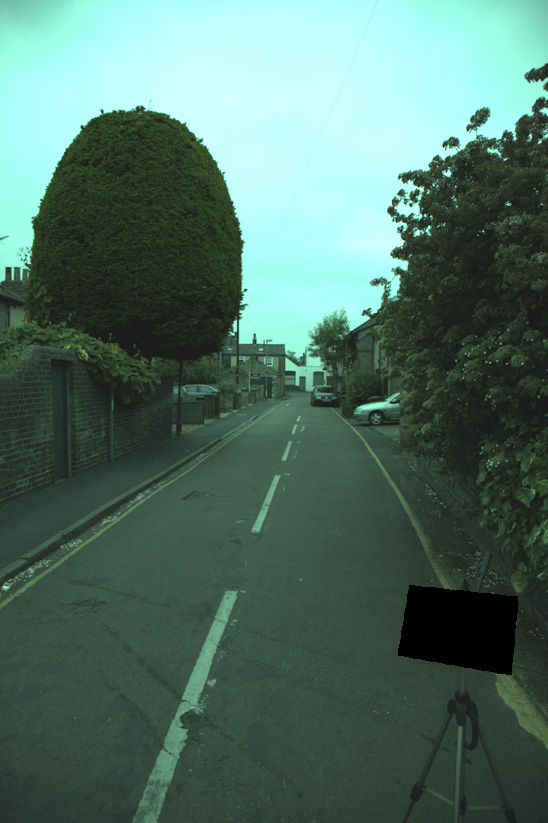} &
\includegraphics[width=0.350\columnwidth]{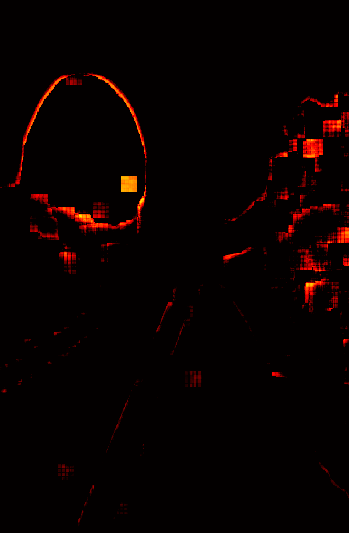} &
\includegraphics[width=0.350\columnwidth]{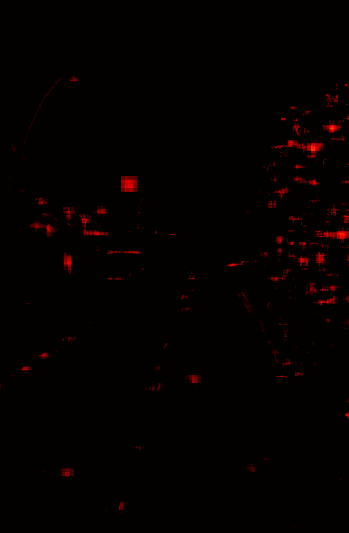} &
\includegraphics[width=0.350\columnwidth]{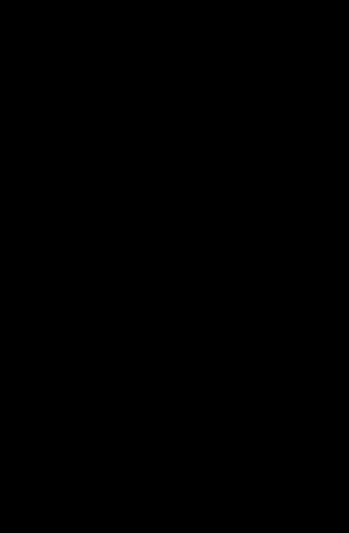} &
\includegraphics[width=0.350\columnwidth]{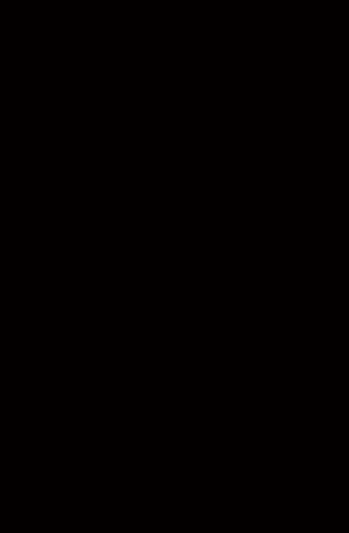} &
\includegraphics[width=0.350\columnwidth]{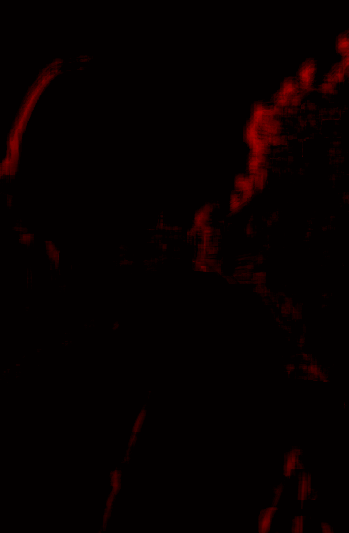} &
\includegraphics[width=0.350\columnwidth]{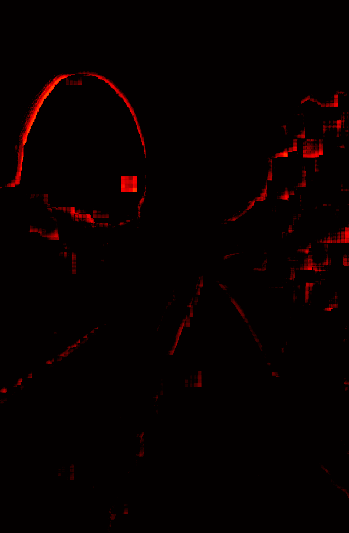} &
\includegraphics[width=0.350\columnwidth]{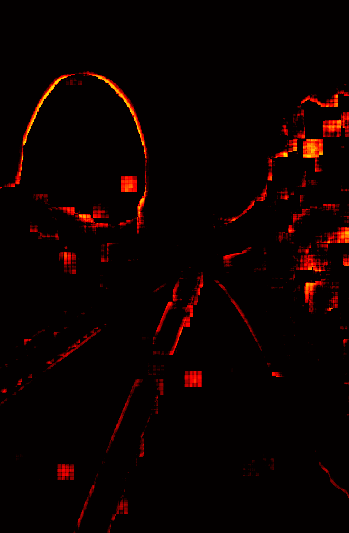} &
\includegraphics[width=0.350\columnwidth]{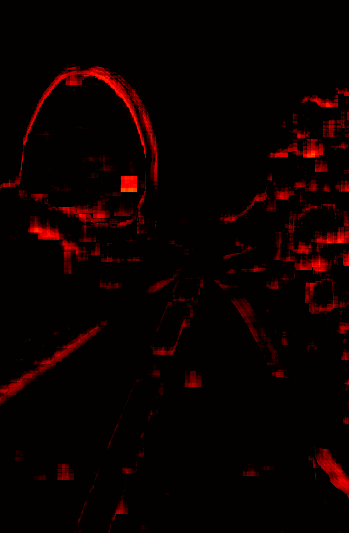} &
\includegraphics[width=0.350\columnwidth]{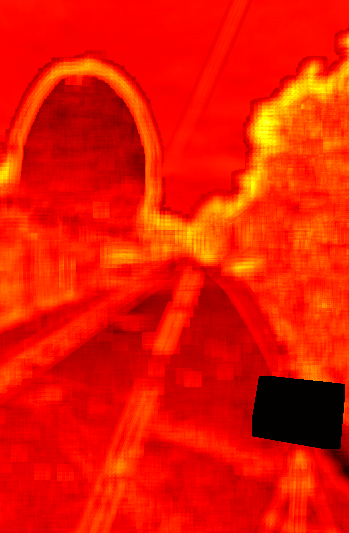} &
\includegraphics[width=0.350\columnwidth]{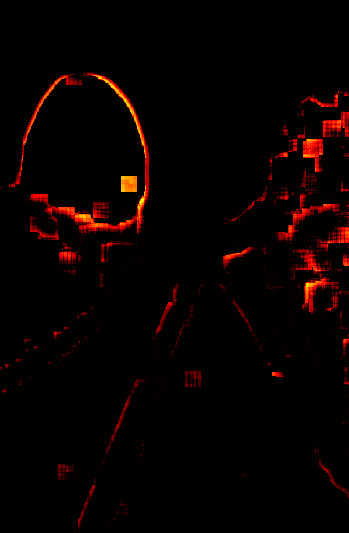} \\
&
\includegraphics[width=0.350\columnwidth]{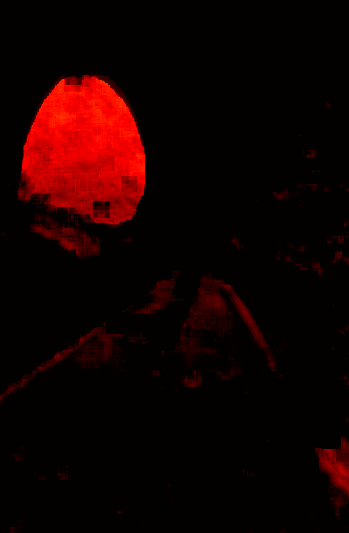} &
\includegraphics[width=0.350\columnwidth]{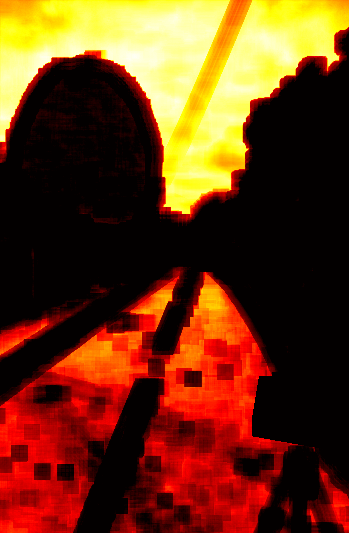} &
\includegraphics[width=0.350\columnwidth]{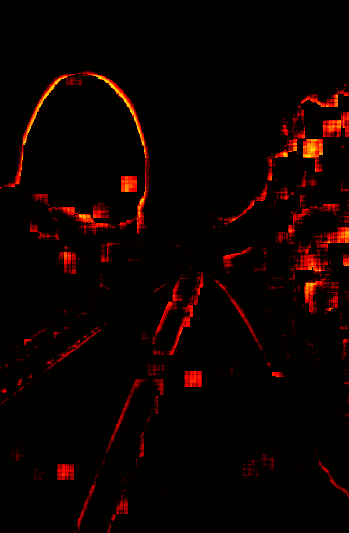} &
\includegraphics[width=0.350\columnwidth]{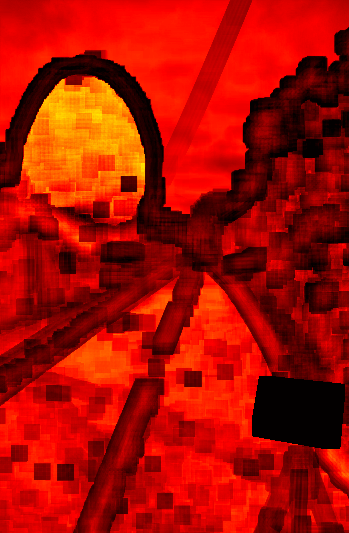} &
\includegraphics[width=0.350\columnwidth]{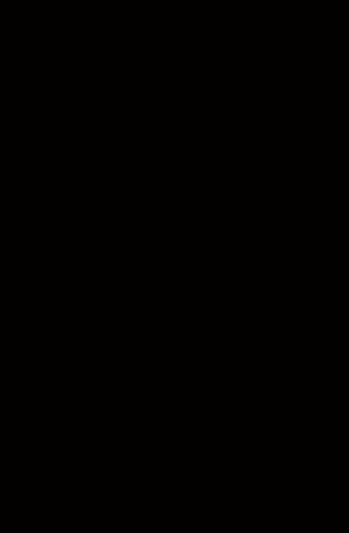} &
\includegraphics[width=0.350\columnwidth]{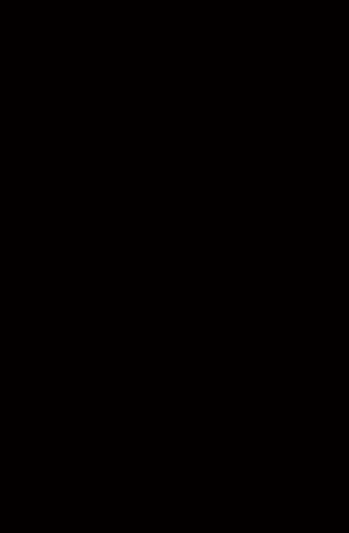} &
\includegraphics[width=0.350\columnwidth]{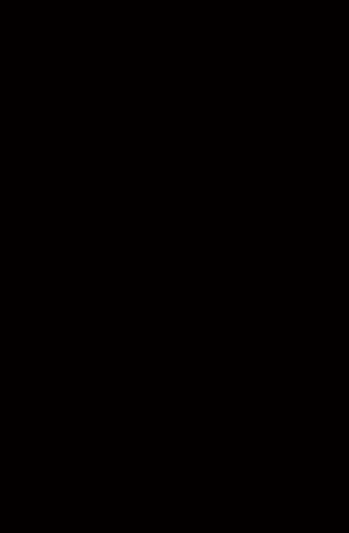} &
\includegraphics[width=0.350\columnwidth]{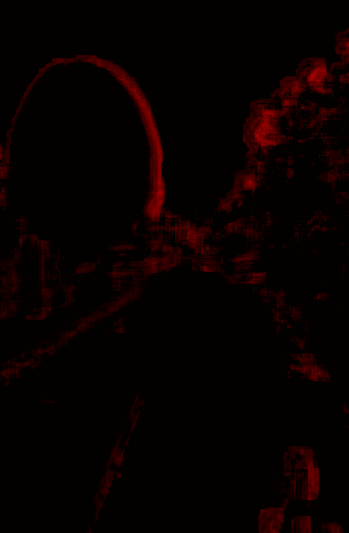} &
\includegraphics[width=0.350\columnwidth]{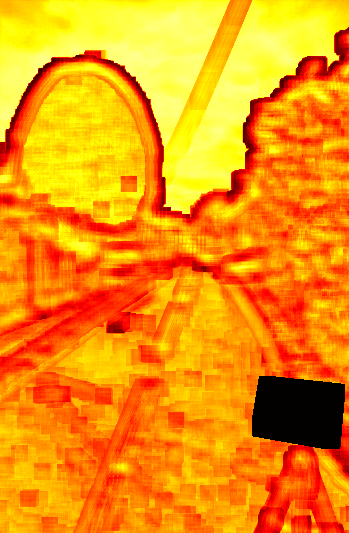} &
\includegraphics[width=0.350\columnwidth]{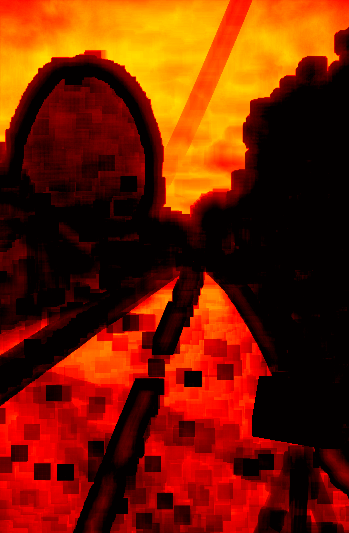} \\
&
\includegraphics[width=0.350\columnwidth]{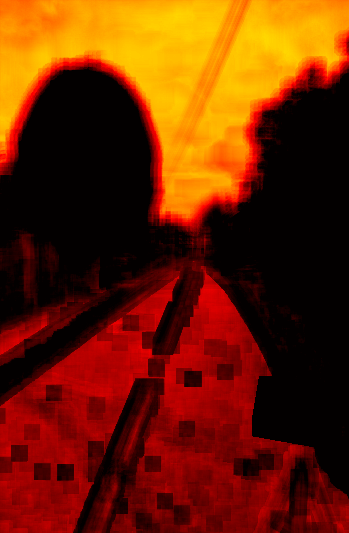} &
\includegraphics[width=0.350\columnwidth]{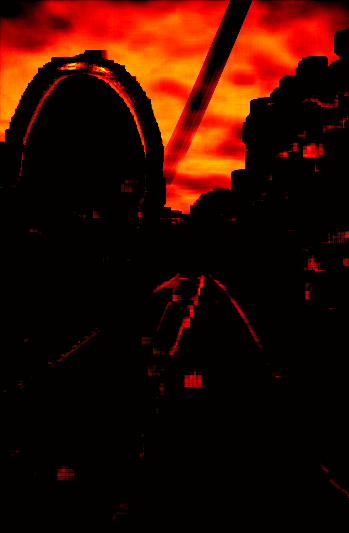} &
\includegraphics[width=0.350\columnwidth]{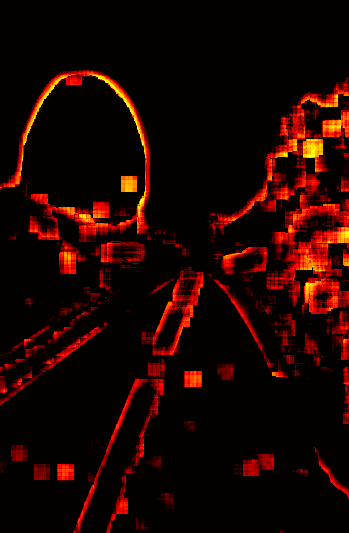} &
\includegraphics[width=0.350\columnwidth]{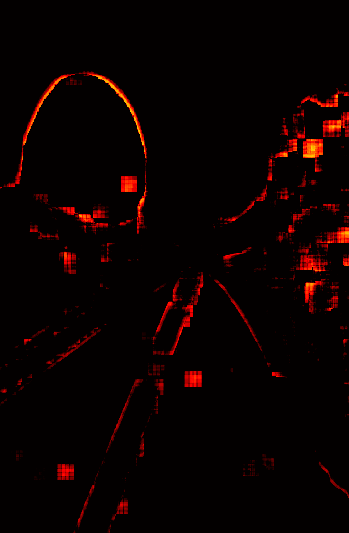} &
\includegraphics[width=0.350\columnwidth]{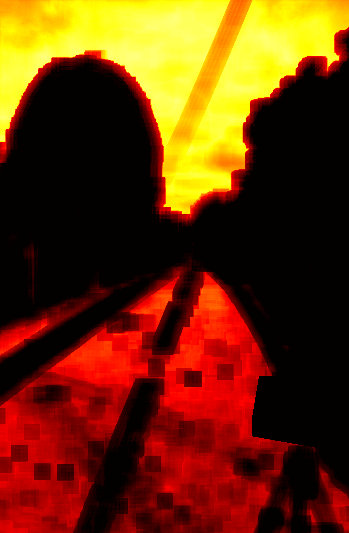} &
\includegraphics[width=0.350\columnwidth]{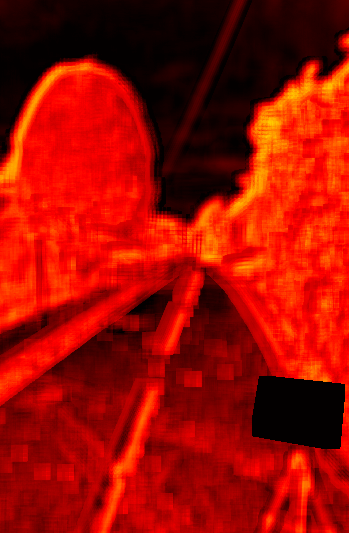} &
\includegraphics[width=0.350\columnwidth]{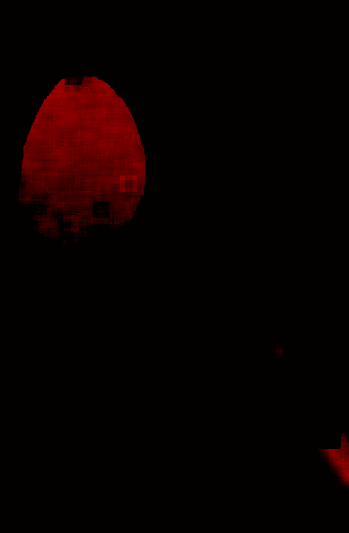} &
\includegraphics[width=0.350\columnwidth]{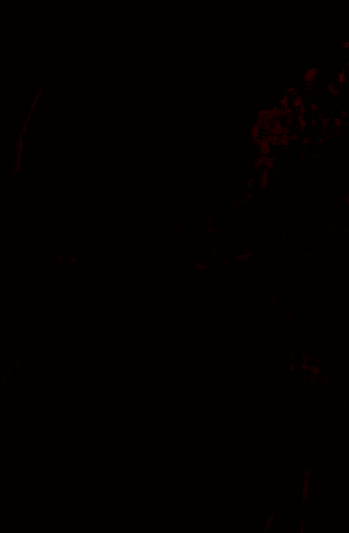} &
\includegraphics[width=0.350\columnwidth]{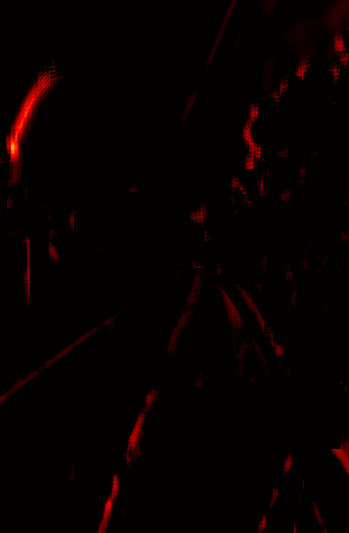} &
\includegraphics[width=0.350\columnwidth]{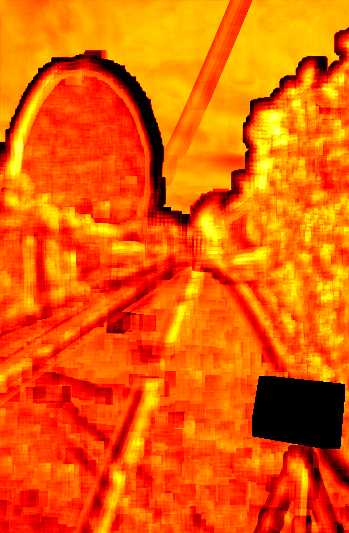} \\
&
\includegraphics[width=0.350\columnwidth]{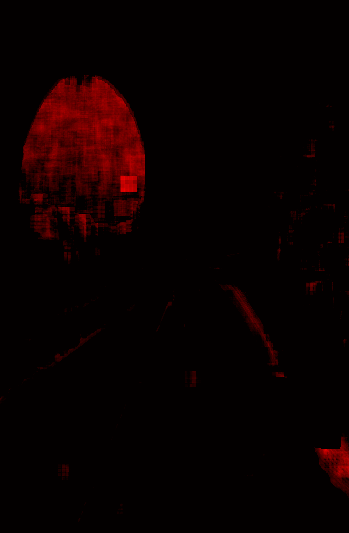} &
\includegraphics[width=0.350\columnwidth]{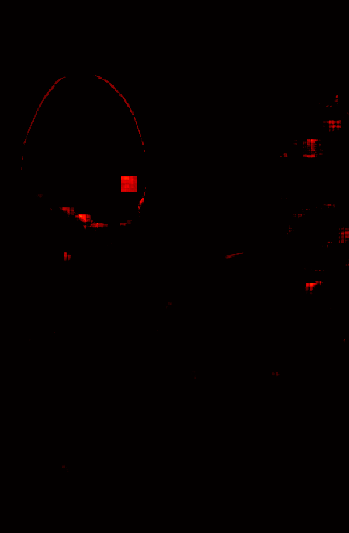} &
\includegraphics[width=0.350\columnwidth]{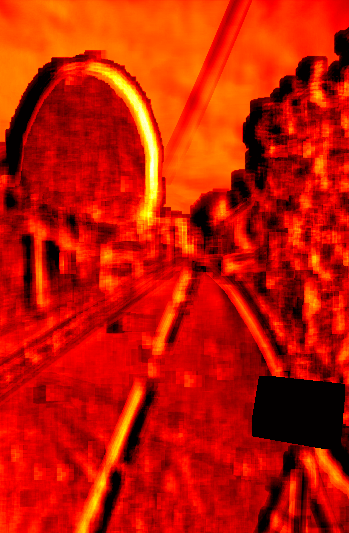} &
\includegraphics[width=0.350\columnwidth]{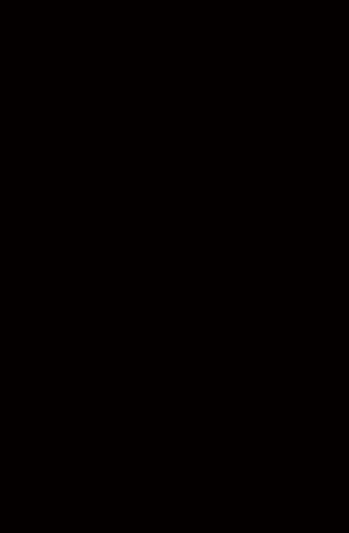} &
\includegraphics[width=0.350\columnwidth]{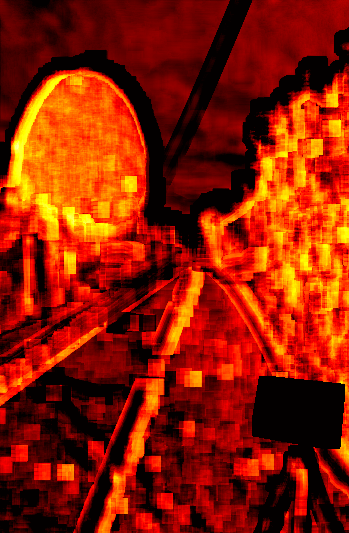} &
\includegraphics[width=0.350\columnwidth]{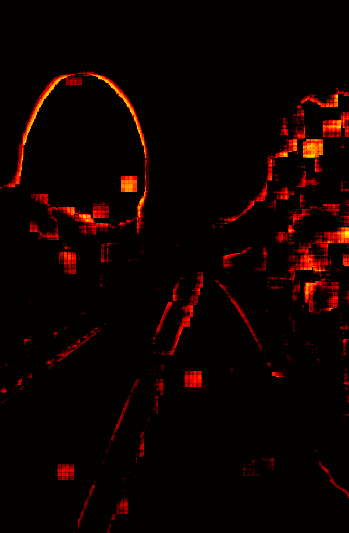} &
\includegraphics[width=0.350\columnwidth]{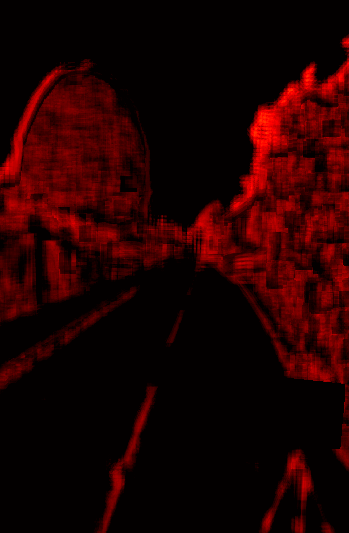} &
\includegraphics[width=0.350\columnwidth]{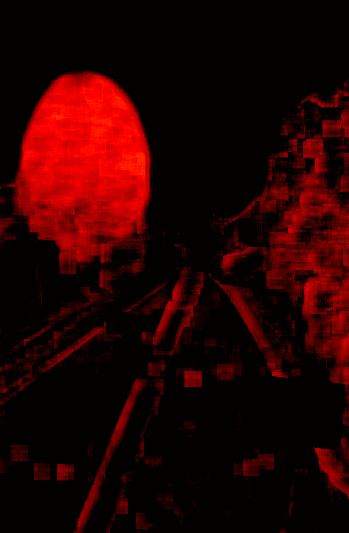} &
\includegraphics[width=0.350\columnwidth]{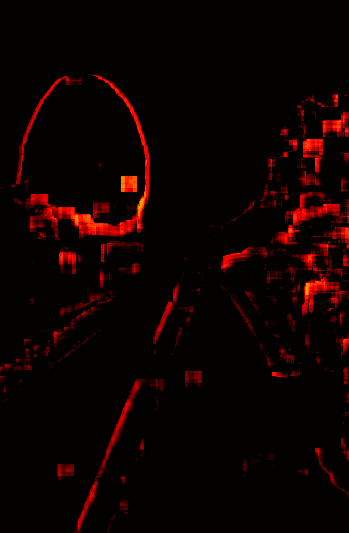} &
\includegraphics[width=0.350\columnwidth]{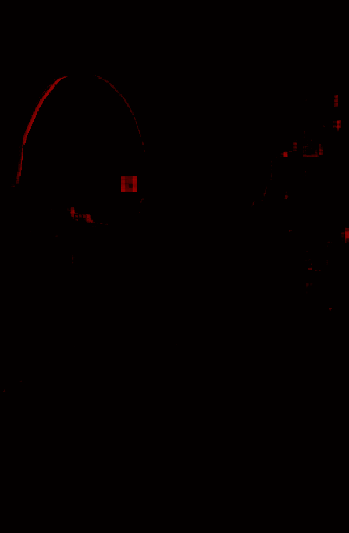} \\
\end{tabular}
}
\caption{Activation maps of the 40 neurons in the fully connected layer on all the patches of an entire image.}
\label{fig:activationMaps40neuC}
\end{figure*}

\ADD{Differently from the work by Funt et al.~\cite{funt1996learning}, our
network exploits some spatial information encoded in the 40 features
that are computed as linear combinations of the 240 convolutions after
that they have been pooled according to a $4 \times 4$ spatial grid.
In fact, as noted by Gijsenij et al.~\cite{gijsenij2007color}, the use
of spatial information brings an improvement over the application of
color constancy to the entire image.
To better understand the role of the 40 features, we report in
Figure~\ref{fig:patchactivations} the ten patches producing their
highest values.  The patches are taken from the first fold of the
Gehler-Shi dataset and are shown after the stretching of the color
channels.  It can be seen how different neurons are activated by different kinds
of patches.  Some of them are specialized in finding uniform patches
of a given dominant color (blue, red, green\dots) that often
correspond to specific content in the input images (sky,
vegetation\dots).  Several neurons are able to identify highlights, an
element that has been previously exploited for color constancy \cite{lee1986method}.  There are also neurons specialized in detecting
strong edges (that have also been used in the past \cite{vandeWGev07}) and patches with complex textures.
Figure~\ref{fig:activationMaps40neuC} shows the 40 activations on the
patches of a whole image, while those of five selected neurons on six
different images are shown in Figure \ref{fig:activationMaps40neu}.
These figures suggest that the network performs a rough analysis of
the content of the image by identifying the main elements of the
scene or by selecting elements that may be useful for the estimation
of illuminant.  For instance, neuron \#8 seems to fire on image edges,
neuron \#17 on highlights, neuron \#22 on sky and bluish texture,
neuron \#27 on skin and orange/reddish texture, neuron \#38 on
vegetation and greenish texture.  The use of semantic concepts share
some similarities with the work of van de Weijer et
al.~\cite{hilevelinfo} where the illuminant is estimated by maximizing
the likelihood of the colors associated to each semantic class.}

\setlength{\tabcolsep}{1.5pt}

\begin{figure*}[!ht]%
\setlength{\tabcolsep}{1.5pt}
\centering
\resizebox{\textwidth}{!} {
\begin{tabular}{cccccc}
\includegraphics[width=0.350\columnwidth]{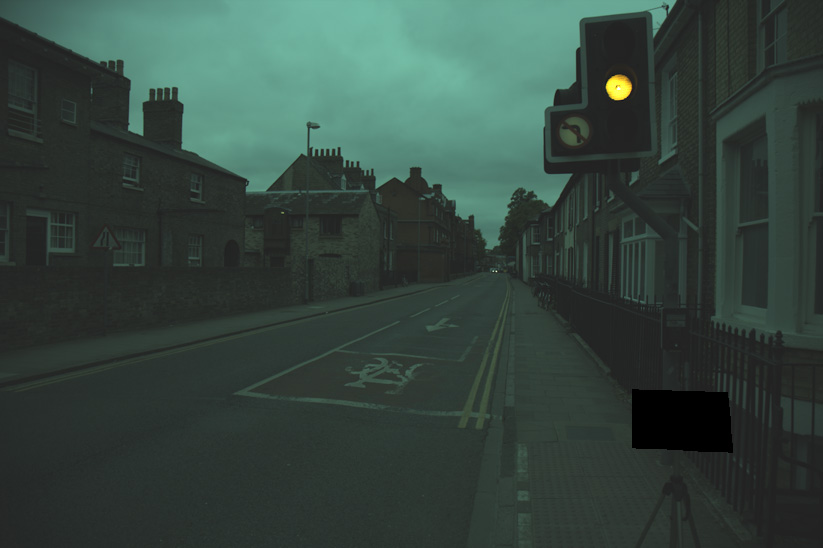} &
\includegraphics[width=0.350\columnwidth]{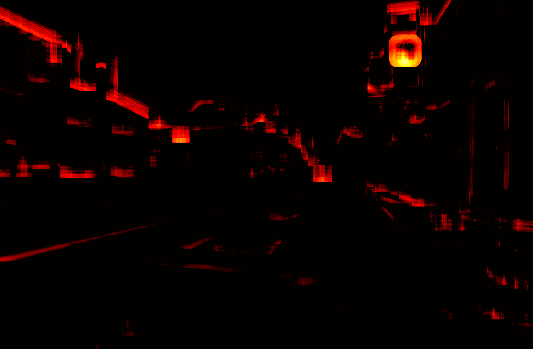} &
\includegraphics[width=0.350\columnwidth]{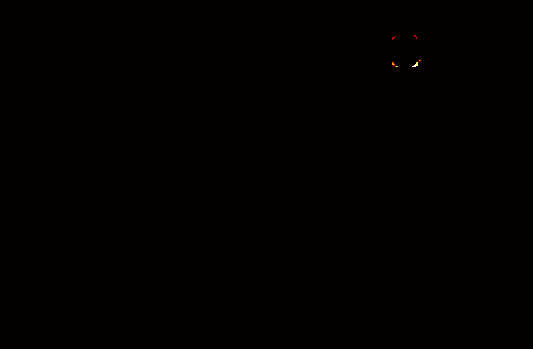} & 
\includegraphics[width=0.350\columnwidth]{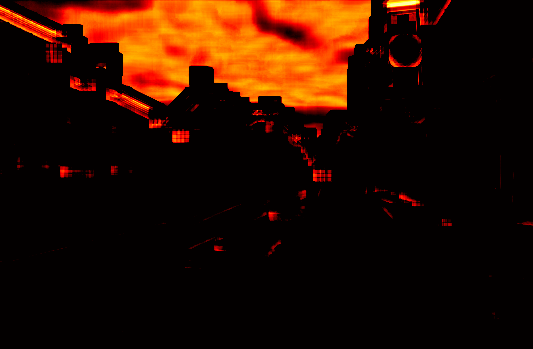} & 
\includegraphics[width=0.350\columnwidth]{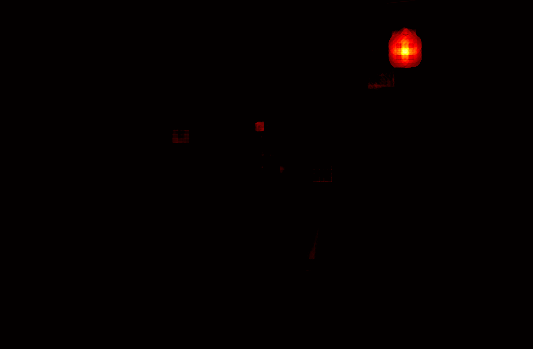} & 
\includegraphics[width=0.350\columnwidth]{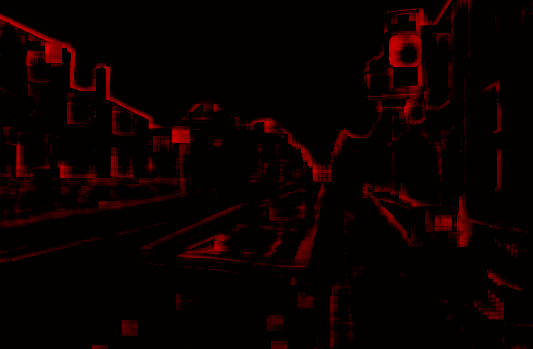} \\
\includegraphics[width=0.350\columnwidth]{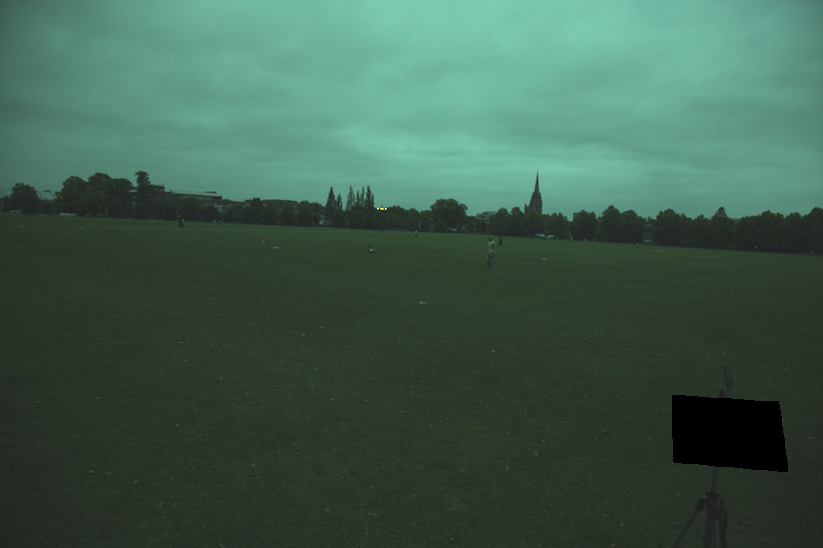} &
\includegraphics[width=0.350\columnwidth]{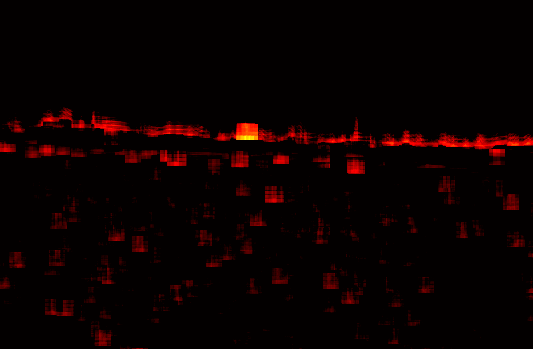} &
\includegraphics[width=0.350\columnwidth]{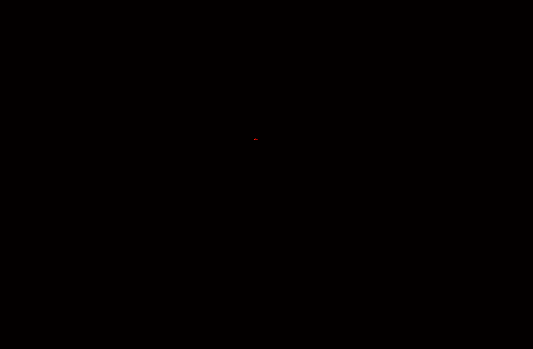} & 
\includegraphics[width=0.350\columnwidth]{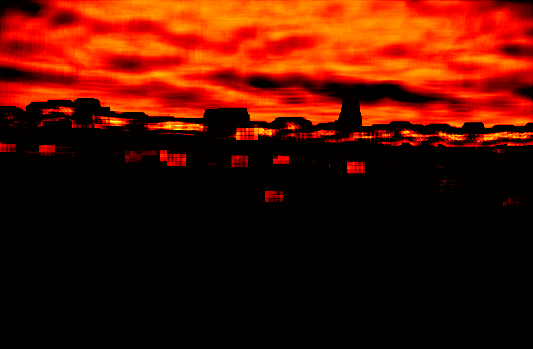} & 
\includegraphics[width=0.350\columnwidth]{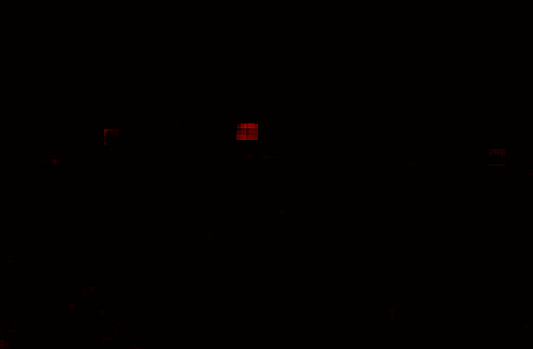} & 
\includegraphics[width=0.350\columnwidth]{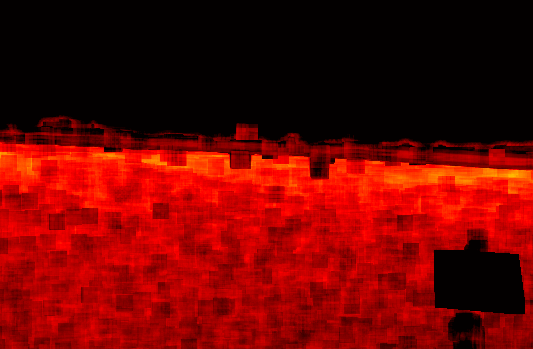} \\
\includegraphics[width=0.350\columnwidth]{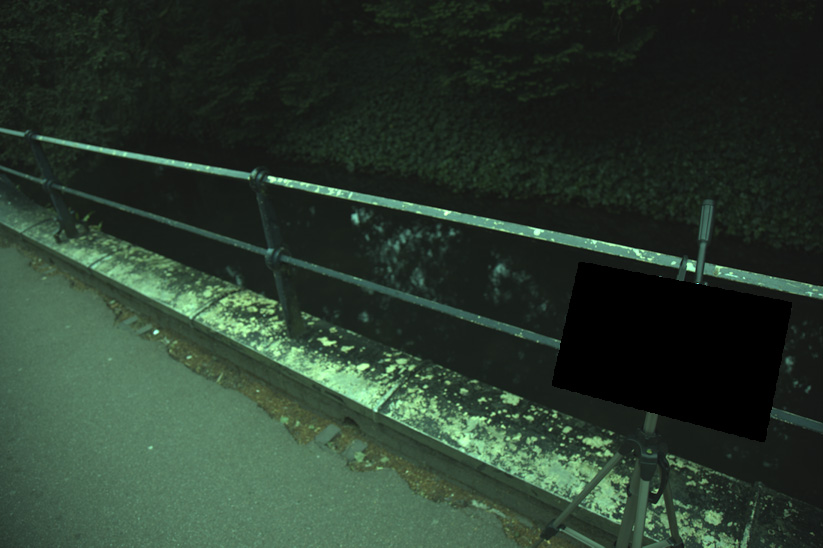} &
\includegraphics[width=0.350\columnwidth]{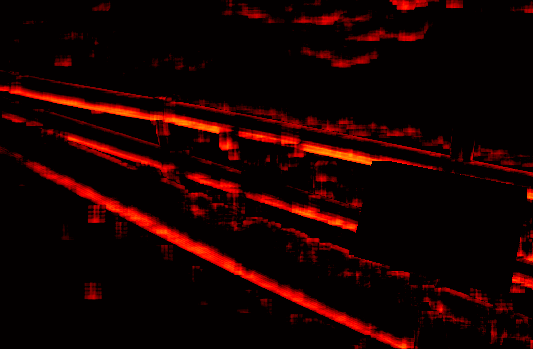} &
\includegraphics[width=0.350\columnwidth]{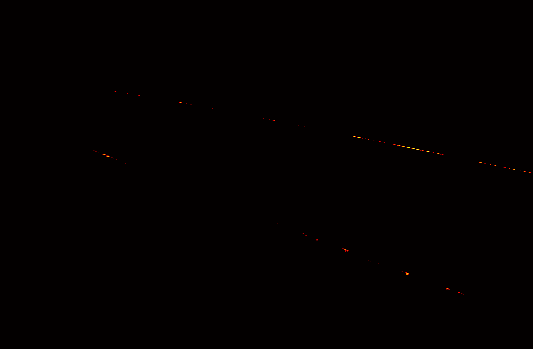} & 
\includegraphics[width=0.350\columnwidth]{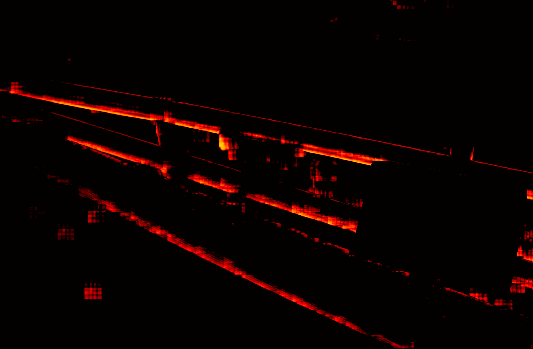} & 
\includegraphics[width=0.350\columnwidth]{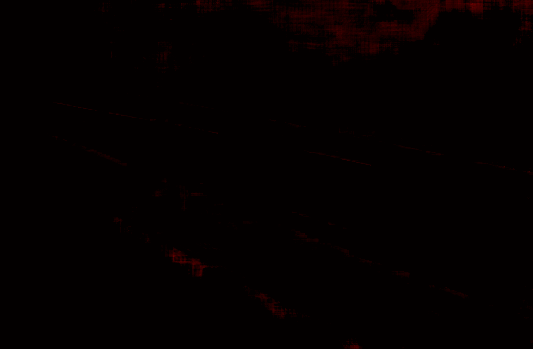} & 
\includegraphics[width=0.350\columnwidth]{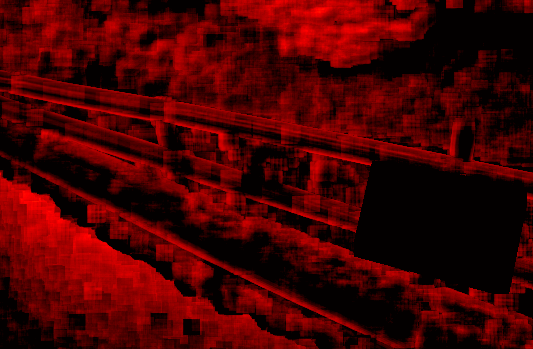} \\
\includegraphics[width=0.350\columnwidth]{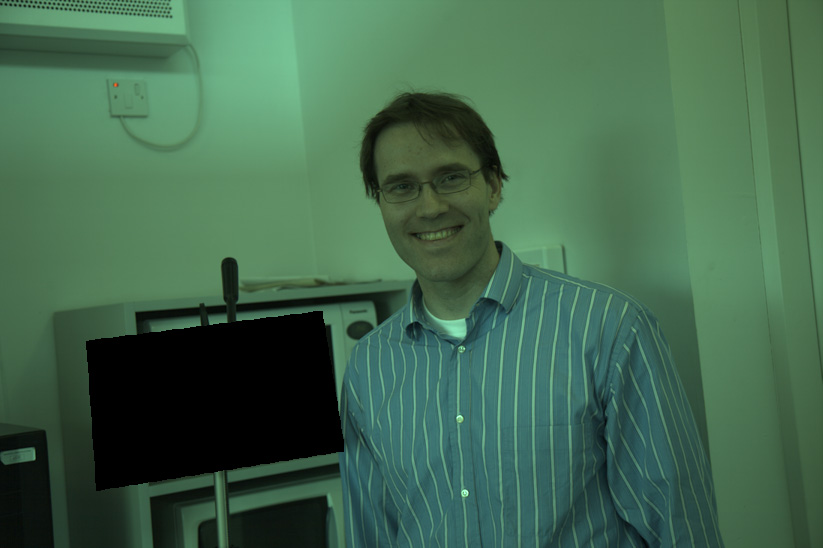} &
\includegraphics[width=0.350\columnwidth]{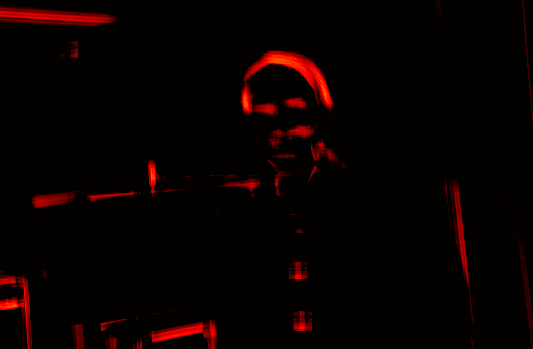} &
\includegraphics[width=0.350\columnwidth]{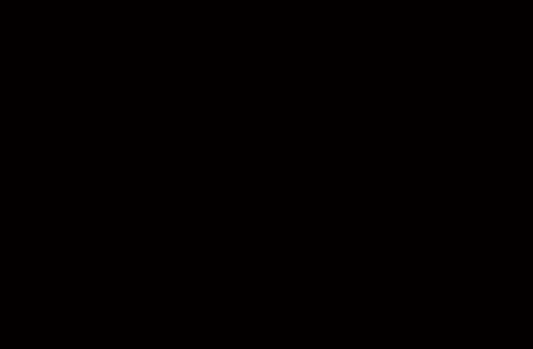} & 
\includegraphics[width=0.350\columnwidth]{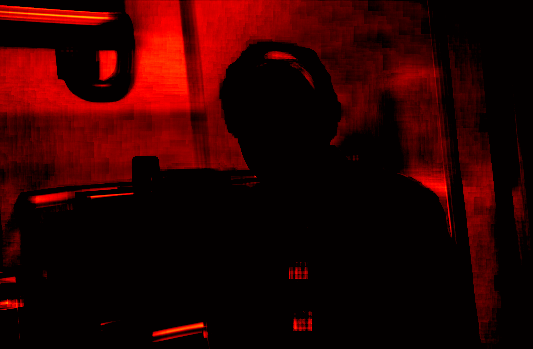} & 
\includegraphics[width=0.350\columnwidth]{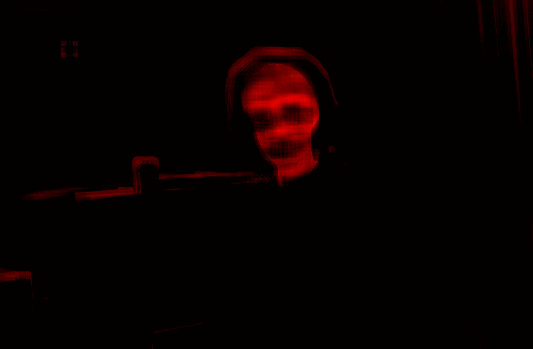} & 
\includegraphics[width=0.350\columnwidth]{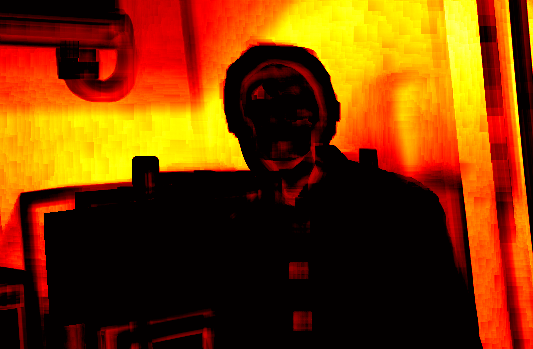} \\
\includegraphics[width=0.350\columnwidth]{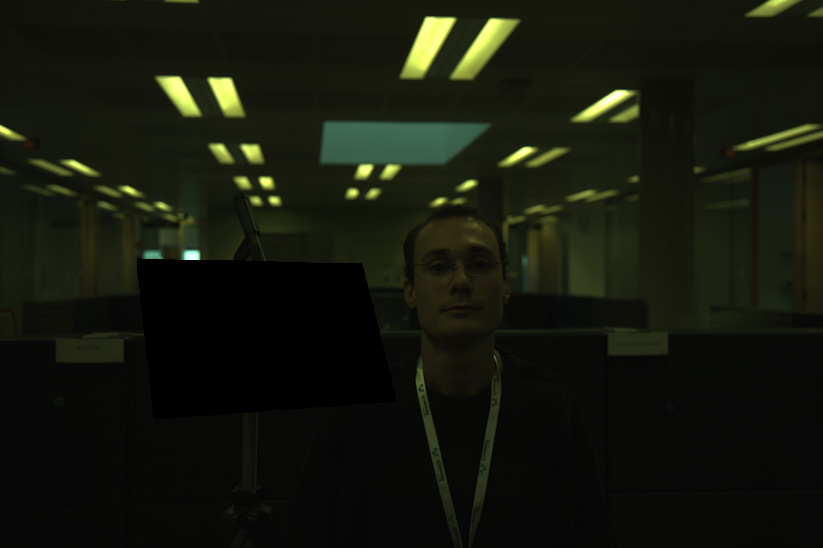} &
\includegraphics[width=0.350\columnwidth]{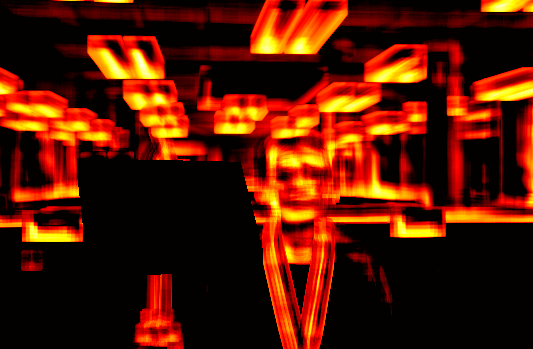} &
\includegraphics[width=0.350\columnwidth]{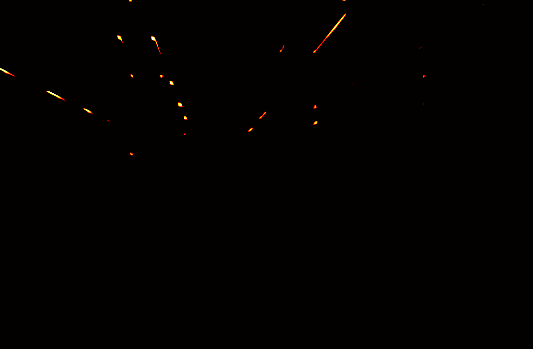} & 
\includegraphics[width=0.350\columnwidth]{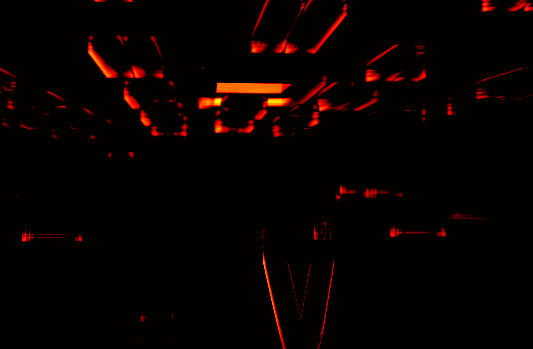} & 
\includegraphics[width=0.350\columnwidth]{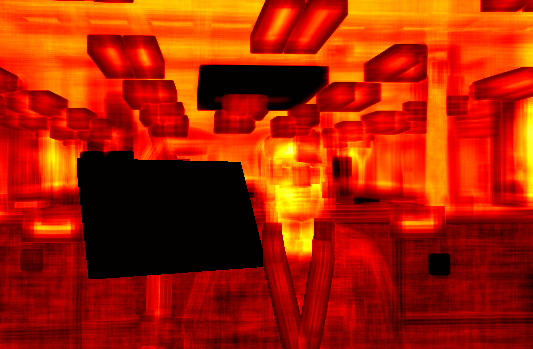} & 
\includegraphics[width=0.350\columnwidth]{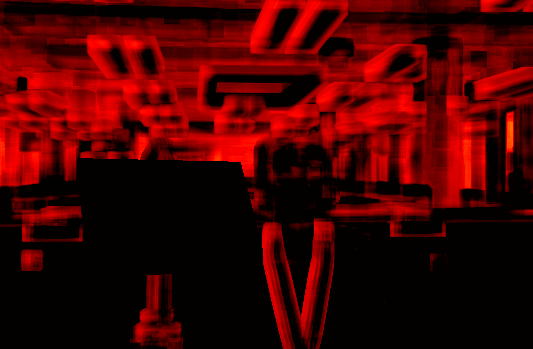} \\
\includegraphics[width=0.350\columnwidth]{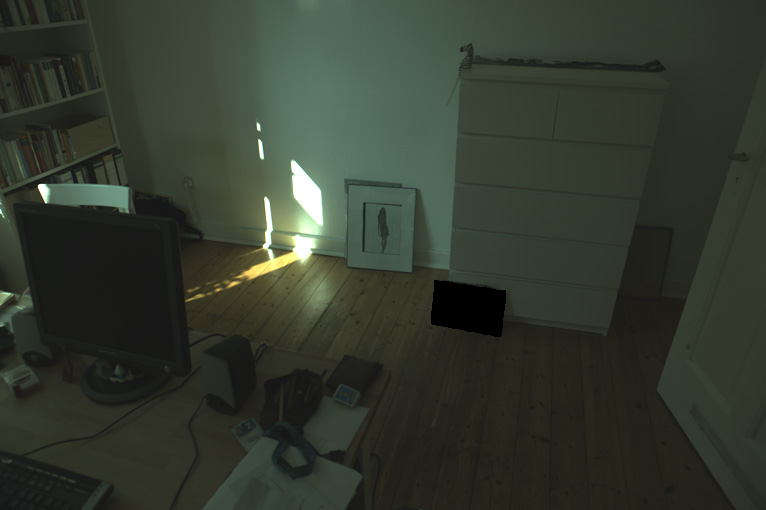} &
\includegraphics[width=0.350\columnwidth]{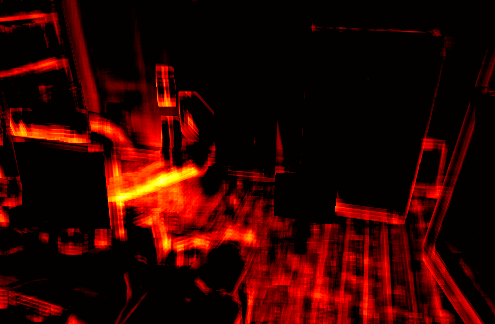} &
\includegraphics[width=0.350\columnwidth]{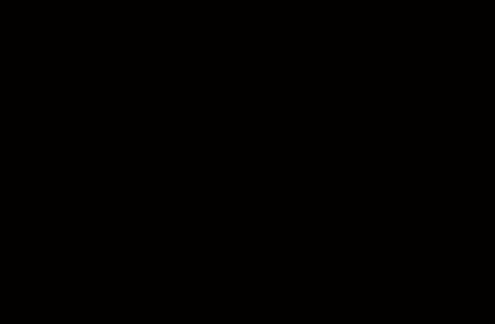} & 
\includegraphics[width=0.350\columnwidth]{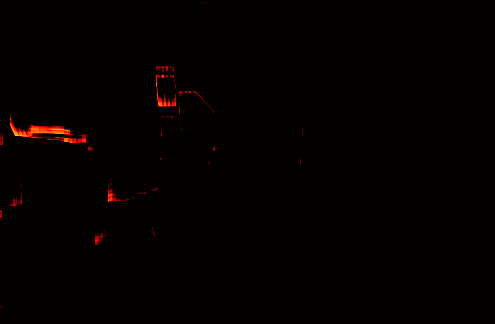} & 
\includegraphics[width=0.350\columnwidth]{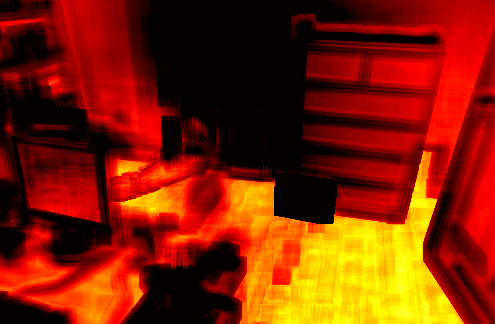} & 
\includegraphics[width=0.350\columnwidth]{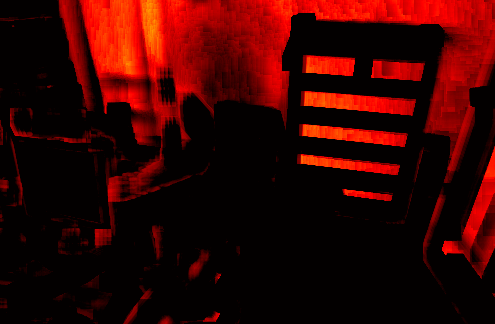} \\
 Input image &  Neuron \#8 &  Neuron \#17 &  Neuron \#22 &  Neuron \#27 &  Neuron \#38 \\
\end{tabular}
}
\caption{Activation maps of five selected neurons in the fully connected layer: neuron 8, 17, 22, 27  and 38.}
\label{fig:activationMaps40neu}
\end{figure*}

\begin{figure*}[!ht]
\includegraphics[width=1.0\textwidth]{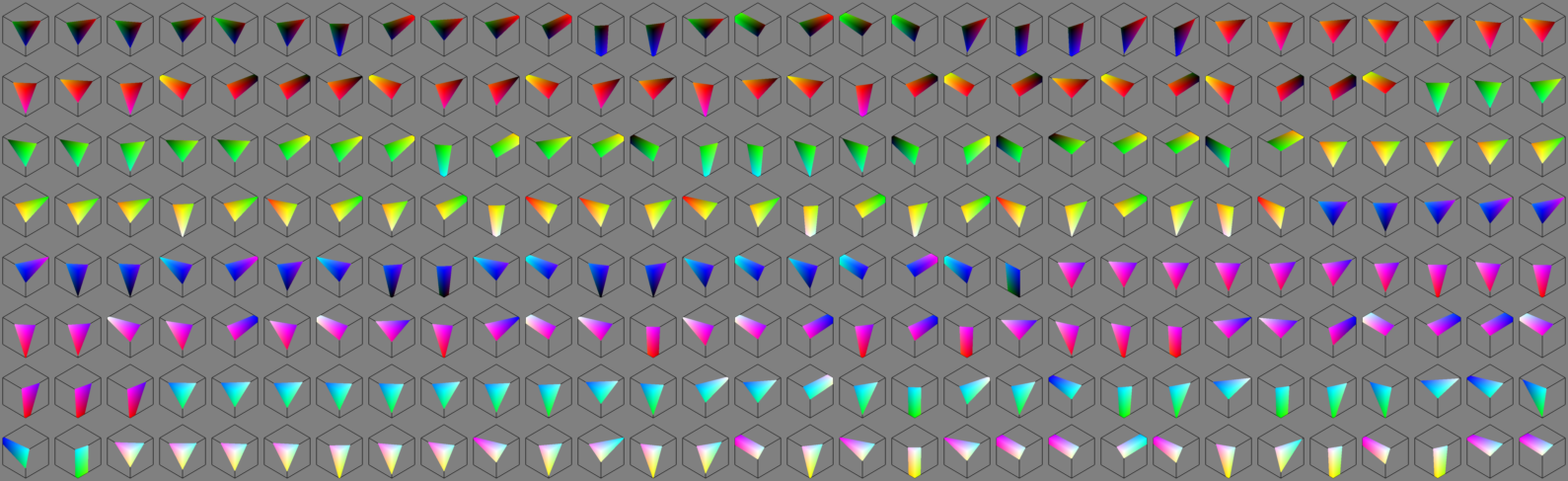}%
\caption{Surface of the RGB cube resulting in the higher 10\% responses of the 240 convolutional filters.}
\label{fig:cubes}%
\end{figure*}

\ADD{Finally, the first layer is of the convolutional kind, and it consists
of 240 units with $1 \times 1$ kernels.  The activation of each
convolutional unit can be seen as the projection over a specific
direction in the RGB cube.  Note that while $1 \times 1$ convolutions
do not exploit spatial information, they also preserve it unaltered
for the subsequent layers.  The combination of the 240 units forms a
sort of ``soft'' quantization of the color space that can be combined
by the pooling units to represent the local color distribution.
Figure~\ref{fig:cubes} shows how different regions of the RGB color
cube activate the 240 convolutional units.  Since each unit
corresponds to a linear projection, the maximum activation always
occur on a vertex of the RGB cube (to improve visualization, cubes are
rotated so that the region of maximum activation is always
front-facing).  It is possible to see that for all the eight vertexes
there are several units with high activations.  In practice this means
that the quantization learned by the network covers the whole color
space instead of being focused on specific colors.  Several units seem
redundant as they activate in presence of very similar colors.  We
observed, in fact, small differences in performance when we reduced
the number of convolutional units (see Figure~\ref{fig:parametri} b).}

\section{Conclusion} 
\label{sec:conclusions}
In this work we have developed a CNN-based color constancy algorithm that combines feature learning and regression as a complete
optimization process, which enables us to employ modern training
techniques to boost performance.  The network has been specially designed
to work on image patches in order to estimate the local illuminant
color. When our method detects a single illuminant in the image, the
local estimates are given as input to a trained local-to-global
\regressor{} which is able to predict the global illuminant with a
high accuracy.
\ADD{The experimental results showed that our algorithm improves the 
state-of-the-art performance on 
images with
a single illuminant. 
Experiments on a synthetically relighted
dataset with multiple illuminants showed that our method outperforms
all the general purpose local illuminant estimation methods in the
state of the art. Results are further confirmed on two real-world datasets with multiple illuminants, where our method is outperformed only by an illuminant estimation method exploiting the presence of faces.}
%
%
The results obtained suggest that a possible future research direction
is that of feeding additional semantic information in the form of
scene category or detected objects to further improve illuminant
estimation performance. 

Currently, our method is articulated in three separate steps.  In the
future we plan to merge them into a single estimation model.  In order
to allow the end-to-end learning of such a model, we are collecting a
large dataset composed by RAW images having both single and multiple
illuminants.

\ifCLASSOPTIONcaptionsoff
  \newpage
\fi



%

%
%

\bibliographystyle{IEEEtran}
\bibliography{colorConstancyCNN_TPAMI_v3_arxiv}

%

\begin{IEEEbiographynophoto}{Simone Bianco}
 obtained the BSc and the MSc degree in Mathematics from
the University of Milano-Bicocca, Italy, respectively in 2003 and 2006.
He received the PhD in Computer Science at Department of Informatics,
Systems and Communication of the University of Milano-Bicocca, Italy,
in 2010, where he currently is a post-doc. His research interests include
computer vision, optimization algorithms, machine learning, and color
imaging.
\end{IEEEbiographynophoto}

\begin{IEEEbiographynophoto}{Claudio Cusano}
   received the Laurea and PhD degrees from the
  University of Milano Bicocca in 2002 and 2006, respectively.  He has
  been a researcher with grant at the ITC institute of the Italian
  National Research Council and then at the Imaging and Vision
  Laboratory of the University of Milano-Bicocca. Currently, he is
  assistant professor at the Department of Electrical, Computer and
  Biomedical Engineering of the University of Pavia. The main topics
  of his research concern 2D and 3D imaging, with a particular focus
  on image analysis and classification, and on face recognition.
\end{IEEEbiographynophoto}


\begin{IEEEbiographynophoto}{Raimondo Schettini}
 is a professor at the University of Milano Bicocca (Italy).
He is head of Imaging and Vision Lab and Vice-Director of the Department
of Informatics, Systems and Communication. He has been associated
with Italian National Research Council (CNR) since 1987 where
he has leaded the Color Imaging lab from 1990 to 2002. He has been
team leader in several research projects and published more than 200
refereed papers and six patents about color reproduction, and image
processing, analysis and classification. He has been recently elected
Fellow of the International Association of Pattern Recognition (IAPR)
for his contributions to pattern recognition research and color image
analysis.
\end{IEEEbiographynophoto}




\end{document}